\documentclass[times,twocolumn,final]{elsarticle}

\usepackage{pifont} 
\usepackage{adjustbox}
\usepackage{algorithmic}
\usepackage{amsfonts}
\usepackage{amsmath}
\usepackage{amssymb}
\usepackage{array}
\usepackage{arydshln}
\usepackage{booktabs}
\usepackage{color}
\usepackage{comment}
\usepackage{etoolbox}
\usepackage{framed}
\usepackage{graphicx}
\usepackage{hyperref}
\usepackage{ifthen}
\usepackage{latexsym}
\usepackage{makecell}
\usepackage{multicol}
\usepackage{multirow}
\usepackage{pifont}
\usepackage{tabularx}
\usepackage{tabulary}
\usepackage{textcomp}
\usepackage{url}
\usepackage{xcolor}
\usepackage{colortbl}
\usepackage{multirow}
\usepackage{booktabs}
\usepackage{wasysym}
\usepackage{cleveref}
\usepackage[
  separate-uncertainty = true,
  multi-part-units = repeat
]{siunitx}

\definecolor{newcolor}{rgb}{.8,.349,.1}
\usepackage{medima}
\usepackage[numbers]{natbib}

\begin{document}

\title{SurgLaVi: Large-Scale Hierarchical Dataset for Surgical Vision-Language \\ Representation Learning}  

\author[1,2]{Alejandra Perez \fnref{fn1}} 
\author[1]{Chinedu Nwoye}
\author[1]{Ramtin Raji Kermani}
\author[1]{Omid Mohareri}
\author[1]{Muhammad Abdullah Jamal\corref{cor1}}
\cortext[cor1]{Corresponding author}
\ead{abdullah.jamal@intusurg.com}

\fntext[fn1]{Work done during an internship at Intuitive Surgical Inc.}

\affiliation[1]{
organization={Intuitive Surgical, Inc.},
city={Sunnyvale},
state={CA},
country={United States}}
            
\affiliation[2]{
organization={Center for Research and Formation in Artificial Intelligence (CinfonIA), Universidad de los Andes},
city={Bogotá},
country={Colombia}}

% \verso{SurgLaVi}

\begin{abstract}
\noindent \textbf{Abstract:} Vision–language pre-training (VLP) offers unique advantages for surgery by aligning language with surgical videos, enabling workflow understanding and transfer across tasks without relying on expert-labeled datasets. However, progress in surgical VLP remains constrained by the limited scale, procedural diversity, semantic quality, and hierarchical structure of existing datasets. In this work, we present SurgLaVi, the largest and most diverse surgical vision–language dataset to date, comprising nearly 240k clip–caption pairs from more than 200 procedures, and featuring hierarchical levels at coarse-, mid-, and fine-level.
At the core of SurgLaVi lies a fully automated pipeline that systematically generates fine-grained transcriptions of surgical videos and segments them into coherent procedural units. To ensure high-quality annotations, it applies dual-modality filtering to remove irrelevant and noisy samples. Within this framework, the resulting captions are enriched with contextual detail, producing annotations that are both semantically rich and easy to interpret. To ensure accessibility, we release \href{https://github.com/aperezr20/SurgLaVi}{SurgLaVi-$\mathbf{\beta}$}, an open-source derivative of 113k clip–caption pairs constructed entirely from public data, which is over four times larger than existing surgical VLP datasets. To demonstrate the value of the SurgLaVi datasets, we introduce SurgCLIP, a CLIP-style video–text contrastive framework with dual encoders, as a representative base model. SurgCLIP achieves consistent improvements across phase, step, action, and tool recognition, surpassing prior state-of-the-art methods, often by large margins. These results validate that large-scale, semantically rich, and hierarchically structured datasets directly translate into stronger and more generalizable representations, establishing SurgLaVi as a key resource for developing surgical foundation models.

\end{abstract}
\maketitle

\section{Introduction}\label{introduction}
\begin{figure}[!t]
  \centering
    \includegraphics[width=0.48\textwidth]{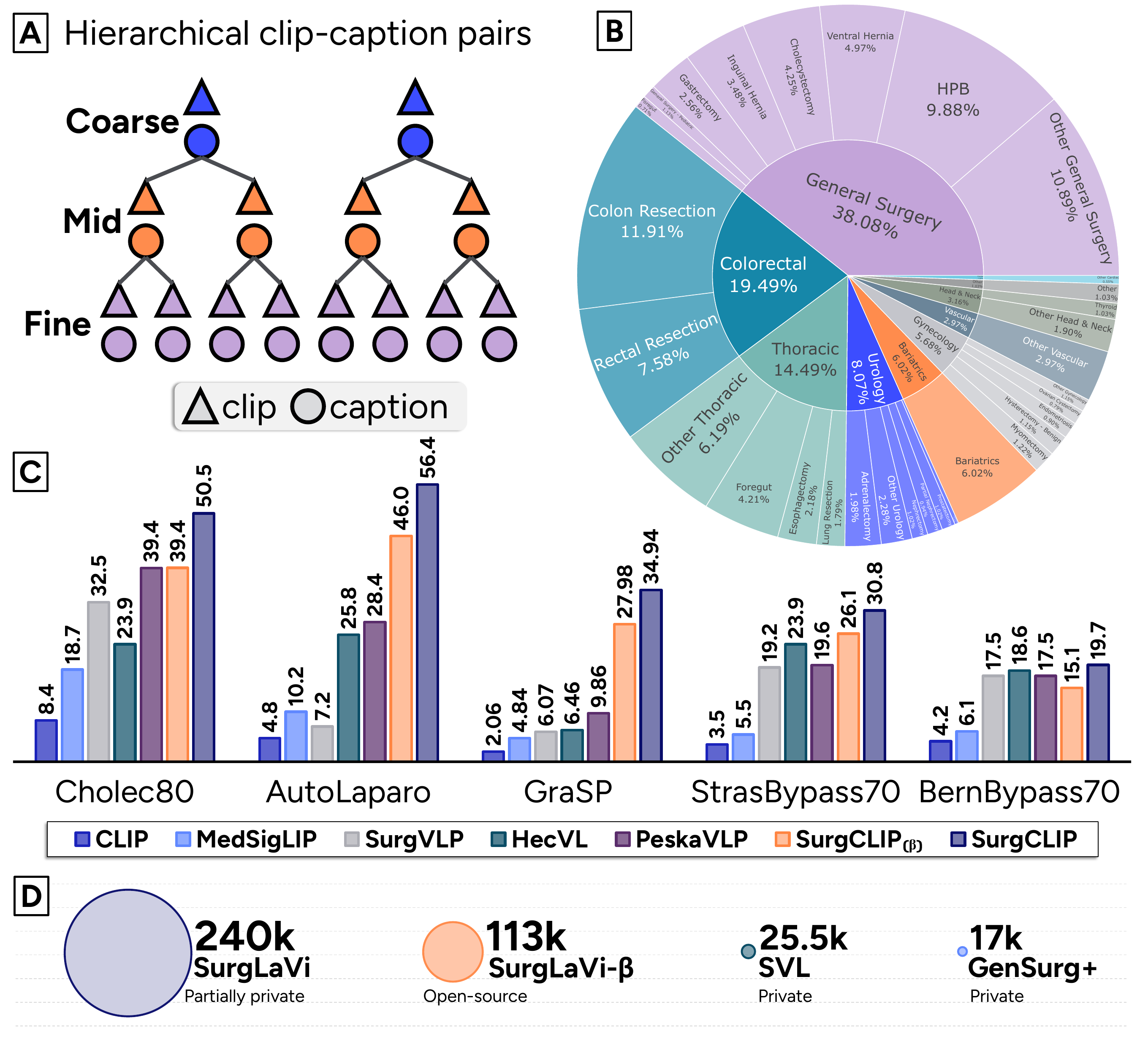}
    \caption{\textbf{Overview of SurgLaVi.} 
(A) Hierarchical clip–caption pairs at \textcolor{black}{coarse, mid, and fine} levels provide multi-scale temporal granularity for language-supervised analysis. 
\textcolor{black}{(B) Specialty and subject video distribution, demonstrating broad coverage across diverse surgical domains.} 
(C) Zero-shot phase recognition \textcolor{black}{F1 performance}  across five datasets, where SurgCLIP trained on SurgLaVi and SurgLaVi$-\mathbf{\beta}$ surpass prior state-of-the-art approaches. 
(D) Dataset scale comparison, showing that SurgLaVi and SurgLaVi$-\mathbf{\beta}$ substantially exceed existing surgical VLP datasets.}
    \label{fig:overview}
\end{figure}

Vision–language pre-training (VLP) aligns visual and textual modalities in a joint embedding space, grounding visual features in language and enabling a wide range of downstream tasks such as open-vocabulary recognition and retrieval~\citep{clip}, captioning~\citep{li2022blip}, visual question answering~\citep{lu2019vilbert}, and spatial grounding~\citep{gu2021open}. Seminal works like CLIP~\citep{clip} and SigLIP~\citep{zhai2023sigmoid} established this paradigm by training on large-scale image–text pairs, demonstrating strong transfer without task-specific supervision and reducing the dependence on manually labeled datasets. Beyond natural images, this approach is especially valuable in surgery: aligning procedural language with surgical videos facilitates workflow understanding, supports transfer across tasks such as phase, step, and instrument recognition, and mitigates the existing bottleneck of scarce expert-labeled datasets in surgery. 

Recent efforts have begun adapting VLP to the surgical domain by segmenting academic surgical videos into clips and pairing them with automatic audio transcriptions~\citep{surgvlp,vidlpro}, demonstrating stronger generalization than models pre-trained exclusively on natural images.
Despite these advances, surgical video–language pre-trained models still lag significantly behind task-specific supervised approaches in downstream performance. While increasingly elaborate pre-training frameworks have been proposed to address this gap~\citep{vidlpro, peskavlp}, we argue that the fundamental limitation lies not primarly in model design but in the datasets themselves, which face constraints across several dimensions. First, current pipelines to build VLP datasets rely on heuristic rules to segment academic surgical videos into clips, which frequently truncate or fragment coherent surgical concepts, resulting in clip–caption pairs that lack complete semantic meaning. Although some works have sought to enrich semantics with broader supervision, such as phase-level annotations or video-level abstracts~\citep{hecvl, peskavlp}, these labels are too coarse to capture the fine-grained, hierarchical structure of surgical workflows, which are composed of actions, tasks, steps, and phases~\citep{sages_consensus}. Second, existing datasets contain a substantial proportion of noisy pairs, meaning the text does not accurately describe the corresponding visual content or contains information irrelevant to surgical workflows. Such noise typically results from temporal misalignment between video and narration due to inaccurate transcription tools, the inclusion of non-surgical segments present in academic presentations, and narrator commentary that diverges from the surgical scene. Third, existing datasets struggle with scale and diversity. For instance, SVL~\citep{surgvlp} offers 25k clip–caption pairs but focuses exclusively on laparoscopic procedures, while GenSurgery+~\citep{vidlpro} broadens coverage to robotic-assisted surgery, yet remains smaller in scale with only 17k pairs. Finally, none of the existing datasets are openly available, further constraining reproducibility and progress in the field. Together, these issues reveal that advancing surgical VLP requires rethinking not only model pre-training frameworks but, more fundamentally, the construction of datasets that better reflect the semantics, structure, and diversity of surgical workflows. 

In response to these limitations, we propose a three-part contribution consisting of: (i) the collection of a large and diverse corpus of surgical videos, and (ii) the design of a novel, fully automated, and scalable pipeline to generate high-quality clip–caption pairs that are temporally precise, semantically rich across different levels of granularity, and representative of real surgical workflows, and (iii) the design of a simple lightweight base model with rigorous and exhaustive experimental study on the dataset, validating that well-curated data enables minimal architectures to surpass more complex models trained on suboptimally designed datasets, highlighting the decisive role of our dataset quality and scale over model complexity.
Our dataset pipeline systematically integrates four key components: (i) \textit{fine-grained transcription} of surgical videos, (ii) \textit{semantic hierarchical segmentation} that partitions videos into clips representing procedural units at \textcolor{black}{coarse-, mid-, and fine-} level granularity, (iii) \textit{dual-modality filtering} to discard non-surgical visual content and incomplete or non-descriptive captions, and (iv) \textit{contextual enrichment} that improves caption quality using preceding narrations and video metadata. Leveraging this pipeline, we build the SurgLaVi dataset, the largest and most diverse surgical vision–language data resource to date, comprising nearly 240k clip–caption pairs from over 200 distinct procedure types,                as illustrated in Fig.~~\ref{fig:overview}. By structuring data hierarchically, SurgLaVi captures complete procedural semantics across multiple temporal scales, enabling representations that are more discriminative at different levels of surgical recognition. Using our automated pipeline, we also construct SurgLaVi$-\mathbf{\beta}$, an open-source derivative of 113k clip–caption pairs built entirely from public data, which is more than 4x larger than existing surgical vision–language datasets and is publicly released \footnote{\url{https://github.com/aperezr20/SurgLaVi} to foster research.} 

To assess SurgLaVi’s utility for representation learning, we present SurgCLIP, a lightweight CLIP-style contrastive framework with dual video and text encoders. We pretrain SurgCLIP on SurgLaVi and explore    its effectiveness on four downstream tasks: phase, step, action, and tool recognition, spanning seven public laparoscopic and robotic surgical datasets. Evaluation under zero-shot as well as few- and full-shot linear probing demonstrates superior performance. \textcolor{black}{Remarkably, even when pre-trained on SurgLaVi$-\mathbf{\beta}$, a smaller subset of the dataset, SurgCLIP consistently exceeds previous state-of-the-art F1 performance on several phase recognition benchmarks, including Cholec80 (+0.3\%), AutoLaparo (+17.54\%), StrasBypass70 (+6.41\%), Heichole (+3.62\%) and GraSP (+18.12\%). When trained on the full SurgLaVi corpus, the model achieves even stronger results, showing substantial zero shot F1 gains across different benchmarks (+10\% on Cholec80, +28\% on AutoLaparo, +11.14\% on StrasBypass70, +1.1\% on BernBypass70, +10.51\% on Heichole, and +25\% on GraSP)}. Beyond phase recognition, SurgCLIP also advances performance on step, action, and tool recognition benchmarks. 
These results highlight that pre-training on SurgLaVi’s hierarchically structured dataset enables a standard model to effectively encode procedural workflows without relying on sophisticated architectures or specialized hierarchical pre-training techniques. Extensive ablation studies further validate the design choices and confirm the robustness of our approach.

\section{Related Work}\label{related_work}

\begin{figure*}[h]
  \centering
    \includegraphics[width=1\textwidth]{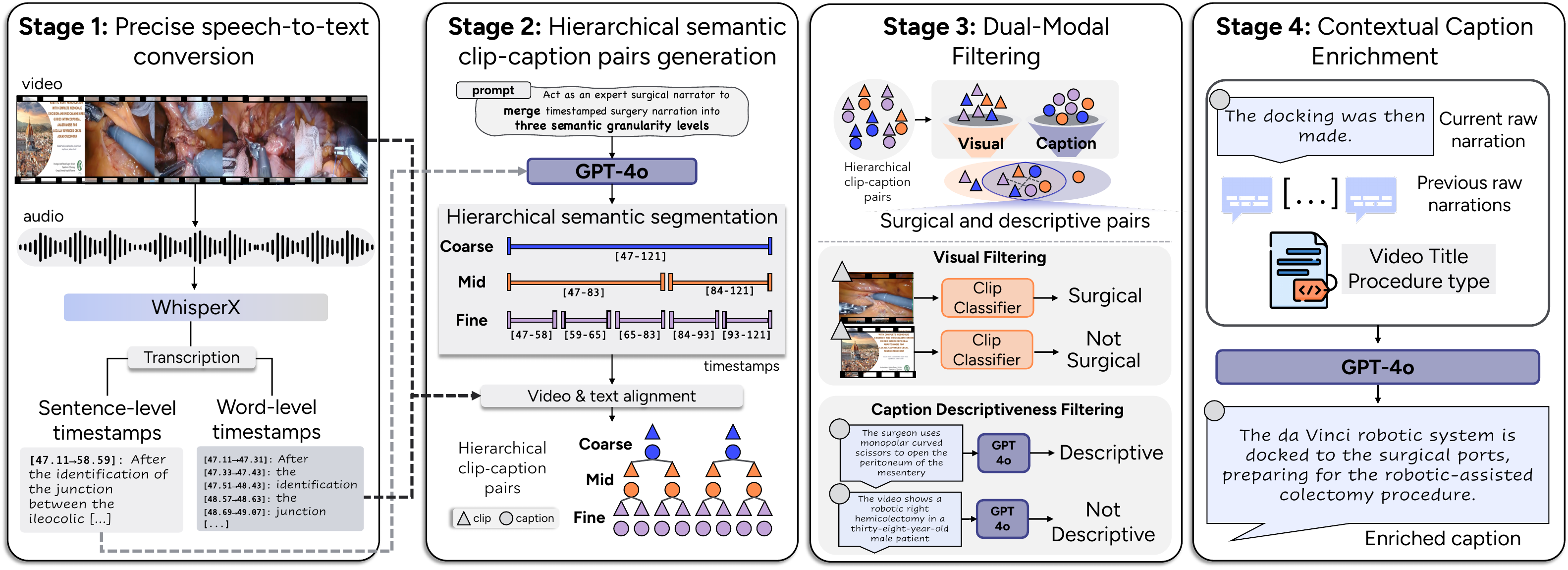}
    \caption{\textbf{SurgLaVi Data Processing Pipeline Overview.} \textbf{Stage 1:} Speech-to-text conversion with fine-grained timestamps. \textbf{Stage 2:} Hierarchical semantic transcript segmentation followed by video–text alignment to generate semantic clip-caption pairs at \textcolor{black}{coarse-, mid-, and fine-level} of granularity. \textbf{Stage 3:} Dual-modality filtering for surgical visual relevance and textual descriptiveness pairs. \textbf{Stage 4:} Contextual caption enrichment using prior context and metadata to enhance  clip-caption semantic alignment.}\label{fig: pipeline}
\end{figure*}

\subsection{Surgical Vision-Language Pre-Training Datasets}
The scarcity of expert surgical annotations has motivated weakly supervised approaches that pair surgical videos with narrations as contextual grounding. The SVL dataset~\citep{surgvlp} pioneered this direction, introducing $\sim$25k clip–caption pairs from lecture videos using dual automatic speech recognition (ASR) systems: AWS Medical Transcribe~\citep{medicalaws} and Whisper~\citep{radford2022whisper}. While this strategy combined strengths in medical terminology recognition and sentence coherence, it suffered from inaccurate timestamps, leading to temporal misalignment. SVL dataset was further limited by sparse random clip sampling, narrowed diversity by just covering laparoscopic procedures, and the absence of workflow structure. To address this lack of workflow representation, HecVL~\citep{hecvl} augmented SVL with phase-level summaries and video-level abstracts, and PeskaVLP~\citep{peskavlp} leveraged large language models to correct narration errors, expand phase descriptions, and generate abstracts. These extensions enriched textual context but ultimately inherited SVL’s small scale, noisy alignments, and narrowed procedural range. To increase coverage, GenSurg+~\citep{vidlpro} turned to publicly available YouTube data, producing $\sim$17k clip–caption pairs across 28 robotic and laparoscopic procedures. While this increased procedural diversity, the dataset was even smaller than SVL and relied on a fixed segmentation of 45-second clips, limiting semantic coherence as well. In parallel, ophthalmological surgery demonstrated the potential of large-scale curation with OphVL~\citep{ophclip}, which assembled $>$375k pairs through ASR denoising with SurgicBERTa~\citep{SurgicBERTa} validation and structured metadata from silent videos. OphVL showed that scale and hierarchical structure can drive strong generalization, but its scope remained confined to a single specialty. Overall, these efforts illustrate an iterative progression, each addressing shortcomings of its predecessors, yet surgical VLP datasets remain fundamentally constrained by scale, temporal alignment, procedural diversity, and semantic quality.

\subsection{Vision-Language Pre-training Models in Surgery}
Computer vision frameworks such as CLIP~\citep{clip}, ALIGN~\citep{jia2021scaling}, and SigLIP~\citep{zhai2023sigmoid} achieve strong performance through standard contrastive pre-training on large-scale image-text corpora, typically containing millions of pairs. In contrast, the surgical domain has needed creative strategies to achieve representation quality with small, noisy pre-training resources. The foundational SurgVLP~\citep{surgvlp} demonstrated that domain-specific pre-training outperformed generic CLIP models by learning joint video-text embeddings through dual-branch contrastive learning, yet it could not  capture hierarchical workflow structure. To address this limitation, HecVL~\citep{hecvl} introduced multi-level contrastive learning across clip-, phase-, and video-level embeddings, creating separate embedding spaces and employing alternating training strategies to capture both short-term actions and long-term procedural context without catastrophic forgetting. However, HecVL's multi-space approach increased model complexity while struggling with textual quality issues. PeskaVLP~\citep{peskavlp} tackled the caption quality problem through hierarchical knowledge augmentation using large language models to correct, expand, and summarize texts at different hierarchical levels, while introducing procedure-aware training with Dynamic Time Warping-based contrastive regularization to enforce temporal alignment between video frames and text sequences. Despite these semantic improvements, the approach required complex alternating training schedules and is computationally intensive. Following these approaches and moving towards unified pre-training frameworks, VidLPRO~\citep{vidlpro} employed a multi-modal fusion approach with cross-attention mechanisms, combining video-text contrastive learning, video-text matching, and masked language modeling objectives in a single stage, though this increased architectural complexity compared to dual-encoder designs. Meanwhile, in ophthalmology, OphCLIP~\citep{ophclip} demonstrated retrieval-augmented learning by incorporating silent videos through a memory bank system that matches narrative videos with relevant procedures, achieving hierarchical representation learning through alternating clip- and video-level pretraining. These collective efforts represent increasingly sophisticated architectural approaches to extract maximum representational value from limited supervision. However, model complexity alone cannot overcome fundamental training constraints, with each innovation introducing new computational or methodological trade-offs.

\section{SurgLaVi Dataset and Processing Pipeline}
\label{sec:processing_pipeline}

\noindent As illustrated in Figure~\ref{fig: pipeline}, we introduce a novel, fully automated and scalable four-stage pipeline for generating hierarchical video-language datasets from surgical videos. This process leverages different foundation models to perform key operations, including transcription, temporal segmentation, and filtering; thereby enabling generation of hierarchical annotations that describe surgical workflows under a language supervision paradigm. The following subsections describe each stage of our proposed pipeline:

\subsection{Stage 1: Speech-to-text conversion}
\label{subsec:stage1}
\noindent  To convert speech narrations from videos to text, we use WhisperX (v3) \textcolor{black}{as it demonstrated superior performance within the surgical domain compared to other ASR models  (Table \ref{tab:whisper_exps} in Appendix)}. This model, denoted as $W(\cdot)$, generates sentence-level timestamped transcriptions and then adds a forced phoneme alignment to generate millisecond-accurate word-level timestamps. This fine-grained alignment enables flexible and accurate video-text pairing in subsequent stages. Formally, given a video dataset $D = \{ (v_1, a_1), (v_2, a_2), \dots, (v_N, a_N) \} $, where $v_i$ is the $i$-th video and $a_i$ is its corresponding audio track, we apply WhisperX to each audio $S_i, \; T_i \leftarrow W(a_i)$ to obtain two terms.

The first term is the \textit{sentence-level transcription}:
\begin{equation}
    S_i = \{ (s_{i_1}, [t_{i_1}^{\text{start}}, t_{i_1}^{\text{end}}]), \dots, (s_{im_i}, [t_{im_i}^{\text{start}}, t_{im_i}^{\text{end}}]) \},
\end{equation}
where $s_{ij}$ is the $j$-th sentence in $v_i$ and $[t_{ij}^{\text{start}}, t_{ij}^{\text{end}}]$ are its start and end timestamps.
The 2nd term is the \textit{word-level transcription}:
\begin{equation}
T_i = \{ (w_{i_1}, [\tau_{i_1}^{\text{start}}, \tau_{i_1}^{\text{end}}]), \dots, (w_{ik_i}, [\tau_{ik_i}^{\text{start}}, \tau_{ik_i}^{\text{end}}]) \},
\end{equation} 
where $w_{il}$ is the $l$-th word in $v_i$ and $[\tau_{il}^{\text{start}}, \tau_{il}^{\text{end}}]$ are its start and end timestamps.
Videos with empty transcriptions ($S_i = \emptyset$) are discarded at this stage to ensure that downstream processing operates only on videos with valid spoken content. We define the global transcription sets as: $S = \bigcup_{i=1}^N S_i, \quad T = \bigcup_{i=1}^N T_i. $

\

\subsection{Stage 2: Hierarchical Semantic Clip–Caption Pair Generation}
\label{subsec:stage2}

\noindent \textit{Hierarchical semantic segmentation:}  
We employ the GPT-4o large language model (LLM) to restructure sentence-level transcriptions \(S_i\) from Stage~1 into semantically coherent segments at three hierarchical levels of surgical granularity. \textcolor{black}{Inspired by the SAGES framework \citep{sages_consensus}, which represents procedural activities through progressively finer units of analysis, we adopt a  hierarchical approach in which segments reflect increasingly granular procedural goals. In our formulation, these are expressed as \textit{coarse}, \textit{mid}, and \textit{fine} segments, enabling a principled temporal decomposition of the surgical workflow that captures high-level context and fine-grained detail.}

For each video's transcription set \(S_i\), we prompt GPT-4o once (Fig.~\ref{fig:prompt_hierarchical} in Appendix) to merge temporally adjacent sentences according to semantic coherence, resulting in hierarchical segmentations:
\begin{equation}
    (C_i, M_i, F_i) = \mathrm{GPT{-}4o}_{\mathrm{seg}}(S_i),
\end{equation}
where,
\begin{itemize}
    \item \(C_i\) denotes the \textit{coarse}-level segments capturing high-level complex procedural structure,
    \item \(M_i\) denotes the \textit{mid}-level segments reflecting intermediate procedural goals,
    \item \(F_i\) denotes the \textit{fine}-level segments representing localized and more simple surgical tasks.
\end{itemize}

\noindent \textit{Video--text alignment:}  
Using hierarchical segments from the previous step, we create temporally aligned clip–caption pairs by matching segment boundaries with word-level timestamps. For each granularity level \textcolor{black}{\(g \in \{\mathrm{coarse}, \mathrm{mid}, \mathrm{fine}\}\)} in video \(v_i\), we extract captions and clips as:
\begin{equation}
x^g_{ij} = v_i\bigl[t^{\text{start},g}_{ij}, \, t^{\text{end},g}_{ij}\bigr]
\end{equation}
\begin{equation}
y^g_{ij} = \bigl\{ w_{ik} \in T_i \;\big|\; 
               t^{\text{start},g}_{ij} \le \tau^{\text{start}}_{ik} \;\wedge\; 
               \tau^{\text{end}}_{ik} \le t^{\text{end},g}_{ij} 
             \bigr\},
\end{equation}
where \(y^g_{ij}\) is the caption formed by concatenating all words within the segment, and \(x^g_{ij}\) is the corresponding video clip.

\noindent The outputs are three multi-granular datasets of clip-caption pairs:
\begin{equation}
\mathcal{D}_{g} = \left\{ \bigl(x^g_{ij}, y^g_{ij}\bigr) \right\}_{i,j}
\quad \text{for} \textcolor{black}{\quad g \in \{\mathrm{coarse}, \mathrm{mid}, \mathrm{fine}\}},
\end{equation}

\noindent This design offers two main benefits. First, it generates video–text pairs at multiple temporal scales capturing progressively finer procedural detail. Second, it naturally augments the dataset by generating multiple segments from the same video, substantially increasing the diversity and quantity of clip–caption pairs without additional data collection.

\

\subsection{Stage 3: Dual-Model Filtering.}  
\label{subsec:stage3}
\noindent We adopt a fine-grained filtering strategy that operates at the clip level, preserving valuable surgical segments rather than discarding entire videos through coarse filtering. This enables preservation of informative surgical segments even when other parts of the same video are irrelevant or noisy. Our goal is to filter out clip–caption pairs that are (i) visually non-surgical or (ii) textually non-descriptive, applying visual and textual checks before retaining a pair. \textcolor{black}{We describe our proposed filtering methodology for each modality in this section and detailed quantitative results in Sec. \ref{sec:pipeline_details_SM} from Appendix.} \\

\noindent \textit{Visual filtering:}  We apply visual filtering at the most granular level, $\mathcal{D}_{\textcolor{black}{\mathrm{fine}}}$ and propagate to corresponding \textcolor{black}{mid- and coarse-} level clips to maintain computational efficiency while ensuring hierarchical consistency. For each \textcolor{black}{fine-level} clip $x^{\textcolor{black}{\mathrm{fine}}}_{ij}$, we uniformly sample $N$ frames $\{f_{ij}^1, \dots, f_{ij}^N\}$ and classify them using a zero-shot SigLIP vision-language model~\citep{open_clip}. We design prompt templates (Fig.~\ref{fig:zero-shot-classification-prompts} in Appendix) to define the \emph{surgical} and \emph{non-surgical} classes. For each class, we compute its embedding as the average of multiple prompts, which we found empirically to produce more robust predictions than representing each class with a single prompt. Each frame $f_{ij}^n$ is assigned to the class whose averaged prompt embedding has the highest cosine similarity with the frame embedding.  

\noindent A \textcolor{black}{fine-level} clip $x^{\textcolor{black}{\mathrm{fine}}}_{ij}$ is classified as \emph{surgical} if more than $50\%$ of its sampled frames are labeled as surgical:
\begin{equation}
\frac{1}{N} \sum_{n=1}^{N} \mathbf{1}\left\{ \text{label}(f_{ij}^{n}) = \text{surgical} \right\} > 0.5
\end{equation}
Once all clips in $\mathcal{D}_{\textcolor{black}{\mathrm{fine}}}$ are classified, labels are propagated directly to their corresponding clips in $\mathcal{D}_{\textcolor{black}{\mathrm{mid}}}$ and $\mathcal{D}_{\textcolor{black}{\mathrm{coarse}}}$ to ensure hierarchical consistency. A higher-level clip inherits its label if all of its constituent fine segments share the same label. If labels are mixed, the higher-level clip is assigned the majority label among its fine segments. This direct propagation ensures consistent filtering throughout the hierarchy while preserving the fine-grained classification performed at the most granular level. \\

\noindent \textit{Text descriptiveness filtering:}  For textual filtering, each caption $y^g_{ij}$ is processed by $\text{GPT-4o}_{\mathrm{textfilter}}$, using a prompt (Fig. \ref{fig:prompt_validity} in Appendix) to instruct the model to determine whether the caption is describing a visual scene. The output is a binary label: \emph{descriptive} or \emph{non-descriptive}. \\

\noindent \textit{Final filtering criterion:}  A clip--caption pair $(x^g_{ij}, y^g_{ij})$ is retained if and only if it is surgical and descriptive. 

\subsection{Stage 4: Contextual Caption Enrichment}  
\label{subsec:stage4}

\noindent In the final stage, we enhance filtered captions by incorporating temporal context and metadata to improve semantic richness and coherence. For each clip--caption pair $(x^g_{ij}, y^g_{ij})$, we apply $\text{GPT-4o}_{\mathrm{enrich}}$ with a designed prompt (Fig. \ref{fig:prompt_enrichment} in Appendix) to instruct the model to generate an enriched caption $\hat{y}^g_{ij}$ conditioned on:  
(i) a fixed window of $N$ preceding captions at the same granularity level,  
(ii) the video title, and  
(iii) the surgical procedure type.

Formally, let  
\begin{equation}
\mathcal{H}^g_{ij} =\left \{ y^g_{i(j-N)}, y^g_{i(j-N+1)}, \dots, y^g_{i(j-1)} \right\}
\end{equation}
be the context window of preceding captions, and let $m_i$ denote the metadata for $v_i$. The enrichment function produces:  
\begin{equation}
\hat{y}^g_{ij} = \text{GPT-4o}_{\mathrm{enrich}}\big(y^g_{ij}, \mathcal{H}^g_{ij}, m_i\big),
\end{equation}
where $\hat{y}^g_{ij}$ integrates prior narrative context and procedural information for a more informative and coherent description.  

\noindent The output of Stage~4 is a set of enriched multi-granular datasets ${\mathcal{D}_{\mathrm{fine}}, \mathcal{D}_{\mathrm{mid}}, \mathcal{D}_{\mathrm{coarse}}}$, where all pairs depict surgical content, contain visually descriptive captions, and include an enhanced version of the caption enriched with contextual and procedural information.

\begin{figure}[h]
  \centering
    \includegraphics[width=0.5\textwidth]{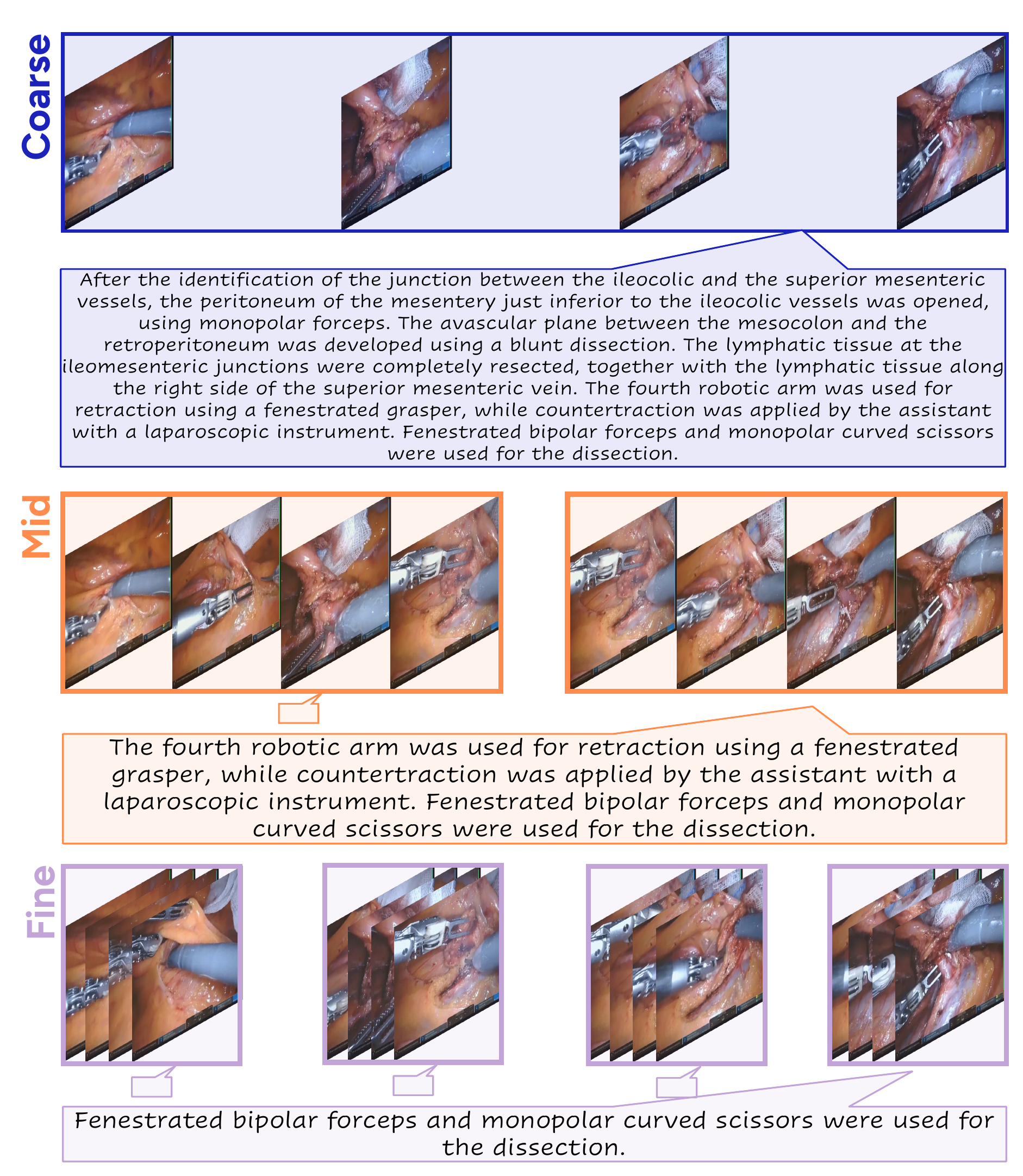}
    \caption{\textbf{Hierarchical clip-caption pair structure.} Example illustrating the three hierarchical levels of granularity in the dataset: (1) \textcolor{black}{\textit{Coarse-level}} pairs have long procedural coverage with complex context-rich descriptions, but lower temporal granularity; (2) \textcolor{black}{\textit{Mid-level}} pairs provide intermediate temporal granularity with moderate complexity captions; (3) \textcolor{black}{\textit{Fine-level}} pairs provide short-duration segments with high temporal resolution and action-focused descriptions}
    \label{fig:clip_caption_example}
\end{figure}

\subsection{SurgLaVi Dataset Overview} 
\label{subsec:overview}
\vspace{1.2mm}

\noindent \textbf{Video Source Collection.} To construct SurgLaVi-$\mathbf{\beta}$, we leveraged prior large-scale web scrapes of surgical YouTube videos provided by LEMON~\citep{che2025lemon} and GenSurgery~\citep{gensurg}, incorporating 3,110 and 4,172 videos, respectively, for a total of 6,812 unique videos. These videos originate from surgeons and institutions that publicly share operative recordings for educational purposes.
To further expand the scale, diversity, and domain coverage of the full dataset, SurgLaVi, we \textcolor{black}{additionally include  2,945 videos sourced from a surgical training archive that is part of an institutionally approved educational repository and adheres to ethical standards for surgical training. These materials do not contain identifiable patient information, and their use for AI research has been formally reviewed and authorized.} This brings the total video source size to 9,758 videos.

\vspace{1.2mm}

\begin{figure}[h]
  \centering
    \includegraphics[width=0.45\textwidth]{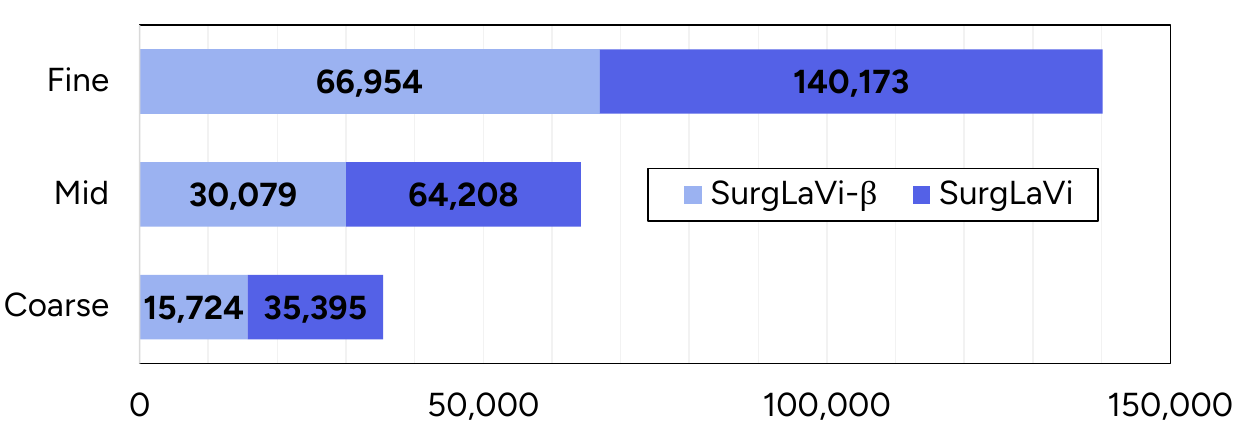}
    \caption{\textbf{Distribution of clip–caption pairs across hierarchical levels.} Histogram showing the number of pairs at the \textcolor{black}{coarse, mid, and fine} levels in  SurgLaVi and SurgLaVi-$\mathbf{\beta}$}.
    \label{fig:levels_counts}
\end{figure}

\noindent \textbf{Quantitative summary.} After processing the video collection through our pipeline, the final SurgLaVi dataset consists of 5,317 videos totaling 843 hours of surgical footage, sampled at varying frame rates from 10 to 60 fps (mean: 29 fps). The dataset contains 239,776 clip–caption pairs organized across three hierarchical granularities: coarse, mid, and fine level (see Fig.~\ref{fig:levels_counts} for the distribution across levels). After filtering, the coarse-level clips, the base level of our hierarchical structure, comprise 664.9 hours of video with 70M frames. SurgLaVi-$\mathbf{\beta}$ represents the open-source subset of SurgLaVi, containing 2,464 videos with 465 hours of surgical footage, sampled at 15 to 30 fps (mean: 28.75 fps), and 112,757 clip–caption pairs following the same hierarchical structure. The filtered coarse-level clips in SurgLaVi-$\mathbf{\beta}$ total 398 hours of video with 41M frames.

% Violin plots 

% Violin plots 
\begin{figure}[h]
  \centering
    \includegraphics[width=0.45\textwidth]{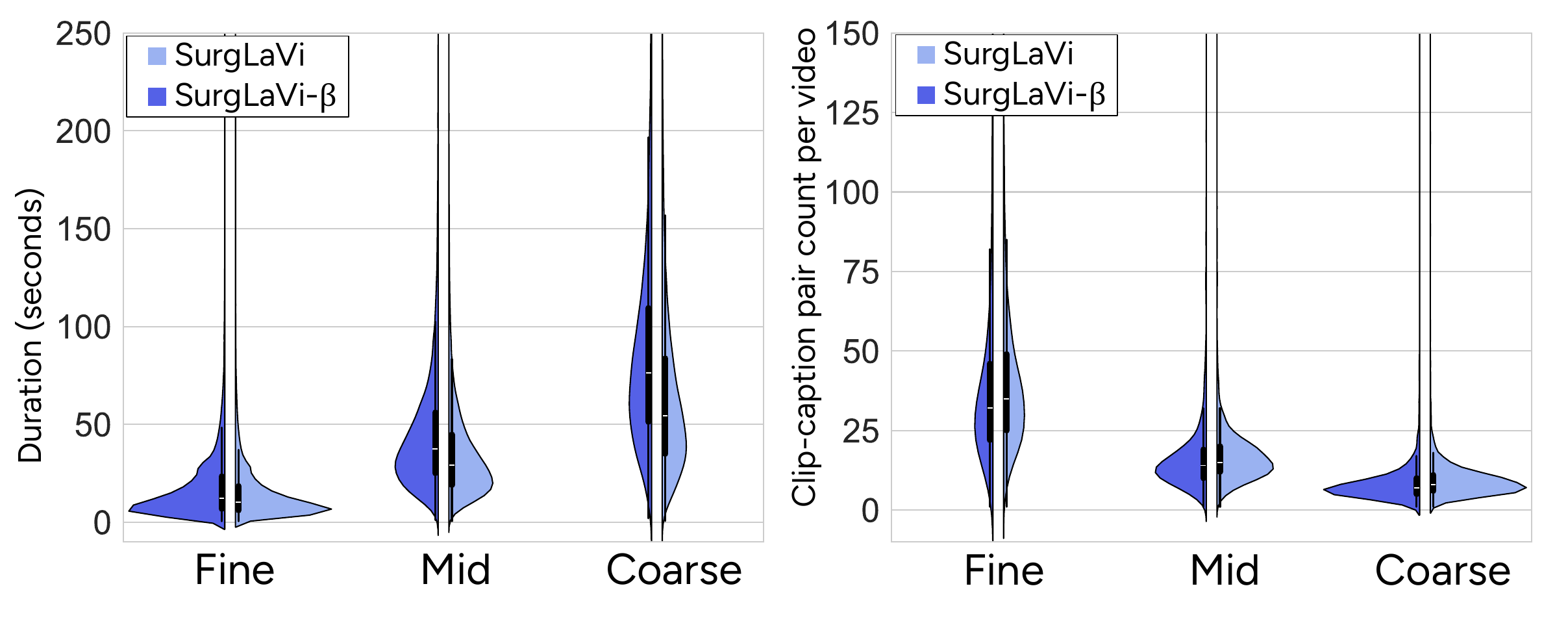}
    \caption{Violin plots comparing SurgLaVi$-\mathbf{\beta}$ and SurgLaVi datasets. \textbf{Left:} Distribution of mean clip durations at \textcolor{black}{coarse, mid, and fine levels}, showing increasing temporal coverage as granularity decreases in both datasets. \textbf{Right:} Distribution of clip counts per video at the same levels, with higher clip counts corresponding to finer granularity.}
    \label{fig:violin_plots}
\end{figure}

\vspace{1.2mm}
\noindent \textbf{Hierarchical structure illustration.}  
To qualitatively demonstrate the hierarchical organization of our dataset, Figure~\ref{fig:clip_caption_example} presents a portion of a representative video segmented into \textcolor{black}{coarse, mid, and fine} levels. Moving from coarse to fine level results in segments of progressively shorter duration and higher temporal resolution, while the corresponding captions evolve from long, context-rich descriptions to concise, action-focused statements. In the illustration, each clip is represented with the same number of sampled frames, highlighting how the effective sampling rate increases across levels: coarse clips provide broader coverage with lower temporal resolution, while fine clips capture fine-grained details with higher temporal resolution. This design explicitly encodes multiple temporal scales, enabling models trained on SurgLaVi to perform multi-scale temporal reasoning over surgical activities.

\vspace{1.2mm}
\noindent \textbf{Hierarchical level distribution.}  
Figure~\ref{fig:violin_plots} illustrates the distribution of clip durations (left) and the number of clips per video (right) across hierarchical levels. In both SurgLaVi and SurgLaVi-$\mathbf{\beta}$, coarse-level clips span longer intervals with relatively few segments per video, mid-level clips provide intermediate coverage, and fine-level clips are shorter but occur more frequently. Despite differences in video sources, the distributions of clip durations and clip counts per level remain consistent between the two datasets. Complementing these temporal statistics, Figure~\ref{fig:levels_counts} presents the overall distribution of clip–caption pairs across levels. As granularity increases from coarse to fine, the number of clips rises accordingly, a trend observed in both the full SurgLaVi dataset and its open-source subset, SurgLaVi-$\mathbf{\beta}$.

\begin{figure}[h]
  \centering
    \includegraphics[width=0.35\textwidth]{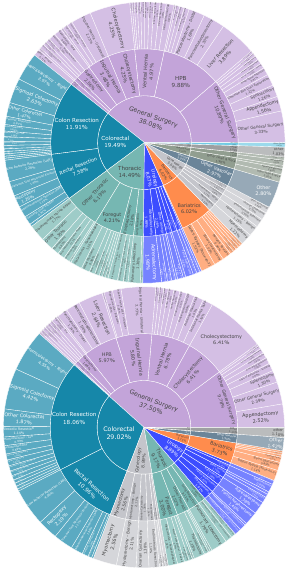}
\caption{\textcolor{black}{\textbf{Dataset distribution.} 
Sunburst visualizations of the SurgLaVi \textbf{(top)} and SurgLaVi$-\mathbf{\beta}$ \textbf{(bottom)} datasets, 
showing the hierarchical distribution of videos across surgical \textit{specialties} (center), 
\textit{subjects} (middle ring), and \textit{procedure types} (outer ring) as defined by the procedure taxonomy. 
Procedure-type classes representing at least $0.2\%$ of all videos are displayed.
\textit{Best viewed online.}}}
    \label{distribution}
\end{figure}

\begin{table}[h]
\centering
\caption{\textbf{Comparison of surgical video-caption datasets.} Total number of videos, clip--caption pairs (in thousands), average clip duration in seconds, and open-source availability. For SurgLaVi datasets, statistics are shown at three annotation granularities: Coarse (C), Mid (M), and Fine (F), with All (A) representing the aggregate across all granularities. Note: SurgLaVi$-\mathbf{\beta}$ is an open-source subset of the complete SurgLaVi dataset.}

\resizebox{0.48\textwidth}{!}{%
\begin{tabular}{lcccc}
\hline
\textbf{Dataset} & \textbf{Videos} & \textbf{\begin{tabular}[c]{@{}c@{}}Clip-Caption\\Pairs (K)\end{tabular}} & \textbf{\begin{tabular}[c]{@{}c@{}}Avg. Clip\\Duration (s)\end{tabular}} &  \textbf{\begin{tabular}[c]{@{}c@{}}Open -\\Source\end{tabular}}  \\
\hline

SVL  & 1,326 & 25.5 & 6.0 & \ding{55} \\
GenSurg+ & 1,800 & 17.0 & 45.0 & \ding{55} \\
\hline
\multirow{4}{*}{\textbf{SurgLaVi}} & \multirow{4}{*}{5,317} 
    & C: 35.4& C: 86.6& \multirow{4}{*}{\ding{55}} \\
    & & M: 64.2 & M: 45.1& \\
    & & F: 140.2& F: 18.2& \\
    \cline{3-4}
    & & \textbf{A: 239.8} & \textbf{A: 28.6} & \\
\hline
\multirow{4}{*}{\textbf{SurgLaVi$-\mathbf{\beta}$}} & \multirow{4}{*}{2,464}
    & C: 15.7& C: 65.4& \multirow{4}{*}{\ding{51}} \\
    & & M: 30.0 & M: 36.1& \\
    & & F: 66.9 & F: 15.2& \\
    \cline{3-4}
    & & \textbf{A: 112.6}& \textbf{A: 35.6} & \\
\hline
\end{tabular}}
\label{tab:dataset_description_comparison}
\end{table}

\vspace{1.2mm}
\noindent \textbf{Comparison with prior datasets.}
Table~\ref{tab:dataset_description_comparison} contrasts SurgLaVi and SurgLaVi$-\mathbf{\beta}$  with existing surgical video–caption datasets. Compared to SVL and GenSurg+, our datasets offer three key advantages: (i) \textit{Scale:} our full and open-source versions contain substantially more videos and clip–caption pairs; (ii) \textit{Hierarchical temporal segmentation:} prior datasets use clips of uniform duration, whereas ours provide clips of varying lengths to capture procedural context at multiple granularities \textcolor{black}{(coarse, mid, fine)};
(iii) \textit{Accessibility:} our dataset includes an openly available subset with captions, while all previous datasets are either entirely private (SVL) or have private captions (GenSurg+).

\vspace{1.2mm}
\noindent \textbf{Dataset diversity.}
Figure~\ref{distribution} depicts the distribution of videos in SurgLaVi and SurgLaVi$-\mathbf{\beta}$ across three taxonomy levels: surgical specialty, subject, and specific procedure type. The full dataset covers 10 specialties, 39 subjects, and 217 procedure types, while the open-source subset spans 9 specialties, 34 subjects, and 170 procedure types. For comparison, the GenSurg+ dataset includes only 28 procedure types (although the specific types differ from those in our taxonomy). We used an internal taxonomy designed by a team of expert surgeons (see Fig. \ref{taxonomy} from Appendix) to define a hierarchical classification scheme from specialty down to specific procedure type and assigned labels in a weakly supervised manner by providing GPT-4o with the taxonomy and each video’s transcription summary and prompting it to classify the video at all three levels (Fig. \ref{fig:prompt_classification} in Appendix).

\section{SurgCLIP}\label{sec:surgclip}

\noindent We establish a base model for our dataset using a dual-encoder video–language model trained with a symmetric contrastive objective. The video encoder is a ViT-B/16 backbone, used in the form of a TimeSFormer model~\citep{bertasius2021space} with divided space–time attention, which decouples spatial and temporal attention blocks. This design offers flexibility during inference, allowing the same model to operate on different temporal inputs, ranging from a single frame to multi-frame clips, without re-training. The text encoder is a BERT-base model~\citep{devlin2019bert} that processes the caption tokens. Both encoders project their modality-specific features into a shared embedding space via learned projection heads.

\begin{figure}[h]
  \centering
    \includegraphics[width=0.35\textwidth]{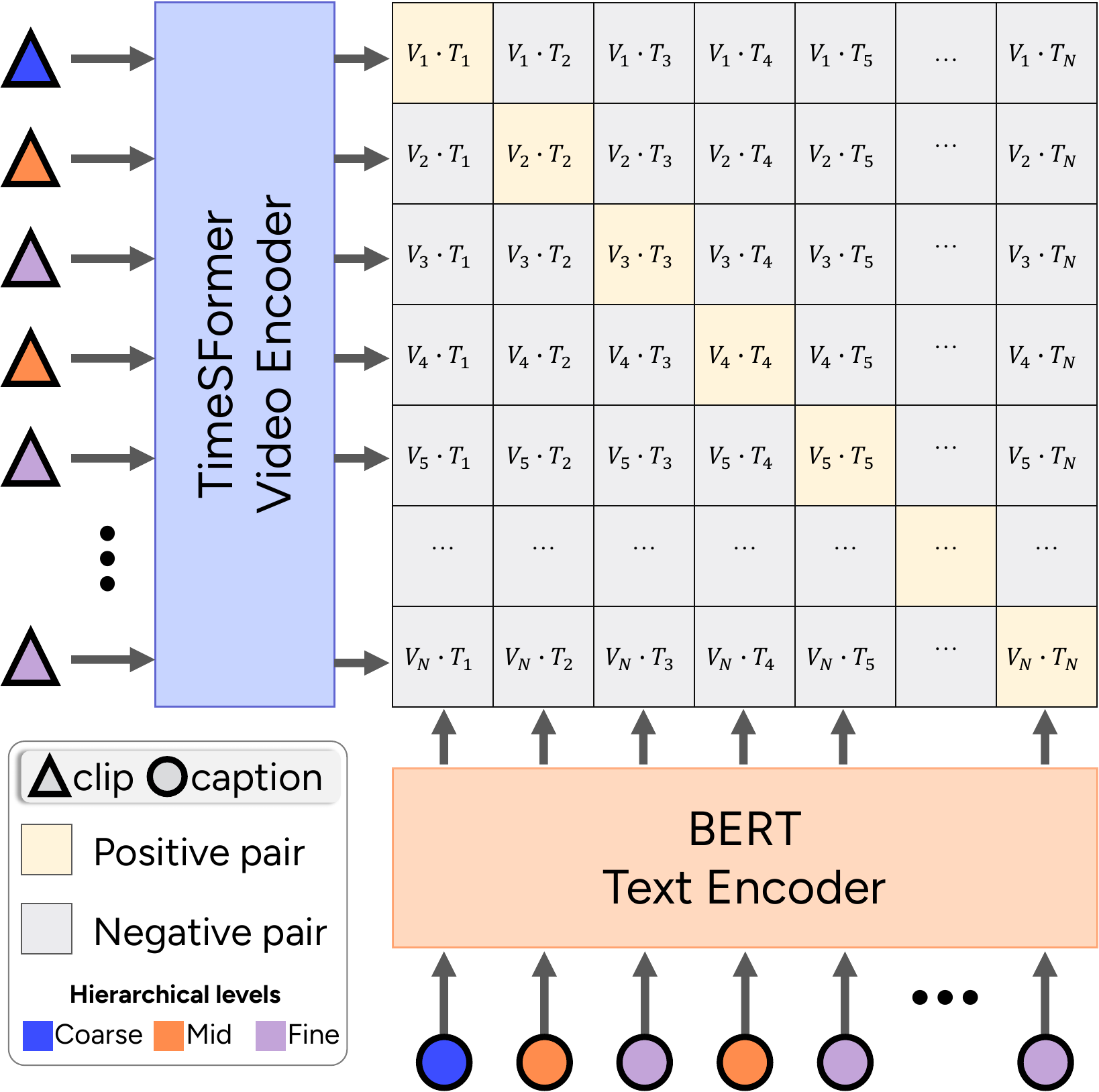}
    \caption{\textbf{SurgCLIP's multi-level contrastive learning for video-text representation learning.} The framework employs dual encoders, a video backbone and text backbone, that project video clips and captions into a shared embedding space. A contrastive cosine similarity matrix enables cross-modal alignment through InfoNCE loss. The model trains on \textcolor{black}{coarse, mid, and fine} clip-caption pairs within each batch, facilitating learning across different levels.}
\label{fig: arch}
\end{figure}

\noindent Given the hierarchical nature of our dataset, we randomly sample clip–caption pairs from all hierarchical levels \textcolor{black}{(coarse, mid, fine)} within each training batch, as shown in Figure \ref{fig: arch}. This strategy ensures that the InfoNCE objective is computed over a mixed-level pool of samples, enabling cross-level alignment and encouraging the model to learn representations transferable across temporal granularities. Specifically, for a batch $\mathcal{B} = \{ (v_i, t_i, \ell_i) \}_{i=1}^B$, where $v_i$ is a video clip, $t_i$ its text caption, and $\ell_i \in \textcolor{black}{\{\mathrm{coarse}, \mathrm{mid}, \mathrm{fine}\}}$ its granularity level, we encode them as: $z^v_i = f_v(v_i), \quad z^t_i = f_t(t_i)$, where $f_v$ and $f_t$ denote the video and text encoders, respectively. The video-to-text and text-to-video InfoNCE losses across all levels are:
\begin{equation}
\mathcal{L}_{v \rightarrow t} = -\frac{1}{B} \sum_{i=1}^B \log \frac{\exp\big( \mathrm{sim}(z^v_i, z^t_i) / \tau \big)}{\sum_{j=1}^B \exp\big( \mathrm{sim}(z^v_i, z^t_j) / \tau \big)},
\end{equation}
\begin{equation}
\mathcal{L}_{t \rightarrow v} = -\frac{1}{B} \sum_{i=1}^B \log \frac{\exp\big( \mathrm{sim}(z^t_i, z^v_i) / \tau \big)}{\sum_{j=1}^B \exp\big( \mathrm{sim}(z^t_i, z^v_j) / \tau \big)},
\end{equation}
where $\mathrm{sim}(\cdot,\cdot)$ denotes cosine similarity and $\tau$ is a learnable temperature parameter. The final training objective is the average of the two directions: $
\mathcal{L} = \frac{1}{2} \left( \mathcal{L}_{v \rightarrow t} + \mathcal{L}_{t \rightarrow v} \right).
$
% $\mathcal{L} = \frac{\mathcal{L}_{v \rightarrow t} + \mathcal{L}_{t \rightarrow v}}{2}.$

\noindent By computing similarities across all pairs in the mixed batch, we implicitly optimize the model for cross-level retrieval, making it sensitive to fine-grained and coarse-grained procedural semantics. To further support this objective, we apply a dynamic temporal sampling strategy that adapts the temporal resolution of clips to their hierarchical level. While the number of frames per clip remains fixed, the sampling stride varies with the clip’s duration. Fine-level clips, short segments focused on fine-grained surgical actions, are sampled at higher frame rates, providing high temporal resolution to capture subtle instrument motions and action-centric captions. In contrast, coarse-level clips cover longer temporal intervals and are sampled at lower frame rates, producing temporally broader representations aligned with procedural descriptions. This combination of mixed-level batching and dynamic temporal sampling exposes the model to multi-scale temporal structure, enabling it to capture fine-grained manipulations and broader procedural context within a unified pre-training framework.

\begin{table*}[h]
\centering
\caption{\textbf{Zero-shot performance comparison across multiple surgical benchmarks covering diverse procedure types, modalities, and task granularities.} Model evaluation on phase, step, action, and tool recognition, reporting accuracy (Acc), F1-score (F1), and mean average precision (mAP) metrics as applicable. SurgCLIP is trained on SurgLaVi dataset, while SurgCLIP$_{(\mathbf{\beta})}$ is trained on the open-source SurgLaVi$-\mathbf{\beta}$ subset. Best results are in bold, second-best are underlined.}
\resizebox{0.95\textwidth}{!}{
\begin{tabular}{@{}cccccccc|cc@{}}
\toprule
Recognition Task        & Dataset                        & Metric               & CLIP          & MedSigLIP & SurgVLP     & HecVL          & PeskaVLP    & SurgCLIP$_{(\mathbf{\beta})}$ & SurgCLIP       \\ \midrule
\multirow{12}{*}{Phase} & \multirow{2}{*}{Cholec80}      & Acc                  & 27.81$_{10.4}$         & 40.94$_{\pm1.14}$     & 49.02$_{\pm10.6}$       & 47.50$_{\pm11.0}$           & 51.43$_{\pm11.01}$       & \underline{57.98}$_{\pm5.25}$ & \textbf{61.29}$_{\pm10.7}$ \\ \cmidrule(l){3-10} 
&                                & F1                   & 8.42$_{\pm2.81}$          & 18.74$_{\pm4.42}$     & 32.46$_{\pm6.43}$       & 23.89$_{\pm5.53}$          & 39.39$_{\pm7.43}$       & \underline{39.42}$_{\pm5.25}$ & \textbf{50.53}$_{\pm8.83}$ \\ \cmidrule(l){2-10} 
& \multirow{2}{*}{AutoLaparo}    & Acc                  & 8.02$_{\pm3.43}$          & 26.54$_{\pm5.39}$     & 10.02$_{\pm5.23}$       & 38.72$_{\pm7.01}$          & 34.94$_{\pm9.06}$       & \underline{55.72}$_{\pm7.65}$ & \textbf{69.14}$_{\pm5.98}$ \\ \cmidrule(l){3-10} 
&                                & F1                   & 4.79$_{\pm1.77}$          & 10.17$_{\pm3.24}$     & 7.19$_{\pm4.15}$        & 25.81$_{\pm4.91}$          & 28.41$_{\pm7.79}$       & \underline{45.95}$_{\pm8.10}$ & \textbf{56.37}$_{\pm4.54}$ \\ \cmidrule(l){2-10} 
& \multirow{2}{*}{StrasBypass70} & Acc                  & 18.52$_{\pm5.99}$         & 20.15$_{\pm3.98}$     & 26.08$_{\pm6.30}$       & 28.59$_{\pm8.52}$          & 29.42$_{\pm6.92}$       & \underline{31.24}$_{\pm6.88}$ & \textbf{32.37}$_{\pm6.30}$ \\ \cmidrule(l){3-10} 
&                                & F1                   & 3.51$_{\pm1.12}$          & 5.54$_{\pm1.15}$      & 19.18$_{\pm4.30}$       & 23.91$_{\pm6.60}$          & 19.64$_{\pm3.97}$       & \underline{26.05}$_{\pm4.39}$ & \textbf{30.78}$_{\pm5.37}$ \\ \cmidrule(l){2-10} 
& \multirow{2}{*}{\textcolor{black}{Heichole}}      & Acc                  & 19.21$_{\pm5.27}$         & 45.36$_{\pm10.7}$     & 45.70$_{\pm10.7}$        & 29.91$_{\pm12.9}$          & 53.33$_{\pm8.47}$       & \underline{56.95}$_{\pm17.59}$ & \textbf{63.84}$_{\pm15.4}$ \\ \cmidrule(l){3-10} 
&                                & F1                   & 6.17$_{\pm2.43}$          & 17.63$_{\pm3.24}$     & 27.04$_{\pm3.92}$       & 22.76$_{\pm7.21}$          & 36.46$_{\pm4.43}$       & \underline{44.00}$_{\pm7.68}$ & \textbf{55.15}$_{\pm8.9}$ \\ \cmidrule(l){2-10} 
& \multirow{2}{*}{BernBypass70}  & Acc                  & 20.73$_{\pm9.14}$         & 30.73$_{\pm8.13}$     & 30.46$_{\pm6.55}$       & \textbf{32.09}$_{\pm8.41}$ & \underline{31.48}$_{\pm8.60}$ & 18.30$_{\pm6.68}$        & 23.90$_{\pm6.58}$           \\ \cmidrule(l){3-10} 
&                                & F1                   & 4.17$_{\pm1.23}$          & 6.14$_{\pm1.34}$      & 17.48$_{\pm2.93}$       & \underline{18.58}$_{\pm5.00}$    & 17.47$_{\pm5.25}$       & 15.06$_{\pm5.54}$       & \textbf{19.68}$_{\pm5.78}$ \\ \cmidrule(l){2-10} 
& \multirow{2}{*}{GraSP}         & Acc                  & 9.93$_{\pm4.79}$          & 12.88$_{\pm7.20}$     & 11.79$_{\pm2.89}$       & 10.99$_{\pm2.99}$          & 13.85$_{\pm3.45}$       & \underline{34.77}$_{\pm8.15}$ & \textbf{41.49}$_{\pm7.87}$ \\ \cmidrule(l){3-10} 
&                                & F1                   & 2.06$_{\pm0.57}$          & 4.84$_{\pm2.30}$      & 6.07$_{\pm1.71}$        & 6.46$_{\pm2.87}$           & 9.86$_{\pm3.14}$        & \underline{27.98}$_{\pm6.53}$ & \textbf{34.94}$_{\pm7.61}$ \\ \midrule
\multirow{2}{*}{Step}   & \multirow{2}{*}{GraSP}         & Acc                  & 3.85$_{\pm2.72}$          & 11.53$_{\pm6.63}$     & 6.02$_{\pm1.11}$        & 1.96$_{\pm1.54}$           & 3.65$_{\pm1.85}$        & \underline{14.15}$_{\pm3.92}$ & \textbf{26.28}$_{\pm7.29}$ \\ \cmidrule(l){3-10} 
&                                & F1                   & 0.83$_{\pm0.61}$          & 1.86$_{\pm1.01}$      & 3.22$_{\pm0.25}$        & 0.56$_{\pm0.38}$           & 2.47$_{\pm0.80}$        & \underline{11.14}$_{\pm4.16}$ & \textbf{16.53}$_{\pm5.44}$ \\ \midrule
\multirow{2}{*}{Action} & \multirow{2}{*}{SARRARP50}     & Acc                  & \textbf{28.8}$_{\pm3.85}$ & 5.49$_{\pm3.24}$      & \underline{17.91}$_{\pm2.86}$ & 4.67$_{\pm3.09}$           & 5.18$_{\pm3.95}$        & 13.94$_{\pm8.18}$       & 17.42$_{\pm7.08}$          \\ \cmidrule(l){3-10} 
&                                & F1                   & 6.28$_{\pm1.14}$          & 3.17$_{\pm1.75}$      & 7.03$_{\pm1.99}$        & 3.03$_{\pm2.43}$           & 2.68$_{\pm1.87}$        & \underline{7.62}$_{\pm2.91}$  & \textbf{7.76}$_{\pm2.71}$  \\ \midrule
Triplet                 & CholeT50                       & mAP                  & 2.50           & 2.98      & 3.04        & 3.63           & \underline{4.49}  & 4.17        & \textbf{5.28}  \\ \midrule
\multirow{3}{*}{Tool}   & Cholec80                       & \multirow{3}{*}{mAP} & 18.44         & 17.92     & 31.19       & 25.11          & \underline{38.88} & 36.77       & \textbf{40.80} \\ \cmidrule(lr){2-2} \cmidrule(l){4-10} 
& \textcolor{black}{Heichole}                       &                      & 21.18         & 20.58     & 22.46       & 19.36          & \underline{32.82} & 31.47       & \textbf{36.79} \\ \cmidrule(lr){2-2} \cmidrule(l){4-10} 
& GraSP                          &                      & 36.00            & 37.18     & 36.93       & 36.75          & 41.05      & \underline{43.06} & \textbf{45.97} \\ \bottomrule
\end{tabular}
}
\label{tab:zero_shot}
\end{table*}

\noindent For detailed implementation details about training schedules, hyperparameters and computational resources used to train SurgCLIP, refer to Sec. \ref{sec:surgclip_training_details} in Appendix.

\section{Experimental Validation}\label{experimental_validation}

\subsection{Zero-Shot Classification}
\label{sec:zero-shot}

\vspace{1.2mm}
\noindent \textbf{Experimental setting and benchmarks.} We evaluate generalization and transferability in a zero-shot setting across laparoscopic and robotic modalities, multiple procedure types, and a wide range of temporal granularities. \textcolor{black}{This includes recognition levels that can match our pre-training as well as benchmarks that operate at even more granular level, such as action, triplet, and tool recognition. Evaluating these additional levels enables us to measure how well the model handles tasks whose temporal granularity exceeds supervision during pretraining}. For phase recognition, we cover five laparoscopic benchmarks: Cholec80~\citep{endonet} and HeiChole~\citep{wagner2023comparative} for cholecystectomy, AutoLaparo~\citep{wang2022autolaparo} for hysterectomy, and StrasByPass and BernByPass~\citep{multibypass} for gastric bypass—as well as the robotic GraSP benchmark~\citep{ayobi2025pixel} for robotic prostatectomy. For step recognition, we use GraSP~\citep{ayobi2025pixel}. Action recognition is evaluated on SARRARP50~\citep{sarrarp50}, which focuses on robotic prostatectomy. Triplet recognition (instrument–verb–target) is assessed using CholecT50~\citep{nwoye2022rendezvous}, and tool presence recognition is evaluated on Cholec80~\citep{endonet},  GraSP~\citep{ayobi2025pixel}, and Heichole~\citep{wagner2023comparative}.

\vspace{1.2mm}

\noindent \textbf{Inference setting.}
All models follow a consistent zero-shot protocol. The frozen vision backbone encodes a 16-frame window centered on each evaluation frame to model temporal dynamics in surgical videos. Then, we mean pool features into a single embedding and compare via cosine similarity to all candidate class embeddings. \textcolor{black}{For phase, step, and action recognition, the predicted label corresponds to the class with the highest similarity score. We evaluate performance using video-wise Accuracy and video-wise macro-averaged F1 as proposed in \citep{funke2023metrics} and report standard deviation across videos. For the video-wise macro-averaged F1, we first compute the F1-score for each class within a procedure and then average these scores uniformly across classes. Next, we average the resulting per-procedure F1-scores across all procedures to avoid bias toward longer videos and to better capture variability between cases. For simplicity, we refer to these metrics as F1 and Accuracy throughout the remainder of this work.} For tool presence and CholecT50 triplets, we treat similarity scores as class probabilities and report mAP. We provide inference prompts specific for each dataset and task in the Appendix.

\vspace{1.2mm}
\noindent \textbf{Results.}
Table~\ref{tab:zero_shot} shows that our approach consistently outperforms prior work across nearly all datasets and tasks. Notably, SurgCLIP pre-trained on SurgLaVi-$\beta$, a smaller subset, already surpasses baselines on most benchmarks, while pre-training on the full SurgLaVi results in further gains. This pattern indicates that our automated curation pipeline produces high-quality supervision from open-source videos, and that additional scale, diversity, and hierarchical structure provide incremental benefits and  robust zero-shot transfer across procedures, modalities, and task granularities. The largest gains occur in high-level temporal tasks such as phase recognition, with improvements of up to +10, +28, and +25 F1 over the strongest prior methods on Cholec80, AutoLaparo, and GraSP, respectively. While action recognition remains particularly challenging, SurgCLIP matches or exceeds previous VLP approaches and delivers consistent gains in tool recognition, indicating that its learned representations extend beyond workflow understanding to object-centric tasks. A remaining limitation, however, is that models primarily rely on global visual features during pre-training and inference, which limits their capacity to capture the fine spatial details and subtle instrument motions necessary for precise action recognition.

\subsection{Linear Probing and Context Optimization}
\label{sec:linear-probing}
\begin{figure}[h]
  \centering
    \includegraphics[width=0.48\textwidth]{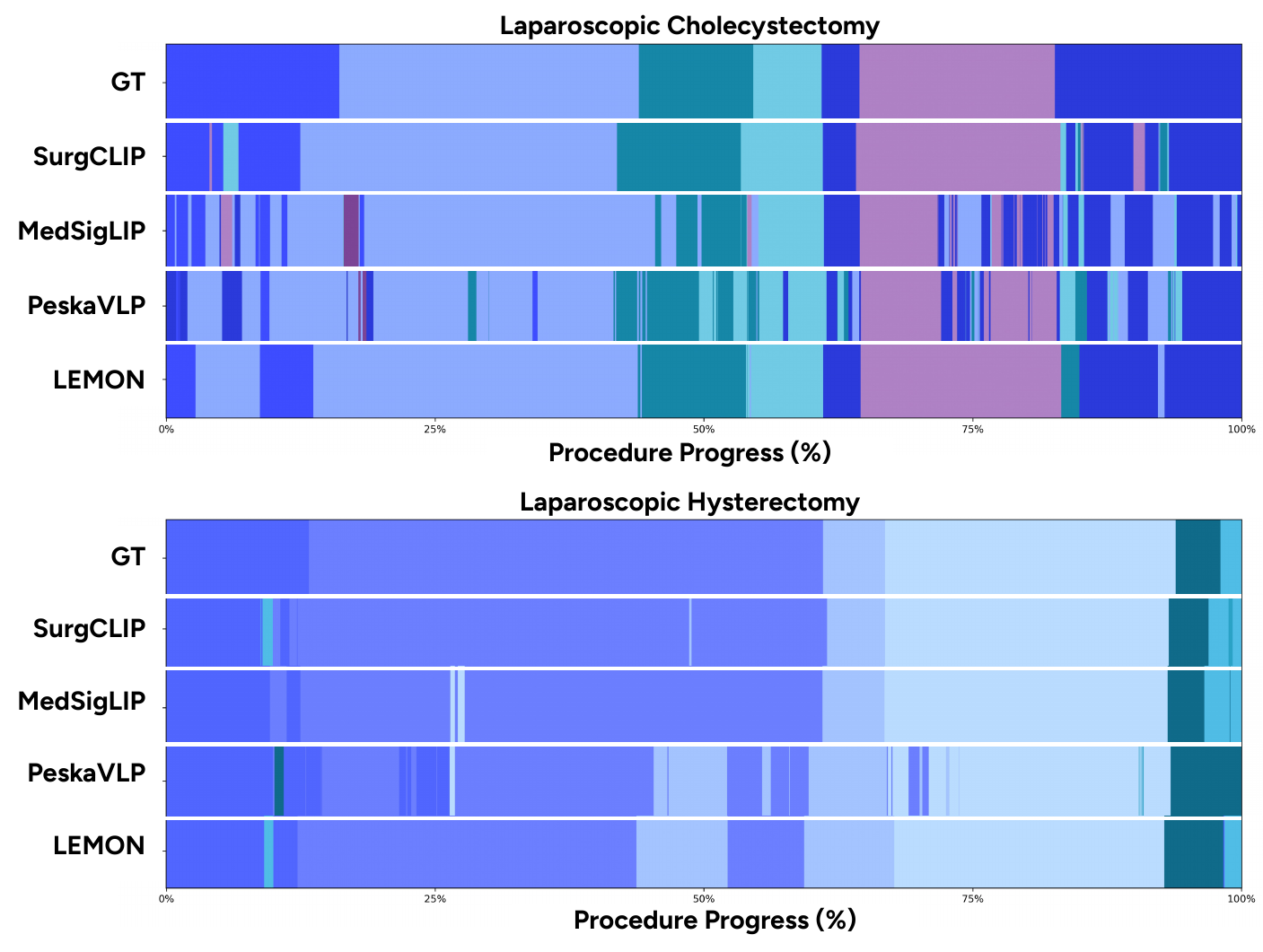}
    \caption{\textbf{Qualitative comparison of surgical phase recognition via linear probing.} Ground truth and predicted phase sequences from different state-of-the-art models and our approach (SurgCLIP) for representative videos from laparoscopic cholecystectomy (Cholec80) and hysterectomy (AutoLaparo) procedures. Colors indicate different surgical phases, with the x-axis showing procedure progress (0-100\%). Our approach achieves superior temporal consistency and phase boundary accuracy across diverse surgical procedures compared to existing methods.}
    \label{fig:qualitative_phase_recognition}
\end{figure}

\begin{figure*}[h]
  \centering
  \includegraphics[width=1\textwidth]{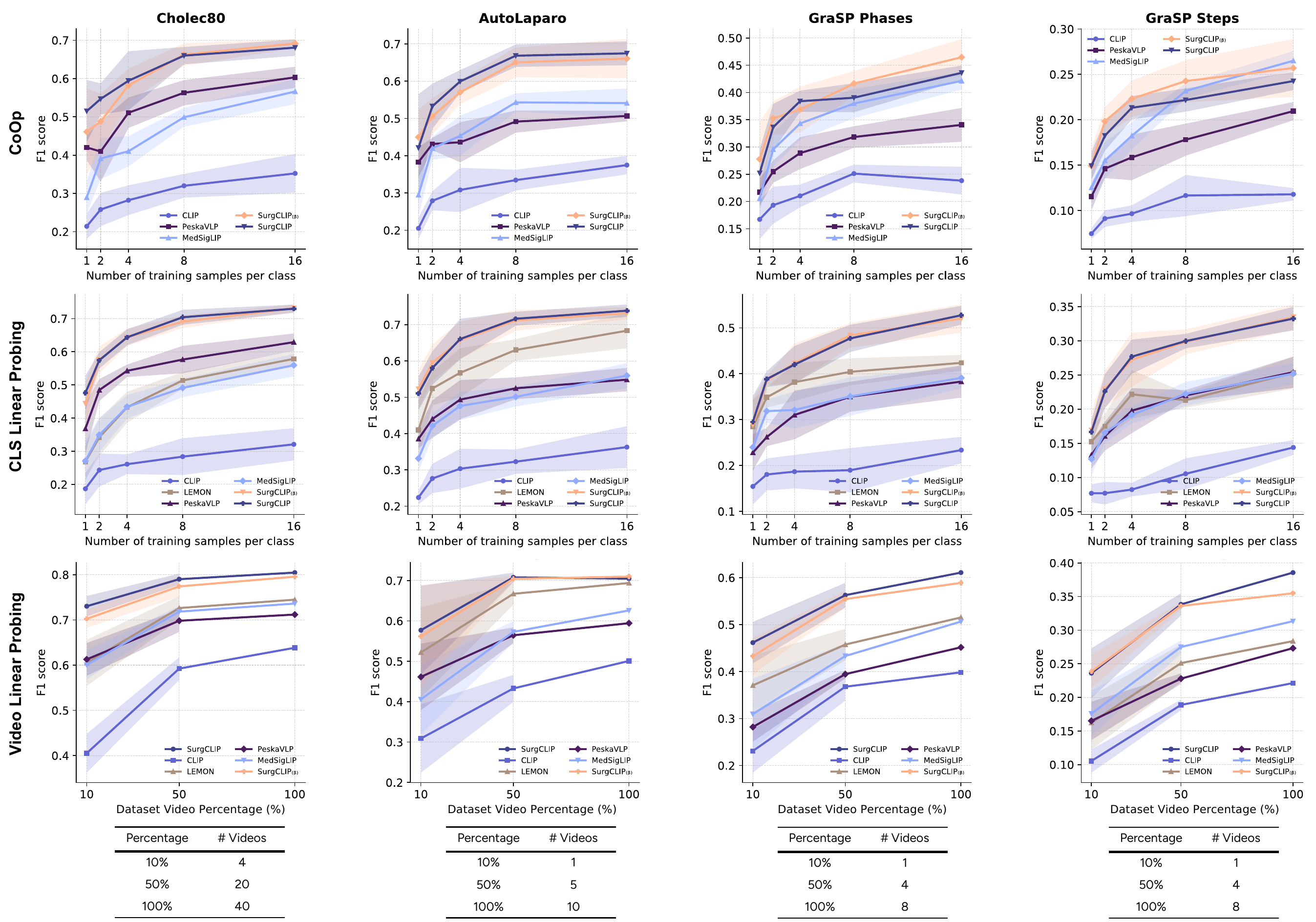}
  \caption{
    \textcolor{black}{\textbf{Comparison of CoOp, CLS Few-Shot, and Video Few-/Full-Shot Linear Probing.}
   F1-scores evaluated under three paradigms: \textbf{(top)} Context Optimization (CoOp),
    \textbf{(middle)} CLS Few-Shot Linear Probing, and \textbf{(bottom)} Video Few-/Full-Shot Linear Probing.
    CoOp and CLS Linear Probing curves are plotted over the number of labeled training examples per class \{1, 2, 4, 8, 16\}.
    Video Linear Probing reports performance using \{10\%, 50\%, 100\%\} of the available labeled training videos in each dataset.
    Compared models include SurgCLIP, CLIP, MedSigLIP, PeskaVLP, and LEMON. Bottom tables report the number of training videos corresponding to each percentage for each dataset.
    (P) and (S) denote phase and step recognition, respectively, and $(\beta)$ marks models pre-trained on SurgLaVi-$\beta$. }
  }
  \label{fig:linear_probing}
\end{figure*}

% Figure 2
\begin{figure}[h]
  \centering
    \includegraphics[width=0.35\textwidth]{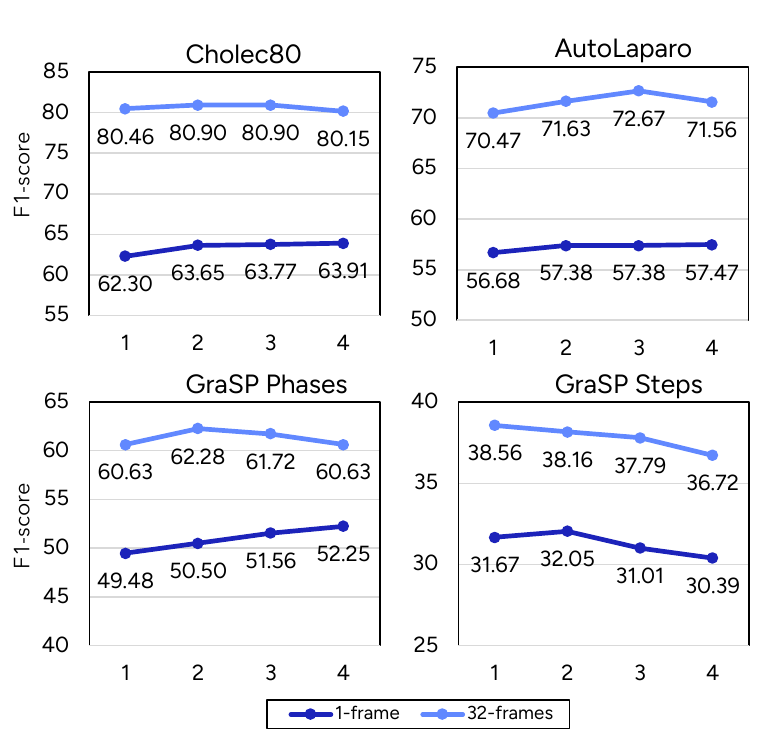}
    \caption{Effect of increasing MLP depth before the linear classifier on downstream performance. For 1-frame embeddings, performance generally increases with more layers, suggesting limited linear separability. For 32-frame embeddings, small gains appear up to two (or three) layers before performance plateaus/declines, indicating that temporal aggregation yields more linearly separable representations where additional classifier capacity offers little benefit.}
    \label{fig:layers}
\end{figure}

% Figure 3
\begin{figure}[]
  \centering
    \includegraphics[width=0.35\textwidth]{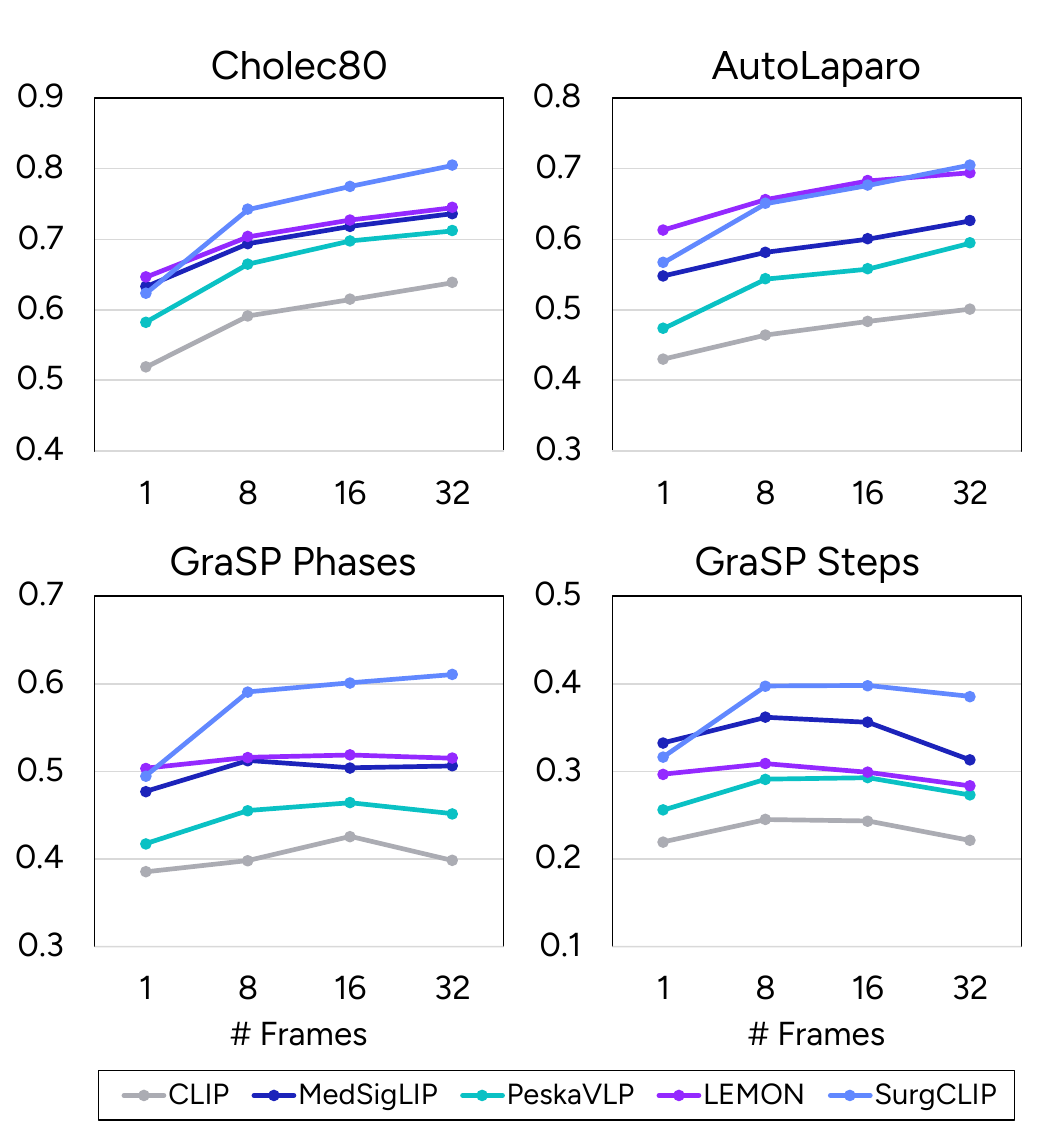}
    \caption{Temporal window size effect on linear probing performance (F1-score). For phase recognition, performance consistently improves with more frames, underscoring the importance of temporal context in workflow analysis. For step recognition, a more granular task, performance declines beyond 8 frames.}
    \label{fig:n_frames}
\end{figure}

\noindent \textbf{Experimental setting and benchmarks.}
To evaluate the quality and transferability of the learned representations, we conduct Linear Probing and \textcolor{black}{Context Optimization (CoOp)~\citep{zhou2022learning}}. On the one hand, linear probing assesses the discriminative capabilities of vision-only representations by training a single linear classifier on frozen visual embeddings from SurgCLIP’s video backbone. On the other hand, \textcolor{black}{CoOp probes multimodal alignment by learning a set of continuous context tokens that condition the text encoder, enabling us to assess how effectively each model’s visual and textual representations interact when guided by task-specific prompts. Together, these two evaluations provide a comprehensive view of representation quality in unimodal and multimodal settings.}

\noindent We conduct these studies on three downstream benchmarks: Cholec80 and AutoLaparo for surgical phase recognition, and GraSP for phase and step recognition. Consistent with our zero-shot analysis, we compare SurgCLIP against CLIP, MedSigLIP, and PeskaVLP. Additionally, we include LEMON~\citep{che2025lemon}, a surgical foundation model, to compare our approach against in-domain, large-scale, self-supervised, purely visual pre-training strategies. All models are evaluated using frozen features and identical inference settings to ensure strict comparability.

\subsubsection{Linear Probing}
\label{subsubsec:linear-prob}
\noindent We consider two complementary few-shot paradigms that reflect different low-effort labeling scenarios: 

\noindent \textcolor{black}{\textbf{(1) Video Few- and Full-Shot Linear Probing.} This paradigm follows prior surgical VLP work~\citep{peskavlp, surgvlp, hecvl}. Here, N-shot denotes the percentage of labeled procedure videos (i.e., full video procedures) sampled from the training split. Once the video of a patient is selected, all frames from that procedure are used to train the linear classifier. We evaluate 10\%, 50\%, and 100\% regimes, where the latter corresponds to full-shot linear probing. This setting reflects a practical annotation scenario in which only a limited number of patient video recordings can be labeled, and annotations are provided at phase/step transition times, which allows all frames within each segment to automatically inherit the corresponding label.}

\noindent \textcolor{black}{ \textbf{(2) CLS Few-Shot Linear Probing: } Inspired by few-shot evaluation protocols in general computer vision~\citep{zhang2022tip, zhou2022learning}, we introduce a second paradigm in which few-shot supervision is defined at the frame level rather than the procedure level. Unlike the video-based regime, where selecting a video provides all frames from that procedure, this setting imposes a stricter constraint by limiting the actual number of annotated samples, not the number of procedures. This also increases training diversity, as sampled frames may originate from multiple surgeries, reflecting a scenario where many patient videos are available but only a small number of frames per class can be annotated. In this setting, we sample 1, 2, 4, 8, and 16 labeled samples per class from the full training split.}

\noindent \textbf{Setting.} All models are trained and evaluated using mean-pooled embeddings from a 32-frame temporal window, following our analysis of temporal context (Fig.~\ref{fig:linear_probing}). For each labeled frame, we construct this window by gathering 32 unlabeled neighboring frames around it. This preserves the supervision at a single annotated frame while allowing the model to leverage local temporal structure. For all few-shot studies, we generate five random subsets of the training data without replacement and report the mean and standard deviation across runs, ensuring that identical subsets are used across all models for strict comparability. More implementation details available in Appendix, Sec. \ref{sec:linear_probing_details}.

\subsubsection{Context Optimization (CoOp)}
\label{subsubsec:coop}
\noindent \textcolor{black}{For multimodal evaluation, we apply Context Optimization (CoOp)~\citep{zhou2022learning} using a single shared learnable context across all classes. Instead of hand-crafted prompts, each class is represented only by its class name, and the learnable context tokens are prepended to this name to form the final text embedding. The vision and text encoders remain frozen, ensuring that CoOp measures the adaptability of the prompt space rather than the encoders themselves.}

\noindent \textcolor{black}{We follow CoOp’s few-shot training protocol, aligned with our CLS Few-Shot Linear Probing setup. Specifically, we sample 1, 4, 8, and 16 labeled frames per class from the training split and optimize only the context tokens. As with linear probing, we construct five random subsets for each shot level and report the mean and standard deviation across runs. Additional implementation details in Appendix, Sec. \ref{sec:linear_probing_details}}.

\vspace{1.2mm}

\noindent \textcolor{black}{\textbf{Results.} As shown in Figure~\ref{fig:linear_probing}, 
SurgCLIP consistently outperforms all baselines across CoOp, CLS Few-Shot, and Video Few-/Full-Shot settings for both SurgLaVi and SurgLaVi-$\beta$. This advantage holds across nearly all datasets and supervision regimes, with the only exception occurring in GraSP step recognition under CoOp. We attribute this performance to three design choices: (1) a video backbone that explicitly models temporal dynamics, (2) language supervision that aligns features with procedural semantics, and (3) hierarchical pre-training with dynamic temporal sampling, which exposes the model to multi-scale surgical activity. Together, these components produce semantically enriched representations that remain highly discriminative even when only a linear layer is trained on top.}

\noindent \textcolor{black}{The results also reveal several consistent trends. First, CoOp improves upon zero-shot inference when more than two labeled samples are available, demonstrating the benefit of learnable contextual prompts over hand-crafted text embeddings; however, it generally remains less effective than linear probing on the visual features. Secondly, using only 16 labeled frames per class often outperforms training on all frames from a single labeled video, and in several cases (AutoLaparo–50\%, GraSP–50\%, Cholec80–10\%) approaches the performance of labeling 4--5 full procedures. This highlights the practical value of sparse, diverse annotations over dense single-procedure labeling. Third, the relative performance of competing models varies across datasets: on Cholec80, PeskaVLP is the strongest non-SurgCLIP baseline, while on AutoLaparo and GraSP, LEMON generally performs best among non-SurgCLIP models. Interestingly, MedSigLIP, despite being trained on broad medical data rather than surgical data, often matches or exceeds PeskaVLP, underscoring the benefit of large-scale medical image–text supervision. Conversely, CLIP is consistently the lowest-performing model, reflecting the substantial domain gap between natural images and surgical videos despite its massive pre-training scale. }

\vspace{1.2mm}

\noindent \textbf{Qualitative analysis.} Figure~\ref{fig:qualitative_phase_recognition} shows comparative qualitative phase recognition results for representative procedures under full-shot linear probing. Overall, SurgCLIP produces more temporally consistent predictions and more accurate phase boundaries compared to existing models. Similar improvements are also observed with LEMON, which benefits from large-scale self-supervised pre-training. In contrast, PeskaVLP and MedSigLIP produce noisier outputs with frequent phase switches, resulting in inconsistent predictions over time.

\vspace{1.2mm}

\noindent \textbf{Scaling number of frames.} We study the effect of varying the temporal window size (1, 8, 16, 32 frames) across all benchmarks under full-shot linear probing (Fig.~\ref{fig:n_frames}). Performance generally improves with more frames, highlighting the importance of temporal context in workflow recognition. The only exception is GraSP step recognition, where performance peaks at 8 frames, which is consistent with the fine-grained, short-duration nature of this task compared to phase recognition. With single-frame inputs, LEMON and MedSigLIP sometimes perform better than SurgCLIP, likely due to LEMON’s frame-level self-supervised training and MedSigLIP’s higher image input resolution. However, once temporal windows are introduced, SurgCLIP exhibits superior gains, demonstrating its ability to learn robust spatio-temporal representations.

\vspace{1.2mm}

\noindent \textbf{Scaling number of layers.} We further examine how classifier depth interacts with embeddings built from different temporal windows  under full-shot linear probing (Fig.~\ref{fig:layers}). For single frame embeddings, performance generally increases as layers are added (except for GraSP Steps, where it declines after one layer) suggesting that these representations are not fully linearly separable and benefit modestly from additional nonlinear capacity. In contrast, embeddings constructed from a 32-frame temporal window improve only up to two layers (three for AutoLaparo) before plateauing or decreasing, consistent with the idea that temporal aggregation around the frame of interest produces more linearly separable representations, where deeper classifiers provide little benefit and may even harm performance.

\subsection{Temporal probing with MS-TCN}\label{subsec:temporal_probing}

\noindent We further evaluate our learned features in a temporal probing setting using a Multi-Stage Temporal Convolutional Network (MS-TCN) head~\citep{farha2019ms}, which models long-range dependencies across full videos. This experiment complements linear probing by testing whether frozen backbone features can support a temporal head capable of capturing procedure-level dynamics. Because temporal modeling requires not only frame-level classification but also coherent and well-aligned segment boundaries, we assess both frame-level metrics (video-wise Accuracy and macro-video-wise F1) and temporal segmentation metrics (Segmental F1@$\{10, 25, 50\}$), following prior work in temporal segmentation~\citep{lea2017temporal, zhang2023surgical}. We conduct this analysis in the full-shot regime across the three surgical phase recognition benchmarks. MS-TCN is trained on frame-level features extracted from each backbone, ensuring a consistent comparison across models. Additional details are provided in Section~\ref{sec:temporal_details} in the Appendix.

\begin{table}[]
\centering
\caption{\textbf{Temporal probing results using MS-TCN head.} Accuracy, F1, and Segmental F1@10 scores for all models in the full-shot regime across three surgical workflow recognition benchmarks. Here, SurgCLIP$_{(\mathbf{\beta})}$ denotes the variant pre-trained on SurgLaVi-$\beta$.}
\resizebox{0.48\textwidth}{!}{
\begin{tabular}{@{}lcccc|cc@{}}
\toprule
\multirow{2}{*}{Metric} & \multicolumn{6}{c}{Model} \\ 
\cmidrule(lr){2-7}
                        & CLIP & MedSigLIP & PeskaVLP & LEMON & SurgCLIP$_{(\mathbf{\beta})}$ & SurgCLIP \\ 
\midrule
\multicolumn{7}{l}{\textbf{Cholec80}} \\
\midrule
Acc & 77.24$_{\pm10.82}$ & 80.88$_{\pm12.02}$ & 85.71$_{\pm11.10}$ & 85.47$_{\pm11.66}$ & 89.72$_{\pm7.79}$ & \textbf{90.11}$_{\pm7.34}$ \\
F1 & 61.95$_{\pm13.26}$ & 69.09$_{\pm13.66}$ & 78.07$_{\pm11.15}$ & 75.77$_{\pm12.40}$ & 82.39$_{\pm9.57}$ & \textbf{83.35}$_{\pm8.84}$ \\
\textcolor{black}{F1@10} & 30.12$_{\pm13.96}$ & 56.97$_{\pm23.49}$ & 54.02$_{\pm22.08}$ & 65.38$_{\pm22.71}$ & \textbf{86.22}$_{\pm17.15}$ & 84.49$_{\pm14.49}$ \\
\textcolor{black}{F1@25} & 27.43$_{\pm13.88}$ & 53.49$_{\pm23.48}$ & 52.76$_{\pm22.67}$ & 63.63$_{\pm23.68}$ & \textbf{85.78}$_{\pm16.97}$ & 83.46$_{\pm15.25}$ \\
\textcolor{black}{F1@50} & 22.14$_{\pm12.62}$ & 45.06$_{\pm24.49}$ & 47.23$_{\pm23.96}$ & 57.38$_{\pm26.47}$ & 78.57$_{\pm20.06}$ & \textbf{77.05}$_{\pm18.26}$ \\
\midrule
\multicolumn{7}{l}{\textbf{AutoLaparo}} \\
\midrule
Acc & 23.95$_{\pm6.53}$ & 25.01$_{\pm5.59}$ & 73.79$_{\pm11.63}$ & 81.50$_{\pm9.17}$ & 86.52$_{\pm6.60}$ & \textbf{86.59}$_{\pm7.64}$ \\
F1 & 11.52$_{\pm3.11}$ & 5.68$_{\pm1.05}$ & 52.98$_{\pm6.58}$ & 58.57$_{\pm5.99}$ & 72.32$_{\pm6.90}$ & \textbf{73.34}$_{\pm7.23}$ \\
\textcolor{black}{F1@10} & 0.11$_{\pm0.13}$ & 14.02$_{\pm2.10}$ & 12.14$_{\pm6.34}$ & 36.93$_{\pm15.99}$ & \textbf{69.43}$_{\pm14.23}$ & 62.96$_{\pm13.47}$ \\
\textcolor{black}{F1@25} & 0.00$_{\pm0.00}$ & 5.40$_{\pm6.27}$ & 10.61$_{\pm5.14}$ & 35.31$_{\pm14.75}$ & \textbf{68.24}$_{\pm14.75}$ & 60.53$_{\pm13.01}$ \\
\textcolor{black}{F1@50} & 0.00$_{\pm0.00}$ & 0.00$_{\pm0.00}$ & 6.88$_{\pm4.29}$ & 31.47$_{\pm14.49}$ & \textbf{59.90}$_{\pm14.15}$ & 50.47$_{\pm17.68}$ \\
\midrule
\multicolumn{7}{l}{\textbf{GraSP}} \\
\midrule
Acc & 32.22$_{\pm6.52}$ & 32.22$_{\pm6.52}$ & 17.85$_{\pm3.02}$ & 57.97$_{\pm4.08}$ & \textbf{66.58}$_{\pm5.97}$ & 63.69$_{\pm6.83}$ \\
F1 & 4.40$_{\pm6.70}$ & 4.40$_{\pm6.70}$ & 7.61$_{\pm11.80}$ & 37.83$_{\pm11.06}$ & \textbf{48.98}$_{\pm13.08}$ & 44.89$_{\pm10.28}$ \\
\textcolor{black}{F1@10} & 0.69$_{\pm8.70}$ & 0.69$_{\pm8.70}$ & 1.82$_{\pm0.87}$ & 22.79$_{\pm5.68}$ & \textbf{38.83}$_{\pm6.80}$ & 38.55$_{\pm4.20}$ \\
\textcolor{black}{F1@25} & 0.00$_{\pm0.00}$ & 0.00$_{\pm0.00}$ & 0.67$_{\pm0.40}$ & 18.08$_{\pm5.42}$ & \textbf{32.02}$_{\pm6.12}$ & 32.10$_{\pm5.55}$ \\
\textcolor{black}{F1@50} & 0.00$_{\pm0.00}$ & 0.00$_{\pm0.00}$ & 0.20$_{\pm0.14}$ & 9.05$_{\pm4.66}$ & \textbf{18.76}$_{\pm5.04}$ & 17.83$_{\pm4.67}$ \\
\bottomrule
\end{tabular}}
\label{tab:temporal_probing}
\end{table}

\vspace{1.2mm}
\noindent \textbf{Results.} Table~\ref{tab:temporal_probing} shows that SurgCLIP achieves the highest F1 and Accuracy performance on Cholec80 and AutoLaparo, while SurgCLIP$_{(\beta)}$ attains the best overall performance in GraSP, confirming that our learned representations transfer effectively to temporal models. On Cholec80 and AutoLaparo, MS-TCN probing provides additional gains compared to 32-frame linear probing, highlighting the value of extended temporal context for these datasets. In contrast, on GraSP phase recognition, temporal probing does not surpass linear probing. We attribute this to two dataset-specific factors: (1) GraSP contains long idle or non-informative intervals between annotated phases, which weakens the temporal consistency within videos; and (2) the limited number of full-length videos reduces the effective sample size for video-level training, whereas frame-wise linear probing mitigates this limitation.

\noindent \textcolor{black}{When examining segmental F1 metrics, we observe that although SurgCLIP yields slightly higher frame-level accuracy and F1 on AutoLaparo and Cholec80, SurgCLIP$_{(\beta)}$ demonstrates substantially improved temporal consistency across nearly all thresholds. These findings suggest that the pretraining distribution in SurgLaVi-$\beta$ produces features with stronger temporal coherence for certain downstream tasks compared to the full SurgLaVi corpus. On GraSP phases, SurgCLIP$_{(\beta)}$ outperforms the full model across frame-level and segmental metrics, which we attribute to the higher proportion of prostatectomy procedures in SurgLaVi-$\beta$ (Figure~\ref{distribution}), resulting in improved domain alignment.}
\vspace{1.2mm}
\noindent \textcolor{black}{
Beyond our two models, the temporal probing results also reveal clear performance differences across existing baselines. CLIP provides very limited temporal signal to MS-TCN, resulting in weak, or even zero, segmental F1 scores and failing to capture procedural regularities. MedSigLIP, despite being trained on medical image-text pairs, remains insufficiently specialized for surgical dynamics and offers only modest improvements. Surgical VLMs such as PeskaVLP perform better at the frame level but still struggle with temporal consistency. Notably, LEMON performs better than CLIP, MedSigLIP, and PeskaVLP, demonstrating that large-scale self-supervised in-domain pretraining produces more informative features for temporal modeling. However, LEMON still trails behind our SurgCLIP models due to the absence of semantic procedural grounding. In contrast, SurgCLIP and SurgCLIP$_{(\beta)}$ deliver consistently superior temporal segmentation, often more than doubling the F1@10 of the next-best baseline, demonstrating that large-scale surgical hierarchical vision-language alignment is crucial for learning representations with strong temporal structure.}

\subsection{Upper-Bound Results: Comparison with Task-Specific Models }
\label{sec:dataset-specific}

\begin{table*}[h]
\centering
\caption{\textcolor{black}{\textbf{Upper-bound results for surgical phase recognition on AutoLaparo and Cholec80.} SurgCLIP models comparison against fully supervised methods. Fully supervised models are trained end-to-end on the full dataset, while our approach uses linear probing with CLS few-shot learning (1-16 samples per class) and video-based splits from the training set (10\%, 50\%, Full). Both SurgCLIP variants demonstrate competitive performance with significantly fewer trainable parameters. $^{\textcolor{red}{\bigstar}}$: trainable, $^{\textcolor{blue}{\CIRCLE}}$: frozen, \textbf{Acc:} Accuracy, \textbf{P:} Precision, \textbf{R:} Recall, \textbf{J:} Jaccard, $^{+}$: 10-second relaxed boundary metrics. Standard deviations reported for few-shot runs.}}
\label{tab:autolaparo_cholec80_sota}
\small
\resizebox{\textwidth}{!}{
\begin{tabular}{@{}clllcccccccc@{}}
\toprule
\textbf{Training} & & & & \multicolumn{4}{c}{\textbf{AutoLaparo}} & \multicolumn{4}{c}{\textbf{Cholec80}} \\
\cmidrule(lr){5-8} \cmidrule(lr){9-12}
\textbf{Samples} & \textbf{Model} & \textbf{Backbone} & \textbf{Head} & \textbf{Acc (\%)} & \textbf{P (\%)} & \textbf{R (\%)} & \textbf{J (\%)} & \textbf{Acc (\%)} & \textbf{P (\%)} & \textbf{R (\%)} & \textbf{J (\%)} \\
\midrule
\multirow{9}{*}{Full} & EndoNet~\citep{endonet} & CNN$^{\textcolor{red}{\bigstar}}$ & Linear Layer$^{\textcolor{red}{\bigstar}}$ & -- & -- & -- & -- & $81.7^{+}$ & $73.7^{+}$ & $79.6^{+}$ & -- \\
& SV-RCNet~\citep{svrcnet} & CNN$^{\textcolor{red}{\bigstar}}$ & LSTM$^{\textcolor{red}{\bigstar}}$ & 75.6 & 64.0 & 59.7 & 47.2 & $85.3^{+}$ & $80.7^{+}$ & $83.5^{+}$ & -- \\
& TMRNet~\citep{tmrnet} & CNN$^{\textcolor{red}{\bigstar}}$ & LSTM$^{\textcolor{red}{\bigstar}}$ & 78.2 & 66.0 & 61.5 & 49.6 & $90.1^{+}$ & $90.3^{+}$ & $89.5^{+}$ & $79.1^{+}$ \\
& TeCNO~\citep{czempiel2020tecno} & CNN$^{\textcolor{red}{\bigstar}}$ & MS-TCN$^{\textcolor{red}{\bigstar}}$ & 77.3 & 66.9 & 64.6 & 50.7 & $88.6^{+}$ & $86.5^{+}$ & $87.6^{+}$ & $75.1^{+}$ \\
& Trans-SVNet~\citep{gao2021trans} & CNN$^{\textcolor{red}{\bigstar}}$ & Transformer / TCN$^{\textcolor{red}{\bigstar}}$ & 78.3 & 64.2 & 62.1 & 50.7 & 89.1 & 84.7 & 83.6 & 72.5 \\
& AVT~\citep{girdhar2021anticipative} & Transformer$^{\textcolor{red}{\bigstar}}$ & Transformer$^{\textcolor{red}{\bigstar}}$ & 77.8 & 68.0 & 82.2 & 50.7 & 86.7 & 77.3 & 82.1 & 66.4 \\
& LoViT~\citep{liu2025lovit} & Transformer$^{\textcolor{red}{\bigstar}}$ & Transformer$^{\textcolor{red}{\bigstar}}$ & 81.4 & 85.1 & 65.9 & 56.0 & 91.5 & 83.1 & 86.6 & 74.2 \\
& SKiT~\citep{liu2023skit} & Transformer$^{\textcolor{red}{\bigstar}}$ & Transformer$^{\textcolor{red}{\bigstar}}$ & 82.9 & 81.8 & 70.1 & 59.9 & 92.5 & 84.6 & 88.5 & 76.7 \\
& Surgformer~\citep{yang2024surgformer} & Transformer$^{\textcolor{red}{\bigstar}}$ & Linear Layer$^{\textcolor{red}{\bigstar}}$ & 85.7 & 82.3 & 75.7 & 66.7 & 92.4 & 87.9 & 89.3 & 79.9 \\
\cmidrule{2-12}
& SurgCLIP$_{(\mathbf{\beta})}$ & Transformer$^{\textcolor{blue}{\CIRCLE}}$ & Linear Layer$^{\textcolor{red}{\bigstar}}$ & 81.9 & 74.4 & 70.4 & 60.6 & 84.8 & 79.8 & 80.0 & 60.6 \\
& SurgCLIP & Transformer$^{\textcolor{blue}{\CIRCLE}}$ & Linear Layer$^{\textcolor{red}{\bigstar}}$ & 84.6 & 71.3 & 71.3 & 61.3 & 85.2 & 80.2 & 80.4 & 67.3 \\
\midrule
\multicolumn{12}{l}{\textit{CLS Few-Shot Linear Probing (samples per class)}} \\
\midrule
1 & \multirow{5}{*}{SurgCLIP$_{(\mathbf{\beta})}$} & \multirow{5}{*}{Transformer$^{\textcolor{blue}{\CIRCLE}}$} & \multirow{5}{*}{Linear Layer$^{\textcolor{red}{\bigstar}}$} & $64.7_{\pm 8.7}$ & $56.9_{\pm 7.4}$ & $55.8_{\pm 6.7}$ & $39.6_{\pm 8.1}$ & $58.0_{\pm 11.0}$ & $49.9_{\pm 9.1}$ & $54.4_{\pm 7.3}$ & $33.9_{\pm 9.4}$ \\
2 & & & & $67.7_{\pm 6.6}$ & $61.1_{\pm 5.5}$ & $60.0_{\pm 6.0}$ & $45.5_{\pm 5.8}$ & $64.1_{\pm 4.9}$ & $58.5_{\pm 3.5}$ & $63.4_{\pm 4.9}$ & $41.6_{\pm 2.2}$ \\
4 & & & & $72.8_{\pm 6.9}$ & $63.8_{\pm 6.0}$ & $64.2_{\pm 7.0}$ & $50.3_{\pm 6.4}$ & $70.1_{\pm 4.3}$ & $63.2_{\pm 2.8}$ & $70.4_{\pm 2.7}$ & $48.7_{\pm 3.2}$ \\
8 & & & & $79.8_{\pm 1.0}$ & $69.8_{\pm 2.6}$ & $71.0_{\pm 1.6}$ & $57.7_{\pm 2.0}$ & $74.4_{\pm 1.1}$ & $67.2_{\pm 0.5}$ & $73.7_{\pm 1.5}$ & $53.1_{\pm 1.4}$ \\
16 & & & & $80.5_{\pm 2.1}$ & $71.1_{\pm 2.3}$ & $72.1_{\pm 1.7}$ & $58.9_{\pm 2.4}$ & $78.5_{\pm 1.4}$ & $70.2_{\pm 1.5}$ & $77.5_{\pm 1.2}$ & $57.5_{\pm 1.8}$ \\
\midrule
1 & \multirow{5}{*}{SurgCLIP} & \multirow{5}{*}{Transformer$^{\textcolor{blue}{\CIRCLE}}$} & \multirow{5}{*}{Linear Layer$^{\textcolor{red}{\bigstar}}$} & $66.0_{\pm 7.0}$ & $57.9_{\pm 6.3}$ & $56.6_{\pm 6.1}$ & $40.9_{\pm 7.3}$ & $58.5_{\pm 9.1}$ & $54.3_{\pm 7.6}$ & $55.7_{\pm 6.3}$ & $36.5_{\pm 6.8}$ \\
2 & & & & $67.5_{\pm 5.8}$ & $61.1_{\pm 4.6}$ & $59.4_{\pm 5.0}$ & $44.5_{\pm 5.2}$ & $62.0_{\pm 4.3}$ & $58.4_{\pm 4.1}$ & $63.2_{\pm 3.7}$ & $41.9_{\pm 4.5}$ \\
4 & & & & $73.8_{\pm 6.4}$ & $65.3_{\pm 5.7}$ & $64.2_{\pm 6.8}$ & $51.0_{\pm 6.8}$ & $68.1_{\pm 3.4}$ & $64.0_{\pm 1.7}$ & $70.4_{\pm 3.1}$ & $49.5_{\pm 2.9}$ \\
8 & & & & $79.9_{\pm 1.3}$ & $69.6_{\pm 0.6}$ & $71.0_{\pm 1.4}$ & $57.2_{\pm 0.8}$ & $74.5_{\pm 0.7}$ & $67.2_{\pm 0.7}$ & $76.5_{\pm 1.3}$ & $55.0_{\pm 1.2}$ \\
16 & & & & $81.2_{\pm 2.3}$ & $71.5_{\pm 2.2}$ & $72.9_{\pm 2.4}$ & $59.8_{\pm 2.4}$ & $77.6_{\pm 1.4}$ & $70.1_{\pm 1.4}$ & $78.9_{\pm 1.1}$ & $58.0_{\pm 1.8}$ \\
\midrule
\multicolumn{12}{l}{\textit{Video Linear Probing (\% of available labeled video cases)}} \\
\midrule
10\% & \multirow{2}{*}{SurgCLIP$_{(\mathbf{\beta})}$} & \multirow{2}{*}{Transformer$^{\textcolor{blue}{\CIRCLE}}$} & \multirow{2}{*}{Linear Layer$^{\textcolor{red}{\bigstar}}$} & $69.0_{\pm 12.9}$ & $58.6_{\pm 14.1}$ & $57.3_{\pm 8.4}$ & $44.5_{\pm 10.8}$ & $76.7_{\pm 2.3}$ & $72.7_{\pm 0.9}$ & $70.4_{\pm 3.3}$ & $55.0_{\pm 1.7}$ \\
50\% & & & & $81.1_{\pm 1.4}$ & $72.0_{\pm 1.3}$ & $67.7_{\pm 1.9}$ & $57.6_{\pm 2.1}$ & $82.7_{\pm 1.0}$ & $77.9_{\pm 1.2}$ & $76.6_{\pm 2.7}$ & $62.8_{\pm 2.4}$ \\
\midrule
10\% & \multirow{2}{*}{SurgCLIP} & \multirow{2}{*}{Transformer$^{\textcolor{blue}{\CIRCLE}}$} & \multirow{2}{*}{Linear Layer$^{\textcolor{red}{\bigstar}}$} & $70.2_{\pm 11.9}$ & $55.6_{\pm 13.9}$ & $57.9_{\pm 8.2}$ & $45.1_{\pm 10.4}$ & $78.6_{\pm 1.0}$ & $74.0_{\pm 0.9}$ & $72.0_{\pm 3.9}$ & $57.4_{\pm 2.2}$ \\
50\% & & & & $82.1_{\pm 1.4}$ & $71.4_{\pm 1.2}$ & $68.5_{\pm 1.9}$ & $58.3_{\pm 2.2}$ & $83.8_{\pm 0.7}$ & $78.5_{\pm 1.0}$ & $78.5_{\pm 1.8}$ & $64.9_{\pm 1.4}$ \\
\bottomrule
\end{tabular}}
\end{table*}

\begin{table}[h]
\centering
\caption{\textcolor{black}{\textbf{Upper-bound results for surgical phase and step recognition on GraSP.} SurgCLIP models comparison against fully supervised methods. Fully supervised models are trained end-to-end on the full dataset, while our approach uses linear probing with CLS few-shot learning (1-16 samples per class) and video-based splits from the training set (10\%, 50\%, Full). Both SurgCLIP variants demonstrate competitive performance with significantly fewer trainable parameters. $^{\textcolor{red}{\bigstar}}$: trainable, $^{\textcolor{blue}{\CIRCLE}}$: frozen. Standard deviations reported for few-shot runs.}}
\label{tab:grasp_results}
\small
\resizebox{0.48\textwidth}{!}{
\begin{tabular}{@{}clllcc@{}}
\toprule
\textbf{Samples} & \textbf{Model} & \textbf{Backbone} & \textbf{Head} & \textbf{F1 Phases (\%)} & \textbf{F1 Steps (\%)} \\
\midrule
\multirow{6}{*}{Full} & SlowFast~\citep{feichtenhofer2019slowfast} & CNN$^{\textcolor{red}{\bigstar}}$ & Linear Layer$^{\textcolor{red}{\bigstar}}$ & 54.7 & 37.1 \\
& TAPIR~\citep{valderrama2022towards} & Transformer$^{\textcolor{red}{\bigstar}}$ & Linear Layer$^{\textcolor{red}{\bigstar}}$ & 58.6 & 43.8 \\
& TAPIS~\citep{ayobi2025pixel} & Transformer$^{\textcolor{red}{\bigstar}}$ & Linear Layer$^{\textcolor{red}{\bigstar}}$ & 63.4 & 45.8 \\
& MuST~\citep{perez2024must} & Transformer$^{\textcolor{red}{\bigstar}}$ & Transformer$^{\textcolor{red}{\bigstar}}$ & 68.8 & -- \\
\cmidrule{2-6}
& SurgCLIP$_{(\mathbf{\beta})}$ & Transformer$^{\textcolor{blue}{\CIRCLE}}$ & Linear Layer$^{\textcolor{red}{\bigstar}}$ & 58.9 & 35.5 \\
& SurgCLIP & Transformer$^{\textcolor{blue}{\CIRCLE}}$ & Linear Layer$^{\textcolor{red}{\bigstar}}$ & 61.1 & 38.5 \\
\midrule
\multicolumn{6}{l}{\textit{CLS Few-Shot Linear Probing (samples per class)}} \\
\midrule
1 & \multirow{5}{*}{SurgCLIP$_{(\mathbf{\beta})}$} & \multirow{5}{*}{Transformer$^{\textcolor{blue}{\CIRCLE}}$} & \multirow{5}{*}{Linear Layer$^{\textcolor{red}{\bigstar}}$} & $28.9_{\pm 5.7}$ & $16.9_{\pm 2.6}$ \\
2 & & & & $38.5_{\pm 0.9}$ & $22.9_{\pm 2.7}$ \\
4 & & & & $42.3_{\pm 4.3}$ & $27.2_{\pm 4.0}$ \\
8 & & & & $48.4_{\pm 2.7}$ & $29.9_{\pm 1.8}$ \\
16 & & & & $52.0_{\pm 3.1}$ & $33.5_{\pm 1.9}$ \\
\midrule
1 & \multirow{5}{*}{SurgCLIP} & \multirow{5}{*}{Transformer$^{\textcolor{blue}{\CIRCLE}}$} & \multirow{5}{*}{Linear Layer$^{\textcolor{red}{\bigstar}}$} & $29.5_{\pm 5.6}$ & $16.7_{\pm 1.7}$ \\
2 & & & & $38.8_{\pm 1.7}$ & $22.6_{\pm 2.5}$ \\
4 & & & & $42.0_{\pm 4.1}$ & $27.7_{\pm 2.5}$ \\
8 & & & & $47.7_{\pm 3.0}$ & $30.0_{\pm 1.1}$ \\
16 & & & & $52.7_{\pm 2.0}$ & $33.3_{\pm 1.7}$ \\
\midrule
\multicolumn{6}{l}{\textit{Video Linear Probing (\% of available labeled video cases)}} \\
\midrule
10\% & \multirow{2}{*}{SurgCLIP$_{(\mathbf{\beta})}$} & \multirow{2}{*}{Transformer$^{\textcolor{blue}{\CIRCLE}}$} & \multirow{2}{*}{Linear Layer$^{\textcolor{red}{\bigstar}}$} & $43.3_{\pm 3.2}$ & $23.9_{\pm 2.5}$ \\
50\% & & & & $55.4_{\pm 2.5}$ & $33.6_{\pm 1.7}$ \\
\midrule
10\% & \multirow{2}{*}{SurgCLIP} & \multirow{2}{*}{Transformer$^{\textcolor{blue}{\CIRCLE}}$} & \multirow{2}{*}{Linear Layer$^{\textcolor{red}{\bigstar}}$} & $46.2_{\pm 4.4}$ & $23.6_{\pm 3.7}$ \\
50\% & & & & $56.3_{\pm 11.2}$ & $33.8_{\pm 2.6}$ \\
\bottomrule
\end{tabular}}
\end{table}

\noindent  \textcolor{black}{Having demonstrated superior performance over existing VLMs and self-supervised foundation models in surgery across multiple evaluation settings including linear probing, temporal probing, and CoOp-based adaptation, we now benchmark SurgCLIP and SurgCLIP$_{(\mathbf{\beta})}$ against task-specific architectures on Cholec80, AutoLaparo, and GraSP (Phases and Steps). We position these fully-supervised methods as upper-bound performance references, as they perform end-to-end training of both backbone and heads (often with dataset-tailored architectural choices), whereas our approach relies on frozen features from our SurgLaVi-pretrained models with only a lightweight linear classification layer being tuned under various few-shot and full-shot settings described in Section~\ref{sec:linear-probing}. To contextualize our approach within the broader surgical phase recognition literature, we adapted our experimental protocol to align with established benchmarks for AutoLaparo and Cholec80. Specifically, we have included Accuracy, Precision, Recall, and Jaccard score used for these datasets as defined in~\citep{funke2023metrics}, and employed an online inference setup to ensure fair comparison with specialized models. }

\noindent \textcolor{black}{Table~\ref{tab:autolaparo_cholec80_sota} presents our results on the AutoLaparo and Cholec80 datasets. On AutoLaparo, our models trained with only 16 samples per class surpass CNN-based baselines with complex temporal heads (LSTM, TCN, Transformer, MS-TCN) across Accuracy, Precision, and Jaccard metrics, despite using only a single trainable linear layer. While performance remains below fully-trainable transformer-based architectures with sophisticated temporal aggregation methods, this gap narrows substantially when full training data is used. Under this setting, SurgCLIP achieves competitive Accuracy (84.6\%), Recall (71.3\%), and Jaccard score (61.3\%), demonstrating strong transfer from our pretraining corpus. We observe a different trend on Cholec80, where the 16-few-shot regime proves insufficient to get close to the performance of specialized models. This reflects the larger training data disparity between AutoLaparo and Cholec80, where specialized models in Cholec80 benefit from 4$\times$ more training videos than those in AutoLaparo. However, with 50\% of the training data, our models achieve performance comparable to CNN-based architectures. With full data availability, SurgCLIP matches the transformer-based AVT model and remains competitive with other heavy transformer-based models, all while requiring only a linear classification layer during training.}

\noindent  \textcolor{black}{For GraSP phase and step recognition (Table~\ref{tab:grasp_results}), our linear probe matches CNN-based baselines using only 16 samples per class. With full training data, SurgCLIP’s linear probe outperforms TAPIR (Transformer-based) and SlowFast (CNN-based) while using minimal trainable parameters and is only 2.34\% F1 lower compared to TAPIS. Performance remains below MuST, which achieves stronger results by employing substantially more parameters through multiple multi-scale transformer heads stacked on top of its transformer-based vision backbone.}

\noindent  \textcolor{black}{Taken together, these results demonstrate that SurgCLIP delivers competitive recognition performance across diverse surgical datasets using only frozen features and a single trainable linear layer, substantially reducing training cost. Although specialized architectures that jointly optimize backbone and temporal aggregation still maintain a 7.4\% advantage on average, the gap is modest relative to their significantly higher architectural complexity and training requirements.}

\subsection{Ablation studies}\label{subsec:ablation_studies}

\noindent To evaluate the contribution of different components in our pre-training setup, we conduct ablation studies on the temporal granularity of pre-training data, our data curation pipeline, \textcolor{black}{and our dynamic sampling during training}. We assess zero-shot performance on phase recognition in Cholec80, AutoLaparo, and GraSP, as well as step recognition in GraSP, enabling us to validate our contributions across datasets with varying surgical workflows and temporal structures.

\vspace{1.2mm}

\noindent \textbf{Ablation on hierarchical pre-training levels.} As shown in Table~\ref{tab:level_ablations}, joint multi-level pre-training that uses fine, mid, and coarse data together achieves the best overall zero shot performance across datasets. This supports the idea that supervision from different temporal scales provides complementary benefits. When each level is used on its own, pre-training with fine-level data or mid-level data produces stronger results than pre-training with coarse-level data. \textcolor{black}{In the context of traditional visual recognition of surgical workflow, visual models often perform better on coarse task such as phases because they last longer, change less frequently, and do not require the same level of temporal precision as granular tasks. However, in our multi-modal setting, fine-level and mid-level data produce a larger number of training samples, and the captions are shorter, more focused, and easier for the model to relate to the visual content, as shown in Figure~\ref{fig:clip_caption_example}. Coarse-level clips are longer and less numerous, and the corresponding captions are broader and more complex, which makes it more difficult for the model to learn strong associations when coarse-level data is used alone. When all levels are combined during pre-training, the model benefits from the clarity and abundance of the fine-grained supervision, while also learning from the richer procedural information contained in coarse-level descriptions.}

\begin{table}[!htpb]
\caption{\textcolor{black}{\textbf{Ablation study on hierarchical pre-training data.} Zero-shot F1-score across surgical datasets when pre-training with Fine- (F), Mid- (M), and Coarse- (C) level data individually and in combination.}}
\resizebox{0.48\textwidth}{!}{
\begin{tabular}{ccccccc}
\toprule
F     & M     & C     & Cholec80       & AutoLaparo     & \begin{tabular}[c]{@{}c@{}}GraSP \\ Phases\end{tabular} & \begin{tabular}[c]{@{}c@{}}GraSP \\ Steps\end{tabular} \\ \hline
x     & x     & \checkmark & 32.34$_{\pm6.54}$& 43.70$_{\pm4.95}$& 20.76$_{\pm4.23}$& 12.85$_{\pm4.03}$
\\
x     & \checkmark & x     & 39.72$_{\pm6.42}$& 43.65$_{\pm3.67}$& 23.73$_{\pm5.44}$& 13.57$_{\pm3.97}$
\\
\checkmark & x     & x     & 39.12$_{\pm7.59}$& 53.48$_{\pm5.62}$& 31.15$_{\pm7.80}$& \textbf{16.81}$_{\pm6.42}$
\\
\checkmark & \checkmark & \checkmark & \textbf{50.53}$_{\pm8.83}$& \textbf{56.37}$_{\pm5.54}$& \textbf{34.94}$_{\pm7.61}$& 16.53$_{\pm5.44}$\\ \bottomrule
\end{tabular}}
\label{tab:level_ablations}
\end{table}
\begin{table}[h]
\centering
\caption{\textcolor{black}{\textbf{Ablation study of the pipeline components.} The model is pre-trained on different dataset versions corresponding to successive stages of our data curation pipeline: Multi-Level Semantic Captioning (MLSC), Dual-Modal Filtering (DMF), and Contextual Enrichment (CE). Zero-shot F1-scores are reported, with the final row showing the performance of the complete method. }}
\resizebox{0.48\textwidth}{!}{%
\begin{tabular}{lcccccc}
\toprule
\textbf{DMF} & \textbf{MLSC} & \textbf{CE} & \textbf{Cholec80} & \textbf{AutoLaparo} & \textbf{\begin{tabular}[c]{@{}c@{}}GraSP \\ Phases\end{tabular}} & \textbf{\begin{tabular}[c]{@{}c@{}}GraSP \\ Steps\end{tabular}} \\ 
\midrule
\multicolumn{7}{l}{\textit{Baseline}} \\
\multicolumn{1}{r}{$\times$} & $\times$ & $\times$ & 35.89$_{\pm5.98}$ & 46.10$_{\pm9.31}$ & 24.59$_{\pm7.71}$ & 13.35$_{\pm4.17}$ \\
\midrule
\multicolumn{7}{l}{\textit{Individual components}} \\
\multicolumn{1}{r}{\checkmark} & $\times$ & $\times$ & 35.88$_{\pm6.03}$ & 43.42$_{\pm7.26}$ & 31.69$_{\pm7.39}$ & 14.33$_{\pm4.43}$ \\
\multicolumn{1}{r}{$\times$} & \checkmark & $\times$ & 36.11$_{\pm7.60}$ & 36.19$_{\pm7.11}$ & 27.50$_{\pm6.88}$ & 12.52$_{\pm4.25}$ \\
\multicolumn{1}{r}{$\times$} & $\times$ & \checkmark & 33.42$_{\pm7.11}$ & $50.06_{\pm6.54}$ & 24.20$_{\pm5.61}$ & 9.55$_{\pm3.14}$ \\
\midrule
\multicolumn{7}{l}{\textit{Pairwise combinations}} \\
\multicolumn{1}{r}{\checkmark} & $\times$ & \checkmark & 33.56$_{\pm8.01}$ & 51.65$_{\pm7.68}$ & 25.79$_{\pm5.39}$ & 11.91$_{\pm3.79}$ \\
\multicolumn{1}{r}{\checkmark} & \checkmark & $\times$ & 42.36$_{\pm8.42}$ & 47.34$_{\pm7.56}$ & 32.02$_{\pm8.15}$ & 14.57$_{\pm4.51}$ \\
\midrule
\multicolumn{7}{l}{\textit{Complete pipeline}} \\
\multicolumn{1}{r}{\checkmark} & \checkmark & \checkmark & \textbf{50.53}$_{\pm8.83}$ & \textbf{56.37}$_{\pm4.54}$ & \textbf{34.94}$_{\pm7.61}$ & \textbf{16.53}$_{\pm5.44}$ \\
\bottomrule
\end{tabular}}
\label{tab:segments_ablations}
\end{table}
\begin{table}[!htpb]
\caption{\textcolor{black}{\textbf{Ablation study on temporal sampling strategies.} Zero-shot F1-scores across multiple surgical datasets when pre-training with fixed clip-sampling strides of varying lengths compared to a dynamic stride strategy.}}
\resizebox{0.48\textwidth}{!}{
\begin{tabular}{@{}cccccc@{}}
\toprule
Mode   & Stride & Cholec80  & AutoLaparo& GraSP Phases   & GraSP Steps    \\ \midrule
\multirow{2}{*}{Fixed} & 1 & 43.75$_{\pm7.35}$ & 34.18$_{\pm9.43}$& 31.36$_{\pm8.87}$ & 14.19$_{\pm5.45}$ \\ \cmidrule(l){2-6} 
  & 4 & 40.45$_{\pm8.06}$& 41.82$_{\pm5.19}$ &24.75$_{\pm7.96}$& 13.15$_{\pm1.31}$\\ \midrule
\multicolumn{2}{c}{Dynamic}& \textbf{50.53}$_{\pm8.83}$ & \textbf{56.37}$_{\pm4.54}$ & \textbf{34.94}$_{\pm7.61}$ & \textbf{16.53}$_{\pm5.44}$ \\ \bottomrule
\end{tabular}}
\label{tab:sampling_ablations}
\end{table}

\noindent \textbf{Ablation on data curation pipeline components.} \textcolor{black}{Table~\ref{tab:segments_ablations} reports the effect of individual components and their combinations in our curation pipeline. We observe that components exhibit varying effectiveness in isolation but demonstrate strong synergistic effects when combined. Examining individual components, we find that their isolated contributions are inconsistent across datasets. Dual-Modal Filtering (DMF) alone yields mixed results: it maintains baseline performance on Cholec80 while showing gains on GraSP Phases (+7.10\%) but degradation on AutoLaparo (-2.68\%). Multi-Level Semantic Captioning (MLSC) in isolation provides minimal improvements and even decreases performance on AutoLaparo (36.19\% vs. 46.10\% baseline). Most notably, Contextual Enrichment (CE) alone consistently underperforms or matches the baseline across most benchmarks, with substantial drops on Cholec80 and GraSP Steps, despite gains on AutoLaparo. This reveals that individual components, when applied in isolation, do not consistently improve performance and may even introduce noise without complementary mechanisms. However, pairwise combinations begin to reveal the synergistic nature of our pipeline. Combining DMF with MLSC achieves notable improvements over the baseline (+6.47\% on Cholec80, +7.43\% on GraSP Phases), demonstrating that semantic diversity (MLSC) paired with quality filtering (DMF) creates more effective training data. Interestingly, pairing DMF with CE results in stronger performance on AutoLaparo (+5.55\% over baseline) but remains suboptimal on other datasets, suggesting that contextual enrichment requires semantic grounding to be fully effective. The complete pipeline with all three components achieves the best performance across all benchmarks, with substantial gains: +14.64\% on Cholec80, +10.27\% on AutoLaparo, +10.35\% on GraSP Phases, and +3.18\% on GraSP Steps compared to baseline. Notably, the full method outperforms even the best pairwise combination by significant margins (e.g., +8.17\% on Cholec80 over DMF+MLSC), demonstrating that all three components are mutually reinforcing rather than independently additive for effective vision-language alignment in surgical video understanding.}

\noindent \textcolor{black}{\textbf{Ablation on temporal sampling strategies.}  Table~\ref{tab:sampling_ablations} compares fixed-stride sampling with our dynamic sampling strategy during pre-training. Our dynamic sampling strategy adapts the sampling stride based on segment duration while maintaining a fixed number of frames per clip: shorter, task-level segments use smaller strides for high temporal resolution, whereas longer, coarse-level segments employ larger strides to increase temporal coverage. This ensures that each clip accurately reflects the temporal coverage described by its paired caption, maintaining proper vision-language alignment while exposing the model to multiple temporal resolutions during training. The results demonstrate the effectiveness of this approach, with our dynamic sampling strategy substantially outperforming fixed-stride alternatives across all benchmarks, resulting +6.78\% to +14.55\% improvement over stride-1 and +9.08\% to +22.19\% over stride-4 across datasets.}

\section{Limitations}\label{sec:limitations}
\noindent \textcolor{black}{
Our work shows that large scale multi-modal pretraining on hierarchically structured and automatically curated surgical video–language data can meaningfully advance generalizable representation learning in surgery, improving transferability across procedures and modalities while narrowing, although not yet closing, the gap with fully supervised and task specific models. These results highlight the promise of vision–language methods for broad surgical intelligence. Nonetheless, several limitations remain. Our pipeline works on narrated surgical videos, which are not uniformly available across all procedure types or clinical settings. In addition, narrations may not always reflect the exact moment shown on screen since surgeons often describe recent actions or upcoming steps, which can occasionally introduce temporal misalignment between captions and visual content. Furthermore, the fully automated curation process operates without expert verification, and although foundation models make it possible to process data at the scale required for effective pretraining, they may introduce noise or bias in the process. Future work could aim to leverage silent videos to complement narrated content by combining self-supervised methods with multi-modal contrastive learning to improve generalizability and robustness of learned representations.}

\section{Conclusion}
We introduce SurgLaVi, the largest and most hierarchically structured video–language dataset in the surgical domain, capturing procedural knowledge at coarse, mid, and fine levels from narrated surgical videos. Our fully automated pipeline transforms raw surgical video collections into temporally precise, semantically rich, multi-granular annotations at unprecedented scale, setting a new benchmark for dataset design. Building on this foundation, SurgCLIP, a lightweight CLIP-style model which we design as a baseline, demonstrates that even standard architectures, when paired with high-quality hierarchical data, can achieve state-of-the-art performance across a wide spectrum of downstream tasks, from phase and step recognition to tool and action detection, in zero-shot and few-shot probing evaluation settings. These findings reveal a fundamental principle: dataset quality and structure can be more decisive than model complexity in capturing complex procedural workflows. SurgLaVi not only accelerates representation learning for surgical AI but also opens transformative opportunities for multimodal reasoning, video–text understanding, and self-supervised procedural modeling, laying the groundwork for the next generation of intelligent, context-aware surgical systems.

%% Loading bibliography style file
%\bibliographystyle{model1-num-names}

% \textbf{Acknowledgments}

\bibliographystyle{IEEEtranN}

\bibliography{cas-refs}

@article{surgvlp,
  title={Learning multi-modal representations by watching hundreds of surgical video lectures},
  author={Yuan, Kun and Srivastav, Vinkle and Yu, Tong and Lavanchy, Joel L and Marescaux, Jacques and Mascagni, Pietro and Navab, Nassir and Padoy, Nicolas},
  journal={Medical Image Analysis},
  pages={103644},
  year={2025},
  publisher={Elsevier}
}

@article{peskavlp,
  title={Procedure-aware surgical video-language pretraining with hierarchical knowledge augmentation},
  author={Yuan, Kun and Navab, Nassir and Padoy, Nicolas and others},
  journal={Advances in Neural Information Processing Systems},
  volume={37},
  pages={122952--122983},
  year={2024}
}

@inproceedings{zhang2022tip,
  title={Tip-adapter: Training-free adaption of clip for few-shot classification},
  author={Zhang, Renrui and Zhang, Wei and Fang, Rongyao and Gao, Peng and Li, Kunchang and Dai, Jifeng and Qiao, Yu and Li, Hongsheng},
  booktitle={European conference on computer vision},
  pages={493--510},
  year={2022},
  organization={Springer}
}

@article{zhou2022learning,
  title={Learning to prompt for vision-language models},
  author={Zhou, Kaiyang and Yang, Jingkang and Loy, Chen Change and Liu, Ziwei},
  journal={International Journal of Computer Vision},
  volume={130},
  number={9},
  pages={2337--2348},
  year={2022},
  publisher={Springer}
}

@inproceedings{hecvl,
  title={Hecvl: Hierarchical video-language pretraining for zero-shot surgical phase recognition},
  author={Yuan, Kun and Srivastav, Vinkle and Navab, Nassir and Padoy, Nicolas},
  booktitle={International Conference on Medical Image Computing and Computer-Assisted Intervention},
  pages={306--316},
  year={2024},
  organization={Springer}
}

@inproceedings{vidlpro,
  title={VidLPRO: A Video-Language Pre-training Framework for Robotic and Laparoscopic Surgery},
  author={Honarmand, Mohammadmahdi and Jamal, Muhammad Abdullah and Mohareri, Omid},
  booktitle={Advancements In Medical Foundation Models: Explainability, Robustness, Security, and Beyond},
  year={2024}
}

@misc{che2025lemon,
      title={LEMON: A Large Endoscopic MONocular Dataset and Foundation Model for Perception in Surgical Settings}, 
      author={Chengan Che and Chao Wang and Tom Vercauteren and Sophia Tsoka and Luis C. Garcia-Peraza-Herrera},
      year={2025},
      eprint={2503.19740},
      archivePrefix={arXiv},
      primaryClass={cs.CV},
      url={https://arxiv.org/abs/2503.19740}, 
}

@article{gensurg,
  title={General surgery vision transformer: A video pre-trained foundation model for general surgery},
  author={Schmidgall, Samuel and Kim, Ji Woong and Jopling, Jeffrey and Krieger, Axel},
  journal={arXiv preprint arXiv:2403.05949},
  year={2024}
}

@article{ophclip,
  title={Ophclip: Hierarchical retrieval-augmented learning for ophthalmic surgical video-language pretraining},
  author={Hu, Ming and Yuan, Kun and Shen, Yaling and Tang, Feilong and Xu, Xiaohao and Zhou, Lin and Li, Wei and Chen, Ying and Xu, Zhongxing and Peng, Zelin and others},
  journal={arXiv preprint arXiv:2411.15421},
  year={2024}
}

@inproceedings{clip,
  title={Learning transferable visual models from natural language supervision},
  author={Radford, Alec and Kim, Jong Wook and Hallacy, Chris and Ramesh, Aditya and Goh, Gabriel and Agarwal, Sandhini and Sastry, Girish and Askell, Amanda and Mishkin, Pamela and Clark, Jack and others},
  booktitle={International conference on machine learning},
  pages={8748--8763},
  year={2021},
  organization={PmLR}
}

@inproceedings{wang2022autolaparo,
  title={Autolaparo: A new dataset of integrated multi-tasks for image-guided surgical automation in laparoscopic hysterectomy},
  author={Wang, Ziyi and Lu, Bo and Long, Yonghao and Zhong, Fangxun and Cheung, Tak-Hong and Dou, Qi and Liu, Yunhui},
  booktitle={International Conference on Medical Image Computing and Computer-Assisted Intervention},
  pages={486--496},
  year={2022},
  organization={Springer}
}

@inproceedings{bertasius2021space,
  title={Is space-time attention all you need for video understanding?},
  author={Bertasius, Gedas and Wang, Heng and Torresani, Lorenzo},
  booktitle={Icml},
  volume={2},
  number={3},
  pages={4},
  year={2021}
}

@inproceedings{devlin2019bert,
  title={Bert: Pre-training of deep bidirectional transformers for language understanding},
  author={Devlin, Jacob and Chang, Ming-Wei and Lee, Kenton and Toutanova, Kristina},
  booktitle={Proceedings of the 2019 conference of the North American chapter of the association for computational linguistics: human language technologies, volume 1 (long and short papers)},
  pages={4171--4186},
  year={2019}
}

@inproceedings{jia2021scaling,
  title={Scaling up visual and vision-language representation learning with noisy text supervision},
  author={Jia, Chao and Yang, Yinfei and Xia, Ye and Chen, Yi-Ting and Parekh, Zarana and Pham, Hieu and Le, Quoc and Sung, Yun-Hsuan and Li, Zhen and Duerig, Tom},
  booktitle={International conference on machine learning},
  pages={4904--4916},
  year={2021},
  organization={PMLR}
}

@inproceedings{zhai2023sigmoid,
  title={Sigmoid loss for language image pre-training},
  author={Zhai, Xiaohua and Mustafa, Basil and Kolesnikov, Alexander and Beyer, Lucas},
  booktitle={Proceedings of the IEEE/CVF international conference on computer vision},
  pages={11975--11986},
  year={2023}
}

@article{ayobi2025pixel,
  title={Pixel-wise recognition for holistic surgical scene understanding},
  author={Ayobi, Nicol{\'a}s and Rodr{\'\i}guez, Santiago and P{\'e}rez, Alejandra and Hern{\'a}ndez, Isabela and Aparicio, Nicol{\'a}s and Dessevres, Eug{\'e}nie and Pe{\~n}a, Sebasti{\'a}n and Santander, Jessica and Caicedo, Juan Ignacio and Fern{\'a}ndez, Nicol{\'a}s and others},
  journal={Medical Image Analysis},
  pages={103726},
  year={2025},
  publisher={Elsevier}
}

@software{open_clip,
  author       = {Ilharco, Gabriel and
                  Wortsman, Mitchell and
                  Wightman, Ross and
                  Gordon, Cade and
                  Carlini, Nicholas and
                  Taori, Rohan and
                  Dave, Achal and
                  Shankar, Vaishaal and
                  Namkoong, Hongseok and
                  Miller, John and
                  Hajishirzi, Hannaneh and
                  Farhadi, Ali and
                  Schmidt, Ludwig},
  title        = {OpenCLIP},
  month        = jul,
  year         = 2021,
  publisher    = {Zenodo},
  version      = {0.1},
  doi          = {10.5281/zenodo.5143773},
  url          = {https://doi.org/10.5281/zenodo.5143773}
}

@article{svrcnet,
  title={SV-RCNet: workflow recognition from surgical videos using recurrent convolutional network},
  author={Jin, Yueming and Dou, Qi and Chen, Hao and Yu, Lequan and Qin, Jing and Fu, Chi-Wing and Heng, Pheng-Ann},
  journal={IEEE transactions on medical imaging},
  volume={37},
  number={5},
  pages={1114--1126},
  year={2017},
  publisher={IEEE}
}

@article{tmrnet,
  title={Temporal memory relation network for workflow recognition from surgical video},
  author={Jin, Yueming and Long, Yonghao and Chen, Cheng and Zhao, Zixu and Dou, Qi and Heng, Pheng-Ann},
  journal={IEEE Transactions on Medical Imaging},
  volume={40},
  number={7},
  pages={1911--1923},
  year={2021},
  publisher={IEEE}
}

@inproceedings{gao2021trans,
  title={Trans-svnet: Accurate phase recognition from surgical videos via hybrid embedding aggregation transformer},
  author={Gao, Xiaojie and Jin, Yueming and Long, Yonghao and Dou, Qi and Heng, Pheng-Ann},
  booktitle={International conference on medical image computing and computer-assisted intervention},
  pages={593--603},
  year={2021},
  organization={Springer}
}

@inproceedings{liu2023skit,
  title={Skit: a fast key information video transformer for online surgical phase recognition},
  author={Liu, Yang and Huo, Jiayu and Peng, Jingjing and Sparks, Rachel and Dasgupta, Prokar and Granados, Alejandro and Ourselin, Sebastien},
  booktitle={Proceedings of the IEEE/CVF international conference on computer vision},
  pages={21074--21084},
  year={2023}
}

@inproceedings{yang2024surgformer,
  title={Surgformer: Surgical transformer with hierarchical temporal attention for surgical phase recognition},
  author={Yang, Shu and Luo, Luyang and Wang, Qiong and Chen, Hao},
  booktitle={International Conference on Medical Image Computing and Computer-Assisted Intervention},
  pages={606--616},
  year={2024},
  organization={Springer}
}

@article{liu2025lovit,
  title={Lovit: Long video transformer for surgical phase recognition},
  author={Liu, Yang and Boels, Maxence and Garcia-Peraza-Herrera, Luis C and Vercauteren, Tom and Dasgupta, Prokar and Granados, Alejandro and Ourselin, Sebastien},
  journal={Medical Image Analysis},
  volume={99},
  pages={103366},
  year={2025},
  publisher={Elsevier}
}

@inproceedings{perez2024must,
  title={Must: Multi-scale t ransformers for surgical phase recognition},
  author={P{\'e}rez, Alejandra and Rodr{\'\i}guez, Santiago and Ayobi, Nicol{\'a}s and Aparicio, Nicol{\'a}s and Dessevres, Eug{\'e}nie and Arbel{\'a}ez, Pablo},
  booktitle={International Conference on Medical Image Computing and Computer-Assisted Intervention},
  pages={422--432},
  year={2024},
  organization={Springer}
}

@inproceedings{feichtenhofer2019slowfast,
  title={Slowfast networks for video recognition},
  author={Feichtenhofer, Christoph and Fan, Haoqi and Malik, Jitendra and He, Kaiming},
  booktitle={Proceedings of the IEEE/CVF international conference on computer vision},
  pages={6202--6211},
  year={2019}
}

@inproceedings{valderrama2022towards,
  title={Towards holistic surgical scene understanding},
  author={Valderrama, Natalia and Ruiz Puentes, Paola and Hern{\'a}ndez, Isabela and Ayobi, Nicol{\'a}s and Verlyck, Mathilde and Santander, Jessica and Caicedo, Juan and Fern{\'a}ndez, Nicol{\'a}s and Arbel{\'a}ez, Pablo},
  booktitle={International conference on medical image computing and computer-assisted intervention},
  pages={442--452},
  year={2022},
  organization={Springer}
}

@inproceedings{girdhar2021anticipative,
  title={Anticipative video transformer},
  author={Girdhar, Rohit and Grauman, Kristen},
  booktitle={Proceedings of the IEEE/CVF international conference on computer vision},
  pages={13505--13515},
  year={2021}
}

@inproceedings{czempiel2020tecno,
  title={Tecno: Surgical phase recognition with multi-stage temporal convolutional networks},
  author={Czempiel, Tobias and Paschali, Magdalini and Keicher, Matthias and Simson, Walter and Feussner, Hubertus and Kim, Seong Tae and Navab, Nassir},
  booktitle={International conference on medical image computing and computer-assisted intervention},
  pages={343--352},
  year={2020},
  organization={Springer}
}

@inproceedings{lea2017temporal,
  title={Temporal convolutional networks for action segmentation and detection},
  author={Lea, Colin and Flynn, Michael D and Vidal, Rene and Reiter, Austin and Hager, Gregory D},
  booktitle={proceedings of the IEEE Conference on Computer Vision and Pattern Recognition},
  pages={156--165},
  year={2017}
}

@article{zhang2023surgical,
  title={Surgical workflow recognition with temporal convolution and transformer for action segmentation},
  author={Zhang, Bokai and Goel, Bharti and Sarhan, Mohammad Hasan and Goel, Varun Kejriwal and Abukhalil, Rami and Kalesan, Bindu and Stottler, Natalie and Petculescu, Svetlana},
  journal={International Journal of Computer Assisted Radiology and Surgery},
  volume={18},
  number={4},
  pages={785--794},
  year={2023},
  publisher={Springer}
}

@article{funke2023metrics,
  title={Metrics matter in surgical phase recognition},
  author={Funke, Isabel and Rivoir, Dominik and Speidel, Stefanie},
  journal={arXiv preprint arXiv:2305.13961},
  year={2023}
}

@article{wagner2023comparative,
  title={Comparative validation of machine learning algorithms for surgical workflow and skill analysis with the heichole benchmark},
  author={Wagner, Martin and others},
  journal={Medical Image Analysis},
  volume={86},
  pages={102770},
  year={2023},
  publisher={Elsevier}
}

@article{SurgicBERTa,
title = {Surgicberta: a pre-trained language model for procedural surgical language},
journal = {International Journal of Data Science and Analytics},
year = {2023},
doi = { https://doi.org/10.1007/s41060-023-00433-5 },
url = { https://link.springer.com/article/10.1007/s41060-023-00433-5 },
author = {Marco Bombieri and Marco Rospocher and Simone Paolo Ponzetto and Paolo Fiorini},
}

@software{medicalaws,
  author       = {AWS},
  title        = {Amazon Transcribe Medical},
  year         = 2023,
  url          = {https://aws.amazon.com/transcribe/medical/}
}

@article{gu2021open,
  title={Open-vocabulary object detection via vision and language knowledge distillation},
  author={Gu, Xiuye and Lin, Tsung-Yi and Kuo, Weicheng and Cui, Yin},
  journal={arXiv preprint arXiv:2104.13921},
  year={2021}
}

@misc{wandb,
title = {Experiment Tracking with Weights and Biases},
year = {2020},
note = {Software available from wandb.com},
url={https://www.wandb.com/},
author = {Biewald, Lukas},
}

@article{endonet,
  author={Twinanda, Andru P. and Shehata, Sherif and Mutter, Didier and Marescaux, Jacques and de Mathelin, Michel and Padoy, Nicolas},
  journal={IEEE Transactions on Medical Imaging}, 
  title={EndoNet: A Deep Architecture for Recognition Tasks on Laparoscopic Videos}, 
  year={2017},
  volume={36},
  doi={10.1109/TMI.2016.2593957}
}

@article{nwoye2022rendezvous,
  title={Rendezvous: Attention mechanisms for the recognition of surgical action triplets in endoscopic videos},
  author={Nwoye, Chinedu Innocent and Yu, Tong and Gonzalez, Cristians and Seeliger, Barbara and Mascagni, Pietro and Mutter, Didier and Marescaux, Jacques and Padoy, Nicolas},
  journal={Medical Image Analysis},
  volume={78},
  pages={102433},
  year={2022},
  publisher={Elsevier}
}

@article{sarrarp50,
  title={Sar-rarp50: Segmentation of surgical instrumentation and action recognition on robot-assisted radical prostatectomy challenge},
  author={Psychogyios, Dimitrios and Colleoni, Emanuele and Van Amsterdam, Beatrice and Li, Chih-Yang and Huang, Shu-Yu and Li, Yuchong and Jia, Fucang and Zou, Baosheng and Wang, Guotai and Liu, Yang and others},
  journal={arXiv preprint arXiv:2401.00496},
  year={2023}
}

@article{multibypass,
  title = {Challenges in multi-centric generalization: phase and step recognition in Roux-en-Y gastric bypass surgery},
  ISSN = {1861-6429},
  url = {http://dx.doi.org/10.1007/s11548-024-03166-3},
  DOI = {10.1007/s11548-024-03166-3},
  journal = {International Journal of Computer Assisted Radiology and Surgery},
  publisher = {Springer Science and Business Media LLC},
  author = {Lavanchy,  Joël L. and Ramesh,  Sanat and Dall’Alba,  Diego and Gonzalez,  Cristians and Fiorini,  Paolo and M\"{u}ller-Stich,  Beat P. and Nett,  Philipp C. and Marescaux,  Jacques and Mutter,  Didier and Padoy,  Nicolas},
  year = {2024},
  month = may 
}

@inproceedings{li2022blip,
  title={Blip: Bootstrapping language-image pre-training for unified vision-language understanding and generation},
  author={Li, Junnan and Li, Dongxu and Xiong, Caiming and Hoi, Steven},
  booktitle={International conference on machine learning},
  pages={12888--12900},
  year={2022},
  organization={PMLR}
}

@inproceedings{farha2019ms,
  title={Ms-tcn: Multi-stage temporal convolutional network for action segmentation},
  author={Farha, Yazan Abu and Gall, Jurgen},
  booktitle={Proceedings of the IEEE/CVF conference on computer vision and pattern recognition},
  pages={3575--3584},
  year={2019}
}

@article{lu2019vilbert,
  title={Vilbert: Pretraining task-agnostic visiolinguistic representations for vision-and-language tasks},
  author={Lu, Jiasen and Batra, Dhruv and Parikh, Devi and Lee, Stefan},
  journal={Advances in neural information processing systems},
  volume={32},
  year={2019}
}

@article{radford2022whisper,
  title={Whisper: Robust Speech Recognition via Large-Scale Weak Supervision},
  author={Radford, Alec and Kim, Jong Wook and Xu, Tao and Brockman, Greg and McLeavey, Christine and Sutskever, Ilya},
  journal={arXiv preprint arXiv:2212.04356},
  year={2022}
}

@article{bain2022whisperx,
  title={WhisperX: Time-Accurate Speech Transcription of Long-Form Audio},
  author={Bain, Max and Huh, Jaesung and Han, Tengda and Zisserman, Andrew},
  journal={INTERSPEECH 2023},
  year={2023}
}

@article{sages_consensus,
  title={SAGES consensus recommendations on an annotation framework for surgical video},
  author={Meireles, Ozanan R and Rosman, Guy and Altieri, Maria S and Carin, Lawrence and Hager, Gregory and Madani, Amin and Padoy, Nicolas and Pugh, Carla M and Sylla, Patricia and Ward, Thomas M and others},
  journal={Surgical endoscopy},
  volume={35},
  number={9},
  pages={4918--4929},
  year={2021},
  publisher={Springer}
}

\clearpage
%\newpage

\setcounter{section}{0}
\section{Appendix}
\setcounter{figure}{0}
\setcounter{table}{0}

\subsection{SurgLaVi Data Processing Pipeline Details}
\label{sec:pipeline_details_SM}

\noindent \textbf{Stage 1: Speech-to-text conversion.}  
We used WhisperX~\citep{bain2022whisperx}, built on Whisper Large v3~\citep{radford2022whisper}, to transcribe audio and generate millisecond-accurate word-level timestamps via forced alignment, enabling reliable video-text pairing for subsequent segmentation. \textcolor{black}{To select the most suitable Automatic Speech Recognition (ASR) model for our pipeline, we conducted a comparative evaluation of several ASR systems on a held-out test set of expert-narrated surgical videos totaling 2.5 hours of spoken content. Ground-truth transcripts were fully generated by surgeons. We computed Word Error Rate (WER) to assess transcription accuracy in the surgical domain. As shown in Table~\ref{tab:whisper_exps}, WhisperX (v3) achieved the lowest WER and most robust domain performance. These results guided our selection of WhisperX as the ASR component in Stage~1.}

\begin{table}[!htpb]
\caption{\textcolor{black}{\textbf{Word Error Rate (WER)} in surgical language transcriptions of different foundation models for Automatic Speech Recognition.}}
\centering
\resizebox{0.25\textwidth}{!}{
\begin{tabular}{@{}ccc@{}}
\toprule
Model                     & Version & WER (\%)    \\ \midrule
\multirow{2}{*}{Whisper Large}  & 2       & 7.69 $\pm$ 2.10  \\ \cmidrule(l){2-3} 
                          & 3       & 17.68 $\pm$ 6.46\\ \midrule
\multirow{2}{*}{WhisperX Large} & 2       & 5.99 $\pm$ 0.73 \\ \cmidrule(l){2-3} 
                          & 3       & \textbf{5.48 $\pm$ 1.04}  \\ \bottomrule
\end{tabular}}
\label{tab:whisper_exps}
\end{table}

\begin{figure}[b!]
  \centering
    \includegraphics[width=0.45\textwidth]{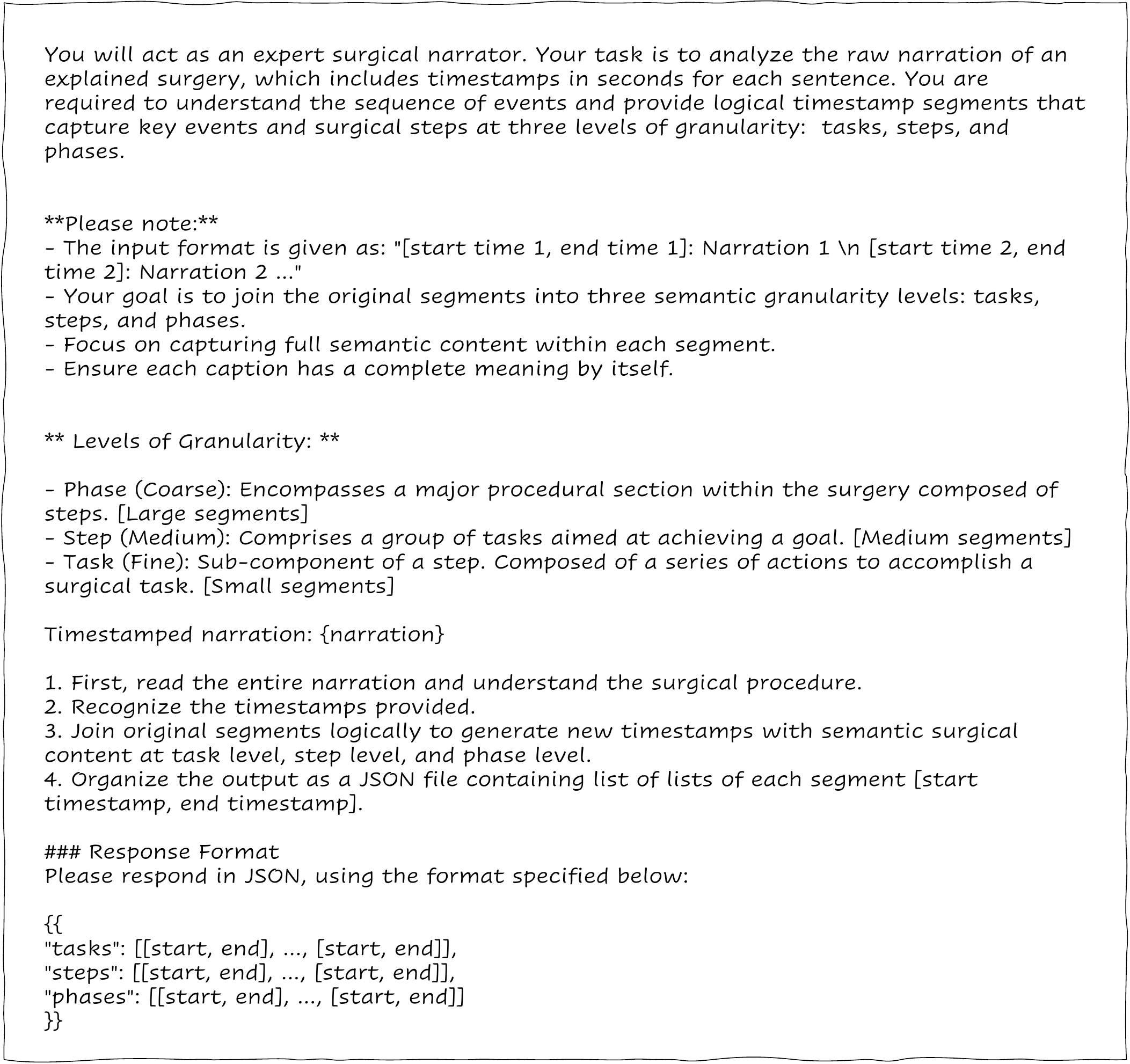}
    \caption{Prompt used for hierarchical semantic segmentation of narrations into tasks, steps, and phases.}
    \label{fig:prompt_hierarchical}
\end{figure}

\noindent \textbf{Stage 2: Hierarchical segmentation.}  
We employed GPT-4o to restructure transcriptions into three hierarchical levels of surgical workflow: phases, steps, and tasks, following the SAGES consensus guidelines~\citep{sages_consensus}. Figure~\ref{fig:prompt_hierarchical} shows the prompting template. Word-level timestamps were then used to align captions with clips, producing multi-granular datasets $\mathcal{D}_{\text{phase}}$, $\mathcal{D}_{\text{step}}$, and $\mathcal{D}_{\text{task}}$.

\begin{figure}[b!]
  \centering
    \includegraphics[width=0.40\textwidth]{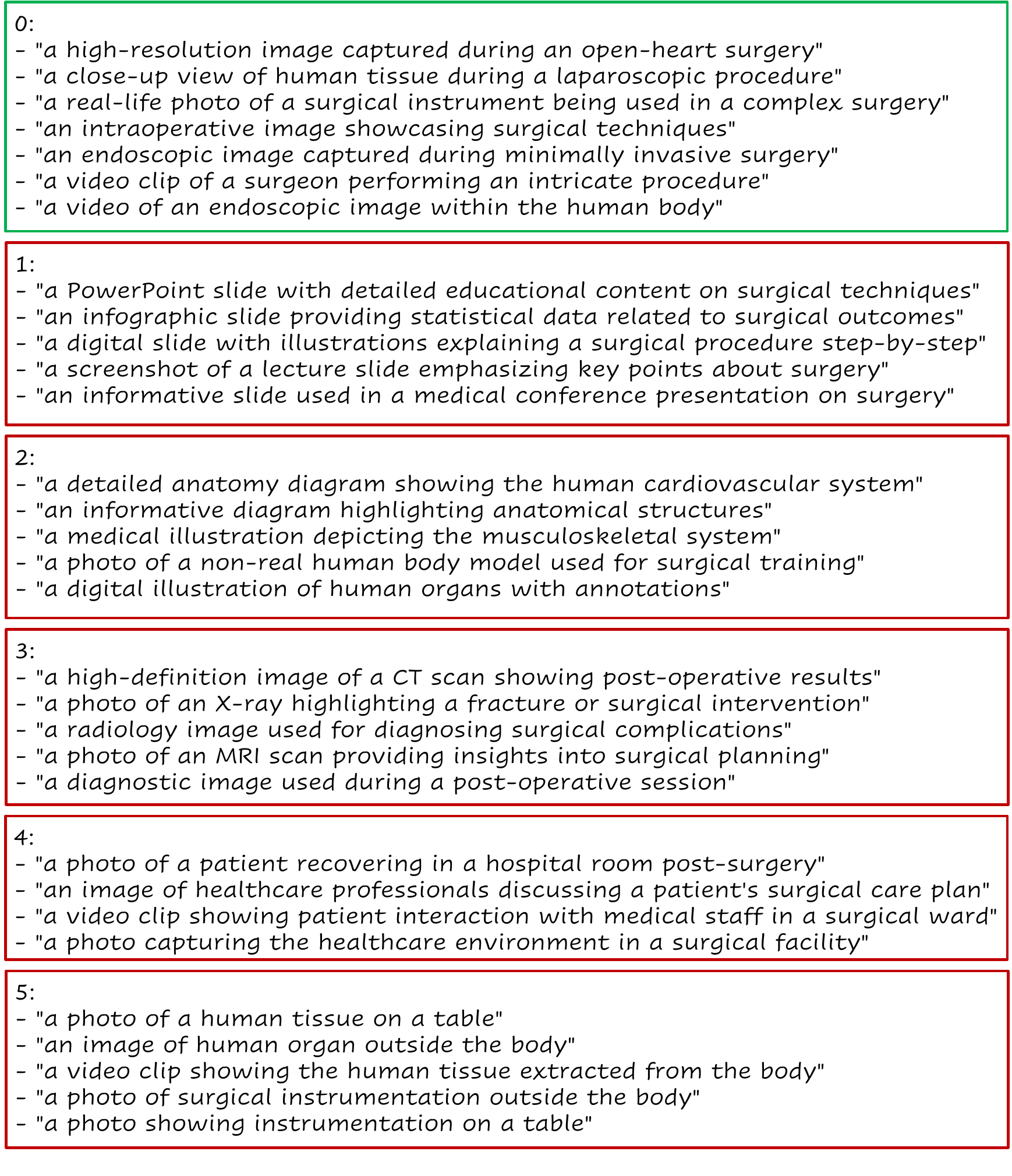}
    \caption{Textual prompts for zero-shot classification into \textit{surgical} (green) and \textit{non-surgical} (red) classes using SigLIP. Class embeddings are obtained by averaging multiple prompts.}
    \label{fig:zero-shot-classification-prompts}
\end{figure}

\begin{figure}[b!]
  \centering
    \includegraphics[width=0.45\textwidth]{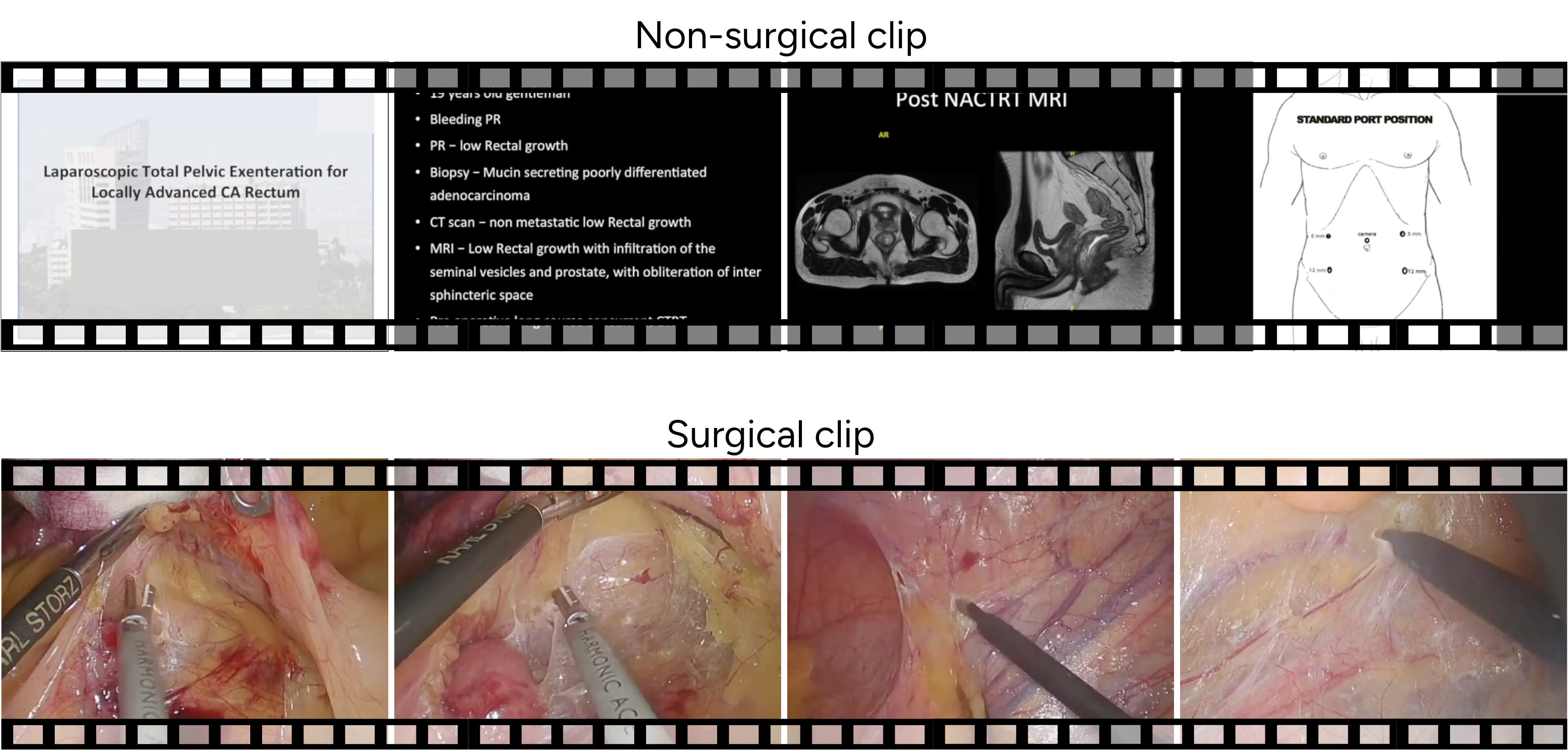}
    \caption{Examples of clips that are labeled as \textit{surgical} and \textit{non-surgical}. }
    \label{fig:clip_filtering_examples}
\end{figure}

\noindent \textbf{Stage 3: Dual-modal filtering.}  
At the clip level, we combined visual and textual checks to remove irrelevant or non-descriptive pairs.  
For visual filtering, we benchmarked three OpenCLIP models~\citep{open_clip}(ViT-G/14, ViT-SO400M/14, and ViT-SO400M-14-SigLIP-384) on 159 manually labeled clips (59 surgical, 100 non-surgical). We adopted SigLIP as it achieved the best results. We compared two strategies for clip classification: majority voting over sampled frames vs. mean-pooled frame embeddings. Majority voting yielded more robust results. With 24 frames per clip and a $>50\%$ voting threshold, we achieved 96.6 F1, 94.29 precision, and 99.0 recall for surgical classification. Figure~\ref{fig:zero-shot-classification-prompts} shows the prompt templates we used for surgical and non-surgical classes, Figure~\ref{fig:clip_filtering_examples} illustrates examples for both classes, and Table~\ref{tab:visual_confusion_matrix} portrays the confusion matrix of the visual filtering classification method chosen.  

\begin{table}[h]
\centering
\begin{minipage}{0.45\linewidth}
\centering
\caption{\textcolor{black}{Visual Filtering Confusion Matrix. (S): Surgical, (NS): Non-Surgical content}}
\small
\begin{tabular}{c|cc}
\toprule
GT $\backslash$ Pred & S & NS \\
\midrule
S      & 99  & 1  \\
NS  & 6   & 53 \\
\bottomrule
\end{tabular}
\label{tab:visual_confusion_matrix}
\end{minipage}
\hfill
\begin{minipage}{0.45\linewidth}
\centering
\caption{\textcolor{black}{Textual Filtering Confusion Matrix. (D) Descriptive, (ND): Non-Descriptive caption}}
\small
\begin{tabular}{c|cc}
\toprule
GT $\backslash$ Pred & D & ND \\
\midrule
D & 140 & 8  \\
ND   &  9  & 55 \\
\bottomrule
\end{tabular}
\label{tab:text_filtering_CM}
\end{minipage}
\end{table}

\begin{figure}[b!]
  \centering
    \includegraphics[width=0.45\textwidth]{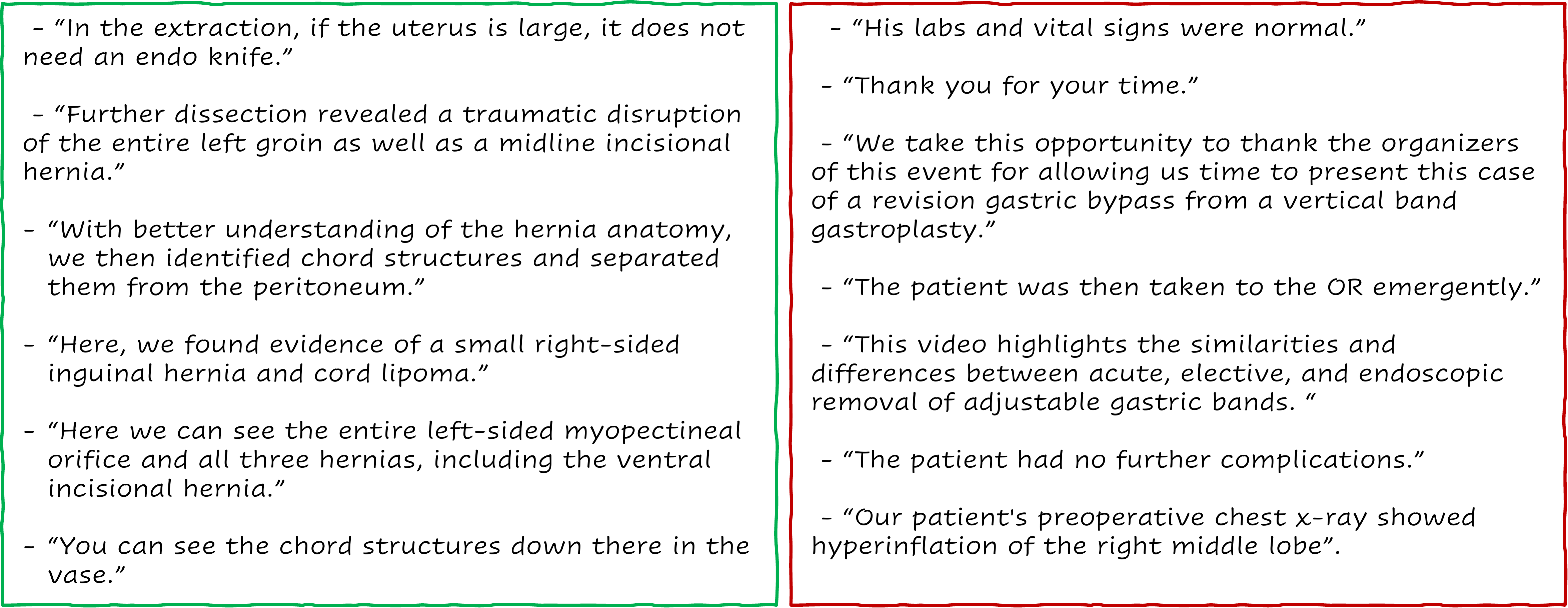}
    \caption{Examples of captions that are labeled as \textit{descriptive} (left) and \textit{non-descriptive} (right). }
    \label{fig:descriptive_filtering_examples}
\end{figure}

\begin{figure}[b!]
  \centering
    \includegraphics[width=0.45\textwidth]{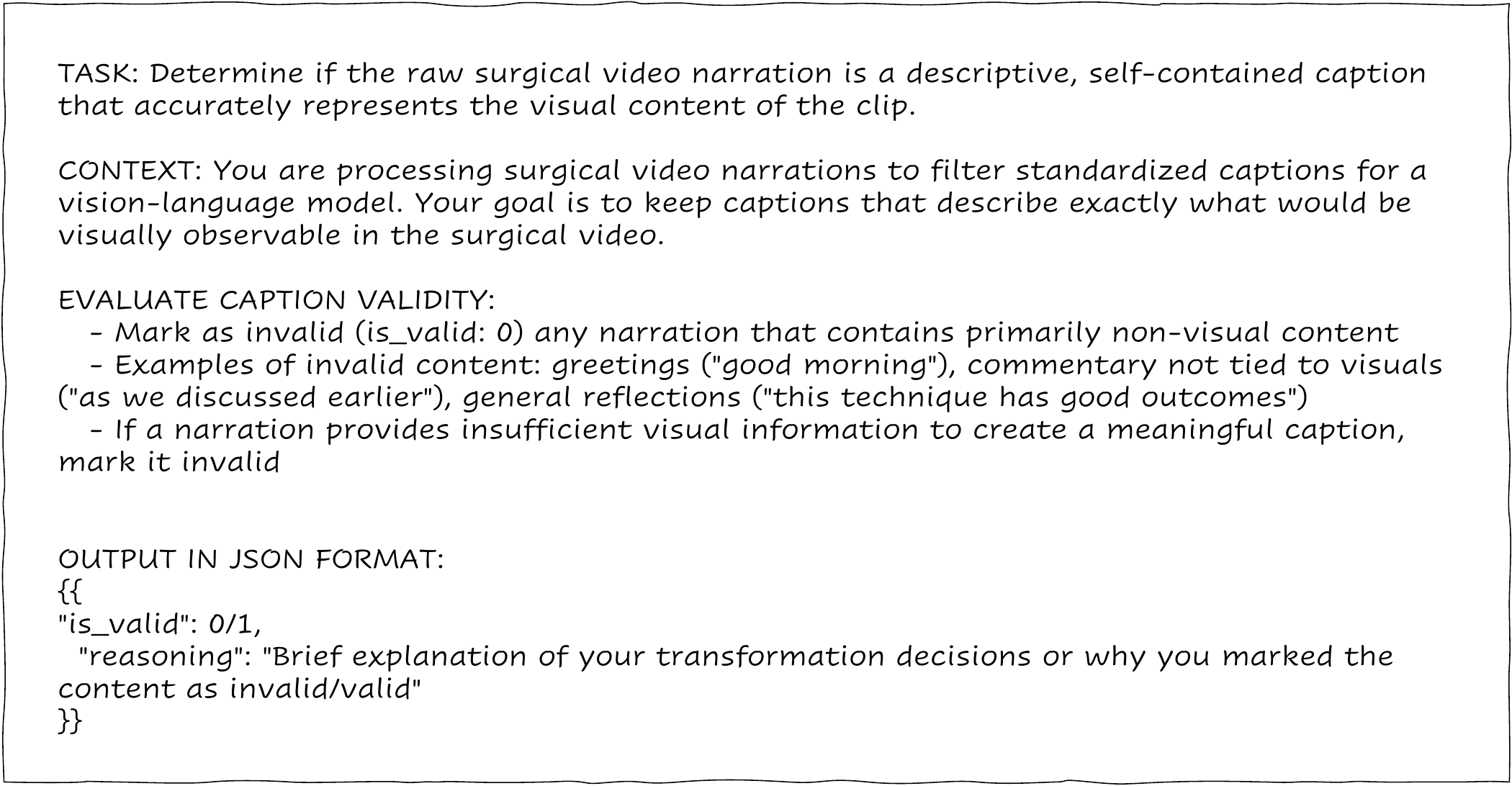}
    \caption{Prompt for caption descriptiveness filtering.}
    \label{fig:prompt_validity}
\end{figure}

\noindent  For textual filtering, we used GPT-4o with a binary decision prompt (Fig.~\ref{fig:prompt_validity}) to retain only descriptive captions; examples of descriptive and non-descriptive captions are displayed in Figure~\ref{fig:descriptive_filtering_examples}. \textcolor{black}{To evaluate textual filtering performance, we randomly selected a set of 212 captions from the full dataset, and an expert in surgical data science classified each as descriptive or non-descriptive of a surgical scene. Using these labels, we measured the LLM’s classification performance and computed the confusion matrix (Table \ref{tab:text_filtering_CM}), achieving 93.95 precision, 94.59 recall, and a 94.27 F1-score}. 

\begin{figure}[b!]
  \centering
    \includegraphics[width=0.40\textwidth]{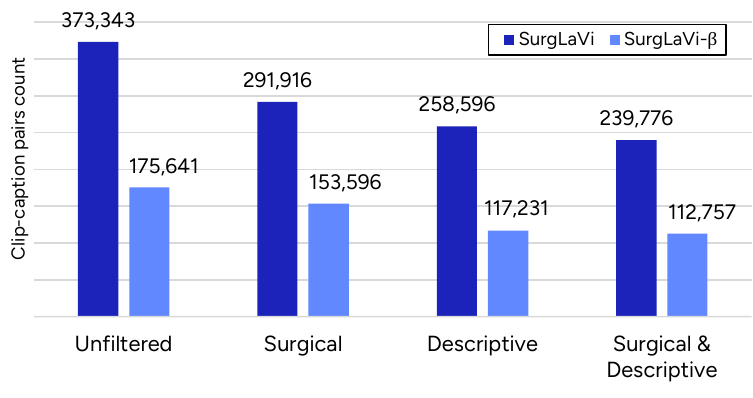}
    \caption{Filtering impact on SurgLaVi$-\mathbf{\beta}$ and SurgLaVi. Clip–caption pairs are progressively reduced through surgical filtering, descriptiveness filtering, and the intersection of both.}
    \label{fig:filtering_histogram}
\end{figure}

A pair was preserved if it passed both filters, and Figure~\ref{fig:filtering_histogram} portrays the raw count and the resulting counts after applying the visual and textual filters separately and in combination.

\begin{figure}[b!]
  \centering
    \includegraphics[width=0.45\textwidth]{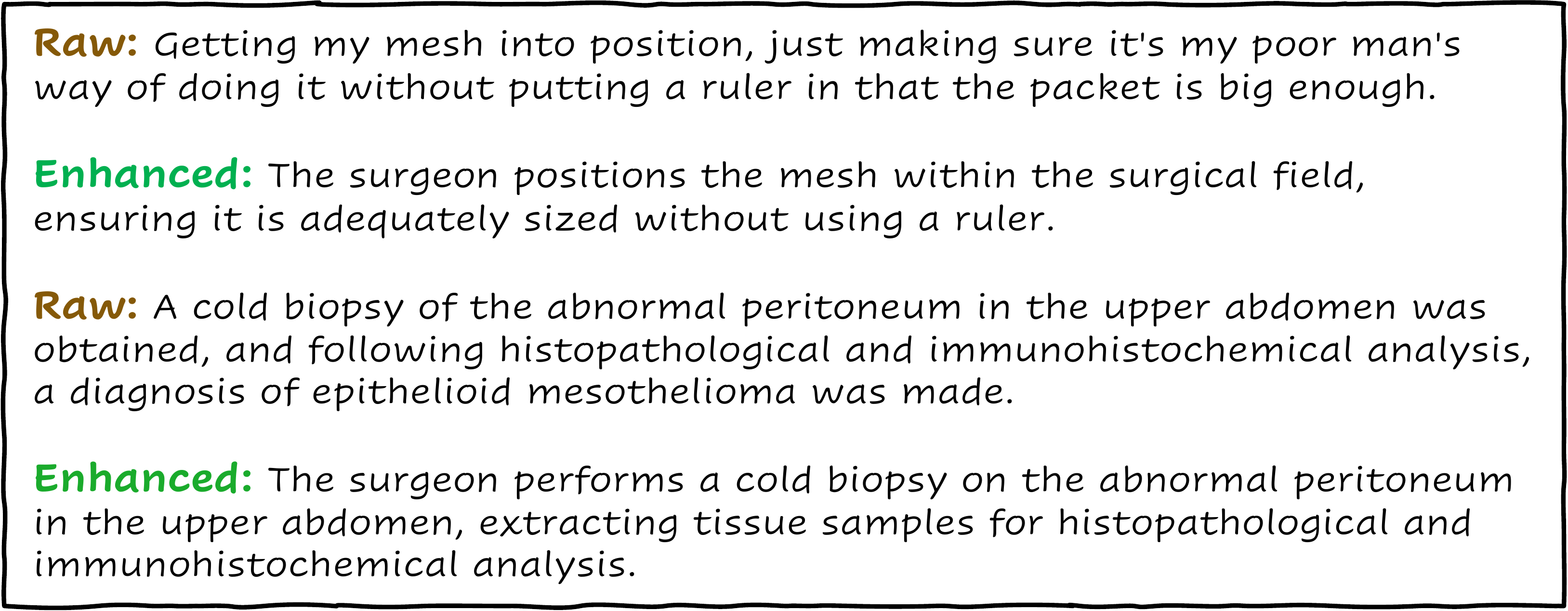}
    \caption{Examples raw and enhanced captions after the contextual enrichment stage}
    \label{fig:raw_enhanced_examples}
\end{figure}

\begin{figure}[b!]
  \centering
    \includegraphics[width=0.45\textwidth]{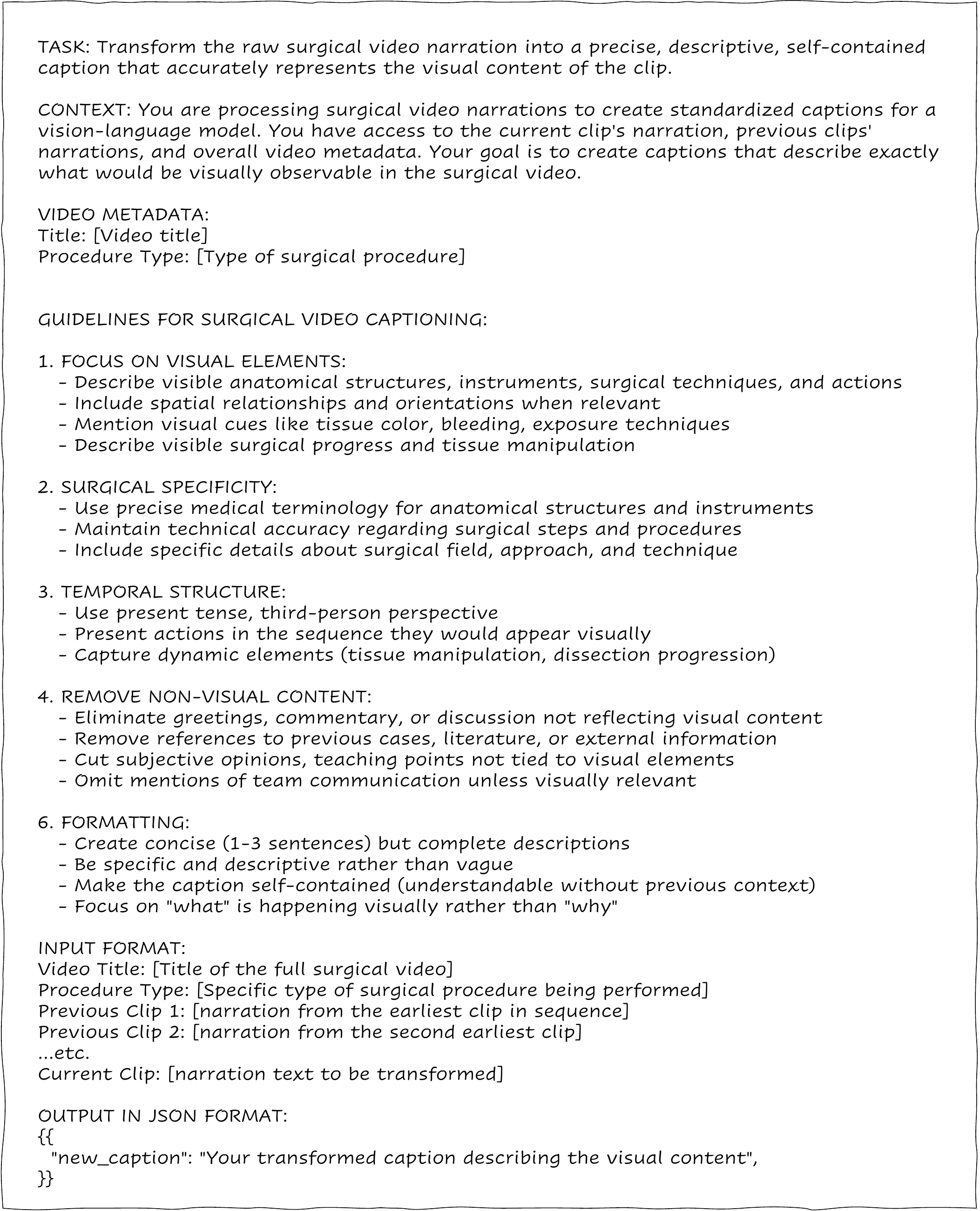}
    \caption{Prompt for contextual caption enrichment.}
    \label{fig:prompt_enrichment}
\end{figure}

\noindent \textbf{Stage 4: Contextual Caption Enrichment.}  
For each clip--caption pair $(x^g_{ij}, y^g_{ij})$, we utilize $\text{GPT-4o}$ using the prompt shown in Figure~\ref{fig:prompt_enrichment}
to instruct the model to generate an enriched caption $\hat{y}^g_{ij}$ conditioned on a fixed window of $N=5$ preceding captions at the same granularity level, the video title, and the surgical procedure type. Figure~\ref{fig:raw_enhanced_examples}
showcases examples of raw narrations along with their corresponding enhanced version.

\subsection{Pre-training Implementation Details}
\label{sec:surgclip_training_details}
\noindent We adopt a CLIP-style dual-encoder with a ViT-B/16 vision backbone (ImageNet-initialized) and a BERT-base text encoder. Outputs (768-d each) are projected into a shared 256-d multimodal space. A TimeSformer block encodes temporal dynamics over 16 frames.  
We optimize a video–text contrastive objective (InfoNCE) with AdamW (lr=$1\times 10^{-4}$, $\beta=(0.9,0.999)$, wd=0.02), cosine decay over 50 epochs, and batch size 312. We train the final model on a single NVIDIA H200 GPU. 

% --------------------------

\subsection{Linear Probing and Context Optimization}
\subsubsection{Linear Probing Details}
\label{sec:linear_probing_details}
\noindent We evaluate frozen visual features by concatenating the 768-dimensional backbone embedding with its 256-dimensional projection head output, yielding a 1024-dimensional representation. Linear classifiers are trained for 50 epochs with a batch size of 256. To ensure fairness across models, we tune all hyperparameters using Weights $\&$ Biases~\citep{wandb} on the validation set provided for each dataset, using single-frame inference for both SurgCLIP and all baselines. Once hyperparameters are selected, the final linear probes are trained using 32-frame temporal windows, where the mean embedding across the window represents the center labeled frame.
Our hyperparameter search considers learning rates ${10^{-3}, 10^{-4}}$, weight decays ${0, 0.01}$, and momentum ${0.9, 0.95}$. 

\subsubsection{CoOp Probing Details}
\label{sec:coop_probing_details}
\noindent CoOp probing follows the same frozen-feature evaluation protocol as linear probing, using the 1024-dimensional representation obtained by concatenating the 768-dimensional backbone embedding with its 256-dimensional projection head output. Unlike linear probing, CoOp optimizes only the learnable context tokens while keeping both vision and text encoders frozen. For all experiments, we optimize using fixed hyperparameters: SGD with momentum~0.9, a cosine learning rate scheduler, learning rate~0.01, and weight decay~0.01. Training runs for up to 500 epochs with early stopping if validation performance does not improve for 10 consecutive epochs. As in linear probing, validation sets provided with each dataset are used to monitor training stability and select the final model.

\noindent \textbf{Results.} Tables \ref{tab:cls_linear_probing}, \ref{tab:full_linear_probing}, and \ref{tab:coop_linear_probing} showcase CLS Linear Probing, Video Linear Probing, and Context Optimization results from Figure \ref{fig:linear_probing}. These tables present the exact F1 scores with their corresponding standard deviations, and additionally include the video-wise accuracy metric.

\begin{table*}[h!]
\centering
\caption{\textbf{CLS-Linear probing comparison across surgical benchmarks.} Model evaluation on phase and step recognition tasks using CLS-Linear probing with varying numbers of training samples per class (1, 2, 4, 8, 16). Reporting F1-score (F1) and accuracy (Acc) metrics with standard deviation across runs. SurgCLIP is trained on SurgLaVi dataset, while SurgCLIP$_{(\mathbf{\beta})}$ is trained on the open-source SurgLaVi$-\mathbf{\beta}$ subset. Best results are in \textbf{bold}, second-best are \underline{underlined}.}

\resizebox{0.9\textwidth}{!}{
\begin{tabular}{@{}cccccccc|cc@{}}
\toprule
Shots & Task & Dataset & Metric & CLIP & LEMON & PeskaVLP & MedSigLIP & SurgCLIP$_{(\mathbf{\beta})}$ & SurgCLIP \\ \midrule

\multirow{8}{*}{1} 
& \multirow{6}{*}{Phase}
& \multirow{2}{*}{AutoLaparo} 
& F1  & 22.40$_{\pm1.92}$ & 41.02$_{\pm8.12}$ & 38.54$_{\pm2.59}$ & 33.14$_{\pm3.43}$ & \textbf{52.29}$_{\pm5.28}$ & \underline{51.02}$_{\pm4.64}$ \\
& & & Acc & 35.96$_{\pm4.60}$ & 54.47$_{\pm9.05}$ & 49.74$_{\pm3.75}$ & 45.35$_{\pm2.99}$ & \textbf{66.89}$_{\pm5.35}$ & \underline{64.72}$_{\pm4.08}$ \\ \cmidrule(l){3-10}

& & \multirow{2}{*}{Cholec80} 
& F1  & 18.69$_{\pm4.74}$ & 26.83$_{\pm5.22}$ & 36.82$_{\pm7.59}$ & 27.06$_{\pm6.12}$ & \underline{44.39}$_{\pm6.40}$ & \textbf{47.60}$_{\pm5.19}$ \\
& & & Acc & 28.51$_{\pm5.54}$ & 34.78$_{\pm7.51}$ & 42.37$_{\pm7.63}$ & 33.74$_{\pm6.86}$ & \textbf{58.22}$_{\pm8.36}$ & \underline{57.82}$_{\pm6.62}$ \\ \cmidrule(l){3-10}

& & \multirow{2}{*}{GraSP} 
& F1  & 15.43$_{\pm3.93}$ & 28.41$_{\pm5.89}$ & 22.83$_{\pm3.96}$ & 23.97$_{\pm3.43}$ & \underline{28.85}$_{\pm5.67}$ & \textbf{29.50}$_{\pm5.60}$ \\
& & & Acc & 21.49$_{\pm3.64}$ & 33.68$_{\pm5.31}$ & 30.20$_{\pm3.89}$ & 29.08$_{\pm4.12}$ & \underline{34.70}$_{\pm4.64}$ & \textbf{35.27}$_{\pm3.92}$ \\ \cmidrule(l){2-10}

& \multirow{2}{*}{Step}
& \multirow{2}{*}{GraSP} 
& F1  & 7.69$_{\pm1.33}$ & 15.24$_{\pm2.34}$ & 13.29$_{\pm1.98}$ & 12.71$_{\pm2.39}$ & \textbf{16.92}$_{\pm2.63}$ & \underline{16.67}$_{\pm1.70}$ \\
& & & Acc & 10.54$_{\pm3.77}$ & 21.61$_{\pm3.07}$ & 17.68$_{\pm3.04}$ & 18.18$_{\pm3.44}$ & \underline{23.52}$_{\pm3.37}$ & \textbf{23.60}$_{\pm2.49}$ \\ \midrule

\multirow{8}{*}{2} 
& \multirow{6}{*}{Phase}
& \multirow{2}{*}{AutoLaparo} 
& F1  & 27.64$_{\pm4.03}$ & 52.41$_{\pm4.62}$ & 44.01$_{\pm5.13}$ & 42.13$_{\pm3.35}$ & \textbf{59.53}$_{\pm4.98}$ & \underline{58.06}$_{\pm4.63}$ \\
& & & Acc & 37.38$_{\pm4.09}$ & 63.82$_{\pm3.95}$ & 53.53$_{\pm3.23}$ & 52.36$_{\pm4.86}$ & \textbf{70.67}$_{\pm4.60}$ & \underline{70.11}$_{\pm4.46}$ \\ \cmidrule(l){3-10}

& & \multirow{2}{*}{Cholec80} 
& F1  & 24.33$_{\pm4.97}$ & 34.20$_{\pm5.61}$ & 48.45$_{\pm3.30}$ & 34.93$_{\pm5.12}$ & \textbf{57.49}$_{\pm4.08}$ & \underline{57.44}$_{\pm2.47}$ \\
& & & Acc & 31.74$_{\pm7.88}$ & 41.03$_{\pm6.45}$ & 50.45$_{\pm5.70}$ & 39.37$_{\pm5.82}$ & \textbf{67.32}$_{\pm4.11}$ & \underline{64.28}$_{\pm4.45}$ \\ \cmidrule(l){3-10}

& & \multirow{2}{*}{GraSP} 
& F1  & 18.04$_{\pm3.50}$ & 34.85$_{\pm2.24}$ & 26.19$_{\pm1.97}$ & 31.85$_{\pm2.55}$ & \underline{38.49}$_{\pm0.91}$ & \textbf{38.83}$_{\pm1.71}$ \\
& & & Acc & 25.27$_{\pm5.50}$ & 41.48$_{\pm3.57}$ & 32.11$_{\pm4.25}$ & 37.45$_{\pm5.36}$ & \underline{43.65}$_{\pm2.36}$ & \textbf{44.14}$_{\pm2.48}$ \\ \cmidrule(l){2-10}

& \multirow{2}{*}{Step}
& \multirow{2}{*}{GraSP} 
& F1  & 7.70$_{\pm1.65}$ & 17.49$_{\pm1.78}$ & 16.04$_{\pm2.10}$ & 16.57$_{\pm1.58}$ & \textbf{22.94}$_{\pm2.70}$ & \underline{22.62}$_{\pm2.48}$ \\
& & & Acc & 14.72$_{\pm6.05}$ & 23.35$_{\pm2.98}$ & 22.98$_{\pm4.29}$ & 24.60$_{\pm4.38}$ & \underline{29.83}$_{\pm4.78}$ & \textbf{30.25}$_{\pm2.95}$ \\ \midrule

\multirow{8}{*}{4} 
& \multirow{6}{*}{Phase}
& \multirow{2}{*}{AutoLaparo} 
& F1  & 30.33$_{\pm5.49}$ & 56.75$_{\pm6.77}$ & 49.35$_{\pm5.38}$ & 47.61$_{\pm3.88}$ & \textbf{65.67}$_{\pm5.16}$ & \underline{66.04}$_{\pm5.55}$ \\
& & & Acc & 37.31$_{\pm5.89}$ & 68.23$_{\pm3.74}$ & 58.85$_{\pm4.83}$ & 56.40$_{\pm3.88}$ & \underline{76.92}$_{\pm4.87}$ & \textbf{76.94}$_{\pm5.47}$ \\ \cmidrule(l){3-10}

& & \multirow{2}{*}{Cholec80} 
& F1  & 26.10$_{\pm2.98}$ & 43.28$_{\pm4.67}$ & 54.26$_{\pm1.88}$ & 43.39$_{\pm3.54}$ & \textbf{64.37}$_{\pm2.60}$ & \underline{64.36}$_{\pm2.41}$ \\
& & & Acc & 35.78$_{\pm4.73}$ & 47.62$_{\pm2.26}$ & 57.60$_{\pm4.65}$ & 50.16$_{\pm3.89}$ & \textbf{71.14}$_{\pm4.00}$ & \underline{69.69}$_{\pm3.29}$ \\ \cmidrule(l){3-10}

& & \multirow{2}{*}{GraSP} 
& F1  & 18.66$_{\pm3.64}$ & 38.16$_{\pm4.08}$ & 31.02$_{\pm5.30}$ & 32.10$_{\pm4.01}$ & \textbf{42.30}$_{\pm4.30}$ & \underline{42.00}$_{\pm4.05}$ \\
& & & Acc & 25.23$_{\pm4.52}$ & 46.42$_{\pm5.58}$ & 37.80$_{\pm6.21}$ & 38.84$_{\pm6.05}$ & \textbf{46.99}$_{\pm4.14}$ & \underline{46.44}$_{\pm3.44}$ \\ \cmidrule(l){2-10}

& \multirow{2}{*}{Step}
& \multirow{2}{*}{GraSP} 
& F1  & 8.23$_{\pm1.00}$ & 22.17$_{\pm3.55}$ & 19.78$_{\pm3.29}$ & 19.17$_{\pm1.27}$ & \underline{27.24}$_{\pm3.96}$ & \textbf{27.68}$_{\pm2.50}$ \\
& & & Acc & 14.17$_{\pm4.61}$ & 29.36$_{\pm6.56}$ & 26.92$_{\pm5.89}$ & 26.33$_{\pm3.52}$ & \textbf{35.19}$_{\pm5.97}$ & \underline{34.88}$_{\pm4.49}$ \\ \midrule

\multirow{8}{*}{8} 
& \multirow{6}{*}{Phase}
& \multirow{2}{*}{AutoLaparo} 
& F1  & 32.27$_{\pm3.35}$ & 63.02$_{\pm2.98}$ & 52.48$_{\pm3.00}$ & 50.14$_{\pm2.67}$ & \underline{71.16}$_{\pm2.83}$ & \textbf{71.62}$_{\pm1.92}$ \\
& & & Acc & 40.52$_{\pm3.43}$ & 73.57$_{\pm3.63}$ & 61.03$_{\pm3.15}$ & 59.69$_{\pm2.87}$ & \underline{80.90}$_{\pm2.07}$ & \textbf{81.41}$_{\pm1.39}$ \\ \cmidrule(l){3-10}

& & \multirow{2}{*}{Cholec80} 
& F1  & 28.38$_{\pm5.55}$ & 51.35$_{\pm2.54}$ & 57.65$_{\pm4.12}$ & 49.11$_{\pm2.75}$ & \underline{69.06}$_{\pm1.91}$ & \textbf{70.41}$_{\pm2.25}$ \\
& & & Acc & 45.18$_{\pm4.50}$ & 55.47$_{\pm5.31}$ & 59.71$_{\pm4.18}$ & 52.36$_{\pm3.30}$ & \underline{75.67}$_{\pm1.92}$ & \textbf{75.97}$_{\pm1.24}$ \\ \cmidrule(l){3-10}

& & \multirow{2}{*}{GraSP} 
& F1  & 18.97$_{\pm4.81}$ & 40.41$_{\pm2.97}$ & 34.98$_{\pm3.26}$ & 35.04$_{\pm4.22}$ & \textbf{48.36}$_{\pm2.69}$ & \underline{47.70}$_{\pm2.97}$ \\
& & & Acc & 25.79$_{\pm5.52}$ & 47.61$_{\pm4.87}$ & 41.33$_{\pm2.96}$ & 41.64$_{\pm7.48}$ & \textbf{53.48}$_{\pm2.72}$ & \underline{52.55}$_{\pm3.88}$ \\ \cmidrule(l){2-10}

& \multirow{2}{*}{Step}
& \multirow{2}{*}{GraSP} 
& F1  & 10.54$_{\pm2.27}$ & 21.31$_{\pm1.09}$ & 22.00$_{\pm0.80}$ & 22.23$_{\pm1.76}$ & \underline{29.85}$_{\pm1.82}$ & \textbf{29.96}$_{\pm1.06}$ \\
& & & Acc & 17.43$_{\pm3.26}$ & 28.79$_{\pm3.31}$ & 30.11$_{\pm2.10}$ & 31.83$_{\pm1.55}$ & \underline{38.51}$_{\pm2.74}$ & \textbf{39.29}$_{\pm2.44}$ \\ \midrule

\multirow{8}{*}{16} 
& \multirow{6}{*}{Phase}
& \multirow{2}{*}{AutoLaparo} 
& F1  & 36.28$_{\pm5.77}$ & 68.33$_{\pm4.84}$ & 54.85$_{\pm3.22}$ & 55.97$_{\pm3.44}$ & \underline{73.15}$_{\pm1.69}$ & \textbf{73.85}$_{\pm1.71}$ \\
& & & Acc & 45.88$_{\pm4.24}$ & 78.10$_{\pm4.71}$ & 63.66$_{\pm2.64}$ & 66.52$_{\pm3.92}$ & \underline{82.89}$_{\pm1.84}$ & \textbf{83.00}$_{\pm1.66}$ \\ \cmidrule(l){3-10}

& & \multirow{2}{*}{Cholec80} 
& F1  & 32.09$_{\pm4.85}$ & 57.86$_{\pm2.35}$ & 62.90$_{\pm2.61}$ & 55.96$_{\pm3.37}$ & \textbf{73.12}$_{\pm1.15}$ & \underline{72.99}$_{\pm1.24}$ \\
& & & Acc & 43.30$_{\pm4.68}$ & 63.46$_{\pm3.34}$ & 66.25$_{\pm4.17}$ & 59.95$_{\pm2.81}$ & \textbf{79.56}$_{\pm1.25}$ & \underline{78.99}$_{\pm1.15}$ \\ \cmidrule(l){3-10}

& & \multirow{2}{*}{GraSP} 
& F1  & 23.36$_{\pm2.89}$ & 42.37$_{\pm1.72}$ & 38.29$_{\pm3.50}$ & 39.09$_{\pm2.20}$ & \underline{52.03}$_{\pm3.07}$ & \textbf{52.74}$_{\pm2.04}$ \\
& & & Acc & 32.11$_{\pm2.83}$ & 51.13$_{\pm1.34}$ & 45.88$_{\pm4.32}$ & 47.29$_{\pm2.44}$ & \underline{57.62}$_{\pm3.75}$ & \textbf{59.05}$_{\pm2.18}$ \\ \cmidrule(l){2-10}

& \multirow{2}{*}{Step}
& \multirow{2}{*}{GraSP} 
& F1  & 14.39$_{\pm1.09}$ & 25.32$_{\pm2.32}$ & 25.41$_{\pm2.29}$ & 25.20$_{\pm1.54}$ & \textbf{33.47}$_{\pm1.85}$ & \underline{33.25}$_{\pm1.72}$ \\
& & & Acc & 23.72$_{\pm3.88}$ & 35.19$_{\pm3.99}$ & 34.79$_{\pm4.75}$ & 34.11$_{\pm3.87}$ & \underline{43.06}$_{\pm3.29}$ & \textbf{43.56}$_{\pm2.52}$ \\ \bottomrule

\end{tabular}
}
\label{tab:cls_linear_probing}
\end{table*}
\begin{table*}[h!]
\centering
\caption{\textbf{Few-shot Context Optimization (CoOp) comparison across surgical benchmarks.} Model evaluation on phase and step recognition tasks using CoOp linear probing with varying numbers of training shots (1, 2, 4, 8, 16). Reporting F1-score (F1) and accuracy (Acc) metrics with standard deviation across runs. SurgCLIP is trained on SurgLaVi dataset, while SurgCLIP$_{(\mathbf{\beta})}$ is trained on the open-source SurgLaVi$-\mathbf{\beta}$ subset. Best results are in \textbf{bold}, second-best are \underline{underlined}.}
\resizebox{0.8\textwidth}{!}{
\begin{tabular}{@{}ccccccc|cc@{}}
\toprule
Shots & Task & Dataset & Metric & CLIP & PeskaVLP & MedSigLIP & SurgCLIP$_{(\mathbf{\beta})}$ & SurgCLIP \\ \midrule

\multirow{8}{*}{1} 
& \multirow{6}{*}{Phase}
& \multirow{2}{*}{AutoLaparo} 
& F1  & 20.54$_{\pm2.87}$ & 38.30$_{\pm3.97}$ & 29.49$_{\pm5.64}$ & \textbf{44.99}$_{\pm7.38}$ & \underline{42.13}$_{\pm14.37}$ \\
& & & Acc & 29.37$_{\pm5.36}$ & 48.05$_{\pm4.44}$ & 40.00$_{\pm8.37}$ & \textbf{54.84}$_{\pm8.51}$ & \underline{49.85}$_{\pm16.47}$ \\ \cmidrule(l){3-9}

& & \multirow{2}{*}{Cholec80} 
& F1  & 21.41$_{\pm3.31}$ & 42.01$_{\pm3.31}$ & 28.95$_{\pm4.24}$ & \underline{46.10}$_{\pm11.29}$ & \textbf{51.49}$_{\pm8.13}$ \\
& & & Acc & 37.24$_{\pm6.68}$ & 47.92$_{\pm5.92}$ & 31.90$_{\pm6.79}$ & \underline{55.01}$_{\pm8.78}$ & \textbf{62.57}$_{\pm3.81}$ \\ \cmidrule(l){3-9}

& & \multirow{2}{*}{GraSP} 
& F1  & 16.72$_{\pm3.60}$ & 21.70$_{\pm2.53}$ & 20.56$_{\pm2.07}$ & \textbf{27.77}$_{\pm6.71}$ & \underline{25.21}$_{\pm4.31}$ \\
& & & Acc & 23.84$_{\pm3.49}$ & 26.99$_{\pm2.75}$ & 25.42$_{\pm3.98}$ & \textbf{32.25}$_{\pm6.68}$ & \underline{30.27}$_{\pm5.19}$ \\ \cmidrule(l){2-9}

& \multirow{2}{*}{Step}
& \multirow{2}{*}{GraSP} 
& F1  & 7.45$_{\pm0.59}$ & 11.53$_{\pm1.51}$ & 12.52$_{\pm1.64}$ & \textbf{14.93}$_{\pm1.85}$ & \underline{14.92}$_{\pm1.52}$ \\
& & & Acc & 13.91$_{\pm1.85}$ & 17.85$_{\pm1.24}$ & 16.33$_{\pm2.92}$ & \textbf{20.75}$_{\pm3.72}$ & \underline{20.20}$_{\pm3.29}$ \\ \midrule

\multirow{8}{*}{2} 
& \multirow{6}{*}{Phase}
& \multirow{2}{*}{AutoLaparo} 
& F1  & 27.92$_{\pm2.56}$ & 43.15$_{\pm1.64}$ & 42.01$_{\pm3.08}$ & \underline{50.57}$_{\pm4.85}$ & \textbf{53.27}$_{\pm6.05}$ \\
& & & Acc & 37.51$_{\pm2.65}$ & 53.80$_{\pm2.21}$ & 52.81$_{\pm4.48}$ & \underline{60.39}$_{\pm5.06}$ & \textbf{63.45}$_{\pm5.04}$ \\ \cmidrule(l){3-9}

& & \multirow{2}{*}{Cholec80} 
& F1  & 25.80$_{\pm4.34}$ & 40.95$_{\pm7.85}$ & 39.12$_{\pm5.37}$ & \underline{48.78}$_{\pm7.25}$ & \textbf{54.68}$_{\pm4.07}$ \\
& & & Acc & 37.00$_{\pm9.09}$ & 46.25$_{\pm8.45}$ & 44.50$_{\pm4.03}$ & \underline{55.58}$_{\pm9.76}$ & \textbf{61.87}$_{\pm1.66}$ \\ \cmidrule(l){3-9}

& & \multirow{2}{*}{GraSP} 
& F1  & 19.32$_{\pm3.39}$ & 25.49$_{\pm2.04}$ & 29.55$_{\pm3.16}$ & \textbf{35.25}$_{\pm3.08}$ & \underline{33.63}$_{\pm4.27}$ \\
& & & Acc & 26.70$_{\pm6.34}$ & 32.07$_{\pm4.14}$ & 34.97$_{\pm5.11}$ & \textbf{38.76}$_{\pm3.56}$ & \underline{38.05}$_{\pm4.81}$ \\ \cmidrule(l){2-9}

& \multirow{2}{*}{Step}
& \multirow{2}{*}{GraSP} 
& F1  & 9.11$_{\pm0.91}$ & 14.61$_{\pm1.04}$ & 15.50$_{\pm2.00}$ & \textbf{19.83}$_{\pm1.53}$ & \underline{18.24}$_{\pm1.65}$ \\
& & & Acc & 13.15$_{\pm0.49}$ & 22.73$_{\pm1.70}$ & 21.82$_{\pm3.65}$ & \textbf{25.06}$_{\pm3.97}$ & \underline{24.24}$_{\pm3.41}$ \\ \midrule

\multirow{8}{*}{4} 
& \multirow{6}{*}{Phase}
& \multirow{2}{*}{AutoLaparo} 
& F1  & 30.83$_{\pm5.94}$ & 43.63$_{\pm5.37}$ & 45.33$_{\pm5.41}$ & \underline{57.02}$_{\pm3.14}$ & \textbf{59.94}$_{\pm3.03}$ \\
& & & Acc & 39.99$_{\pm5.87}$ & 54.87$_{\pm5.03}$ & 54.69$_{\pm7.52}$ & \underline{69.15}$_{\pm1.90}$ & \textbf{70.96}$_{\pm2.35}$ \\ \cmidrule(l){3-9}

& & \multirow{2}{*}{Cholec80} 
& F1  & 28.23$_{\pm3.88}$ & 51.07$_{\pm3.98}$ & 41.01$_{\pm3.96}$ & \underline{58.19}$_{\pm4.35}$ & \textbf{59.42}$_{\pm7.72}$ \\
& & & Acc & 42.32$_{\pm4.03}$ & 56.35$_{\pm5.59}$ & 44.53$_{\pm3.32}$ & \textbf{65.73}$_{\pm4.69}$ & \underline{65.53}$_{\pm9.49}$ \\ \cmidrule(l){3-9}

& & \multirow{2}{*}{GraSP} 
& F1  & 21.03$_{\pm2.08}$ & 28.88$_{\pm2.93}$ & 34.31$_{\pm2.96}$ & \underline{36.88}$_{\pm4.20}$ & \textbf{38.40}$_{\pm1.94}$ \\
& & & Acc & 29.12$_{\pm2.39}$ & 34.82$_{\pm4.45}$ & 40.88$_{\pm1.99}$ & \underline{42.83}$_{\pm3.09}$ & \textbf{45.19}$_{\pm3.22}$ \\ \cmidrule(l){2-9}

& \multirow{2}{*}{Step}
& \multirow{2}{*}{GraSP} 
& F1  & 9.65$_{\pm0.91}$ & 15.85$_{\pm2.51}$ & 18.23$_{\pm1.50}$ & \textbf{22.31}$_{\pm2.02}$ & \underline{21.32}$_{\pm1.29}$ \\
& & & Acc & 15.66$_{\pm4.27}$ & 21.42$_{\pm4.22}$ & 24.31$_{\pm3.71}$ & \textbf{29.99}$_{\pm3.01}$ & \underline{26.49}$_{\pm1.48}$ \\ \midrule

\multirow{8}{*}{8} 
& \multirow{6}{*}{Phase}
& \multirow{2}{*}{AutoLaparo} 
& F1  & 33.47$_{\pm2.78}$ & 49.16$_{\pm2.97}$ & 54.32$_{\pm2.45}$ & \underline{65.06}$_{\pm4.00}$ & \textbf{66.84}$_{\pm3.00}$ \\
& & & Acc & 43.19$_{\pm2.63}$ & 59.71$_{\pm4.04}$ & 63.58$_{\pm2.26}$ & \underline{74.32}$_{\pm2.44}$ & \textbf{77.24}$_{\pm2.63}$ \\ \cmidrule(l){3-9}

& & \multirow{2}{*}{Cholec80} 
& F1  & 32.00$_{\pm3.10}$ & 56.29$_{\pm3.31}$ & 49.89$_{\pm2.47}$ & \textbf{66.27}$_{\pm2.94}$ & \underline{65.99}$_{\pm2.26}$ \\
& & & Acc & 40.52$_{\pm6.02}$ & 61.86$_{\pm6.49}$ & 53.67$_{\pm2.02}$ & \textbf{73.36}$_{\pm4.19}$ & \underline{71.57}$_{\pm3.09}$ \\ \cmidrule(l){3-9}

& & \multirow{2}{*}{GraSP} 
& F1  & 25.12$_{\pm1.63}$ & 31.83$_{\pm1.96}$ & 37.98$_{\pm2.69}$ & \textbf{41.64}$_{\pm2.31}$ & \underline{39.04}$_{\pm2.66}$ \\
& & & Acc & 35.08$_{\pm2.14}$ & 38.52$_{\pm2.22}$ & 45.23$_{\pm2.44}$ & \textbf{47.94}$_{\pm1.24}$ & \underline{43.00}$_{\pm2.39}$ \\ \cmidrule(l){2-9}

& \multirow{2}{*}{Step}
& \multirow{2}{*}{GraSP} 
& F1  & 11.65$_{\pm2.27}$ & 17.80$_{\pm1.77}$ & \underline{23.17}$_{\pm0.82}$ & \textbf{24.26}$_{\pm2.32}$ & 22.17$_{\pm1.28}$ \\
& & & Acc & 18.13$_{\pm4.71}$ & 27.48$_{\pm3.58}$ & \underline{33.19}$_{\pm4.32}$ & \textbf{33.73}$_{\pm1.82}$ & 28.01$_{\pm1.77}$ \\ \midrule

\multirow{8}{*}{16} 
& \multirow{6}{*}{Phase}
& \multirow{2}{*}{AutoLaparo} 
& F1  & 37.50$_{\pm2.49}$ & 50.67$_{\pm1.42}$ & 54.13$_{\pm3.90}$ & \underline{66.06}$_{\pm5.27}$ & \textbf{67.47}$_{\pm3.19}$ \\
& & & Acc & 45.97$_{\pm3.78}$ & 61.44$_{\pm1.50}$ & 63.34$_{\pm3.74}$ & \underline{75.76}$_{\pm5.33}$ & \textbf{78.70}$_{\pm2.91}$ \\ \cmidrule(l){3-9}

& & \multirow{2}{*}{Cholec80} 
& F1  & 35.23$_{\pm5.08}$ & 60.32$_{\pm2.76}$ & 56.62$_{\pm3.28}$ & \textbf{69.16}$_{\pm1.18}$ & \underline{68.07}$_{\pm2.13}$ \\
& & & Acc & 44.77$_{\pm4.89}$ & 65.77$_{\pm1.70}$ & 61.01$_{\pm2.67}$ & \textbf{75.51}$_{\pm2.37}$ & \underline{74.63}$_{\pm2.45}$ \\ \cmidrule(l){3-9}

& & \multirow{2}{*}{GraSP} 
& F1  & 23.83$_{\pm2.54}$ & 34.06$_{\pm3.10}$ & 42.13$_{\pm1.58}$ & \textbf{46.47}$_{\pm3.35}$ & \underline{43.61}$_{\pm1.34}$ \\
& & & Acc & 34.02$_{\pm3.53}$ & 41.26$_{\pm3.85}$ & 49.24$_{\pm1.66}$ & \textbf{51.24}$_{\pm4.42}$ & \underline{46.79}$_{\pm2.20}$ \\ \cmidrule(l){2-9}

& \multirow{2}{*}{Step}
& \multirow{2}{*}{GraSP} 
& F1  & 11.78$_{\pm0.69}$ & 20.95$_{\pm1.00}$ & \textbf{26.50}$_{\pm1.07}$ & \underline{25.70}$_{\pm3.20}$ & 24.26$_{\pm1.00}$ \\
& & & Acc & 18.58$_{\pm2.43}$ & 28.45$_{\pm1.63}$ & \textbf{35.68}$_{\pm2.69}$ & \underline{33.59}$_{\pm5.02}$ & 30.95$_{\pm2.75}$ \\ \bottomrule

\end{tabular}
}
\label{tab:coop_linear_probing}
\end{table*}
\begin{table*}[h!]
\centering
\caption{\textbf{Video few-/full-shot linear probing comparison across surgical benchmarks.} Model evaluation on phase and step recognition tasks using few and full linear probing with 10\%, 50\%, and 100\% of the available training videos of each dataset. Reporting F1-score (F1) and accuracy (Acc) metrics with standard deviation across runs. SurgCLIP is trained on SurgLaVi dataset, while SurgCLIP$_{(\mathbf{\beta})}$ is trained on the open-source SurgLaVi$-\mathbf{\beta}$ subset. Best results are in \textbf{bold}, second-best are \underline{underlined}.}
\resizebox{0.7\textwidth}{!}{
\begin{tabular}{@{}cccccccc|cc@{}}
\toprule
Shots & Task & Dataset & Metric & CLIP & LEMON & PeskaVLP & MedSigLIP & SurgCLIP$_{(\mathbf{\beta})}$ & SurgCLIP \\ \midrule

\multirow{8}{*}{10\%} 
& \multirow{6}{*}{Phase}
& \multirow{2}{*}{AutoLaparo} 
& F1  & 30.89$_{\pm8.45}$ & 52.20$_{\pm11.15}$ & 46.15$_{\pm8.22}$ & 40.58$_{\pm8.59}$ & \underline{56.19}$_{\pm12.66}$ & \textbf{57.70}$_{\pm11.00}$ \\
& & & Acc & 44.83$_{\pm6.61}$ & 68.14$_{\pm12.02}$ & 58.04$_{\pm9.85}$ & 55.34$_{\pm8.66}$ & \underline{70.12}$_{\pm15.44}$ & \textbf{71.27}$_{\pm12.95}$ \\ \cmidrule(l){3-10}

& & \multirow{2}{*}{Cholec80} 
& F1  & 40.52$_{\pm4.29}$ & 60.49$_{\pm5.09}$ & 61.27$_{\pm3.62}$ & 60.05$_{\pm3.18}$ & \underline{70.24}$_{\pm2.01}$ & \textbf{73.02}$_{\pm2.26}$ \\
& & & Acc & 56.02$_{\pm4.25}$ & 72.43$_{\pm1.98}$ & 68.29$_{\pm1.66}$ & 69.24$_{\pm1.60}$ & \underline{79.43}$_{\pm1.12}$ & \textbf{80.16}$_{\pm1.53}$ \\ \cmidrule(l){3-10}

& & \multirow{2}{*}{GraSP} 
& F1  & 23.10$_{\pm4.61}$ & 37.07$_{\pm6.30}$ & 28.20$_{\pm3.11}$ & 30.96$_{\pm7.55}$ & \underline{43.29}$_{\pm3.90}$ & \textbf{46.17}$_{\pm4.41}$ \\
& & & Acc & 39.60$_{\pm2.99}$ & 53.95$_{\pm4.08}$ & 43.51$_{\pm2.49}$ & 48.61$_{\pm7.28}$ & \underline{54.31}$_{\pm3.21}$ & \textbf{57.21}$_{\pm4.46}$ \\ \cmidrule(l){2-10}

& \multirow{2}{*}{Step}
& \multirow{2}{*}{GraSP} 
& F1  & 10.54$_{\pm1.73}$ & 16.23$_{\pm3.98}$ & 16.51$_{\pm2.85}$ & 17.58$_{\pm3.36}$ & \textbf{23.85}$_{\pm2.46}$ & \underline{23.60}$_{\pm3.70}$ \\
& & & Acc & 31.56$_{\pm4.01}$ & 41.70$_{\pm6.61}$ & 34.59$_{\pm3.56}$ & 41.11$_{\pm4.00}$ & \underline{43.63}$_{\pm2.53}$ & \textbf{44.03}$_{\pm4.40}$ \\ \midrule

\multirow{8}{*}{50\%} 
& \multirow{6}{*}{Phase}
& \multirow{2}{*}{AutoLaparo} 
& F1  & 43.29$_{\pm3.35}$ & 66.72$_{\pm2.73}$ & 56.45$_{\pm1.95}$ & 57.30$_{\pm2.62}$ & \underline{70.42}$_{\pm0.75}$ & \textbf{70.79}$_{\pm1.18}$ \\
& & & Acc & 56.29$_{\pm1.33}$ & 80.04$_{\pm2.29}$ & 68.90$_{\pm2.37}$ & 71.01$_{\pm2.40}$ & \underline{82.69}$_{\pm0.66}$ & \textbf{83.47}$_{\pm0.87}$ \\ \cmidrule(l){3-10}

& & \multirow{2}{*}{Cholec80} 
& F1  & 59.23$_{\pm2.49}$ & 72.62$_{\pm2.62}$ & 69.81$_{\pm2.46}$ & 71.83$_{\pm1.24}$ & \underline{77.40}$_{\pm2.24}$ & \textbf{79.00}$_{\pm1.23}$ \\
& & & Acc & 69.73$_{\pm1.90}$ & 81.01$_{\pm1.44}$ & 75.75$_{\pm1.46}$ & 78.16$_{\pm1.06}$ & \underline{84.47}$_{\pm1.11}$ & \textbf{85.35}$_{\pm0.71}$ \\ \cmidrule(l){3-10}

& & \multirow{2}{*}{GraSP} 
& F1  & 36.78$_{\pm2.92}$ & 45.76$_{\pm3.36}$ & 39.46$_{\pm0.90}$ & 43.32$_{\pm2.63}$ & \underline{55.44}$_{\pm1.78}$ & \textbf{56.27}$_{\pm2.62}$ \\
& & & Acc & 50.58$_{\pm3.02}$ & 60.64$_{\pm2.33}$ & 54.10$_{\pm1.41}$ & 58.83$_{\pm2.01}$ & \underline{64.83}$_{\pm1.71}$ & \textbf{65.67}$_{\pm2.45}$ \\ \cmidrule(l){2-10}

& \multirow{2}{*}{Step}
& \multirow{2}{*}{GraSP} 
& F1  & 18.88$_{\pm0.79}$ & 25.10$_{\pm2.81}$ & 22.79$_{\pm0.78}$ & 27.50$_{\pm0.96}$ & \underline{33.60}$_{\pm0.54}$ & \textbf{33.83}$_{\pm1.65}$ \\
& & & Acc & 39.26$_{\pm1.11}$ & 50.61$_{\pm1.31}$ & 42.81$_{\pm1.23}$ & 49.10$_{\pm3.20}$ & \textbf{53.41}$_{\pm1.38}$ & \underline{51.92}$_{\pm2.06}$ \\ \midrule

\multirow{8}{*}{100\%} 
& \multirow{6}{*}{Phase}
& \multirow{2}{*}{AutoLaparo} 
& F1  & 50.09 & 69.39 & 59.44 & 62.60 & \textbf{70.98} & \underline{70.47} \\
& & & Acc & 63.01 & 83.04 & 70.19 & 76.64 & \underline{83.24} & \textbf{84.99} \\ \cmidrule(l){3-10}

& & \multirow{2}{*}{Cholec80} 
& F1  & 63.84 & 74.47 & 71.18 & 73.61 & \underline{79.55} & \textbf{80.46} \\
& & & Acc & 73.04 & 82.71 & 77.39 & 79.77 & \underline{86.24} & \textbf{86.45} \\ \cmidrule(l){3-10}

& & \multirow{2}{*}{GraSP} 
& F1  & 39.84 & 51.53 & 45.17 & 50.65 & \underline{58.89} & \textbf{61.08} \\
& & & Acc & 55.97 & 64.92 & 60.38 & 65.03 & \underline{68.76} & \textbf{70.26} \\ \cmidrule(l){2-10}

& \multirow{2}{*}{Step}
& \multirow{2}{*}{GraSP} 
& F1  & 22.13 & 28.37 & 27.32 & 31.33 & \underline{35.48} & \textbf{38.56} \\
& & & Acc & 44.38 & 53.48 & 48.21 & 54.94 & \textbf{58.11} & \underline{55.93} \\ \bottomrule

\end{tabular}
}
\label{tab:full_linear_probing}
\end{table*}

\begin{figure*}[t]
  \centering
\includegraphics[width=0.5\textwidth]{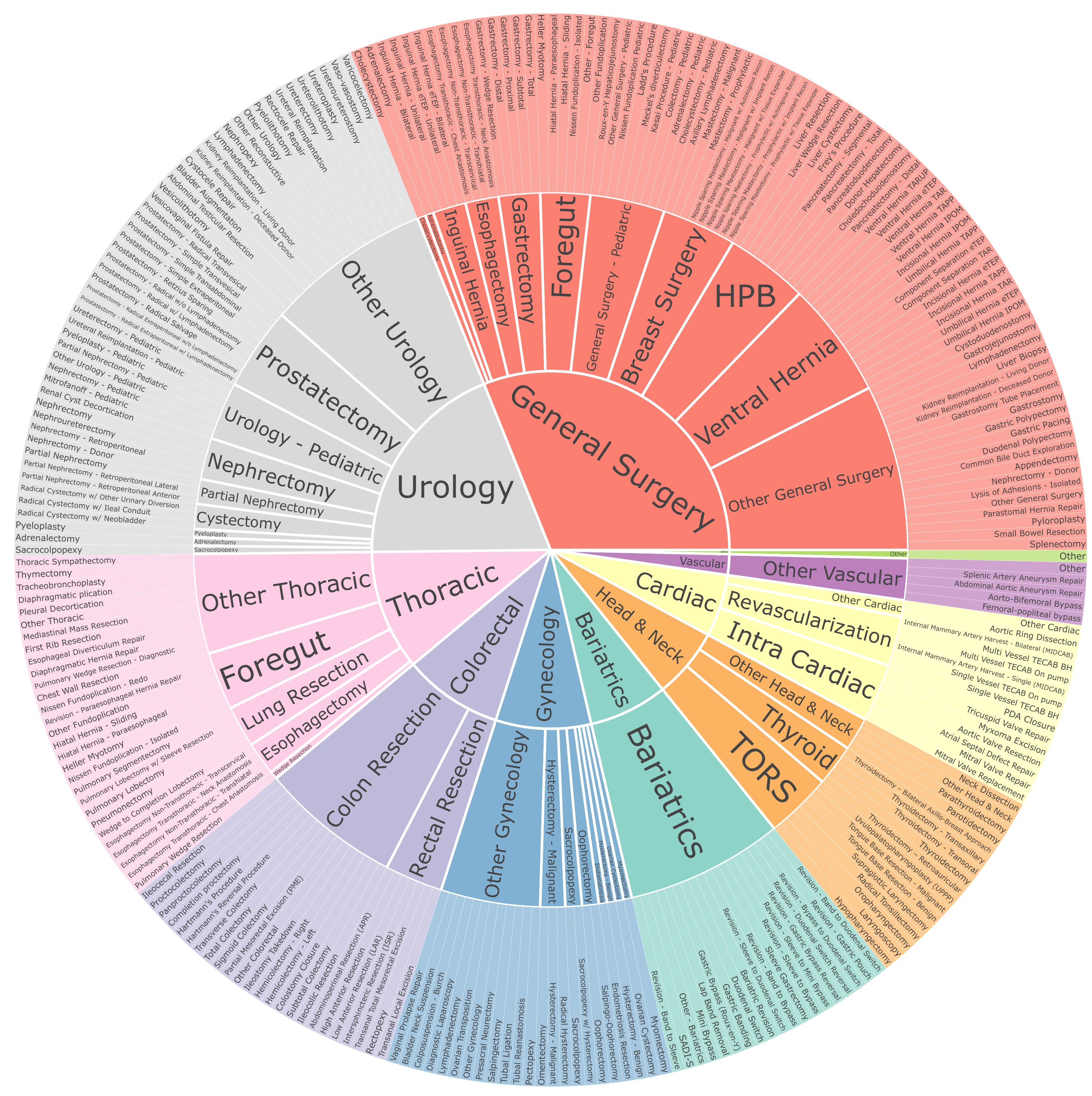}
    \caption{\textbf{Surgical taxonomy.} Sunburst visualization of hierarchical classification from specialties (inner ring) to subjects (middle) and procedures (outer). Wedge size is proportional to procedure count.}
    \label{taxonomy}
\end{figure*}

% --------------------------

\subsection{Temporal Probing Details}
\label{sec:temporal_details}
\noindent We probe temporal transfer using an MS-TCN head trained on frozen frame-level features. Each batch is one full video. Across models/datasets, we use 3 stages, 8 layers per stage, and 64 feature maps. Optimization: AdamW, lr=0.01, wd=0.01, cosine schedule, 100 epochs. These settings are identical across models for fairness. Results are reported in Table~\ref{tab:temporal_probing}.

\subsection{Zero-shot Prompt Design}
\begin{figure}[h]
  \centering
    \includegraphics[width=0.40\textwidth]{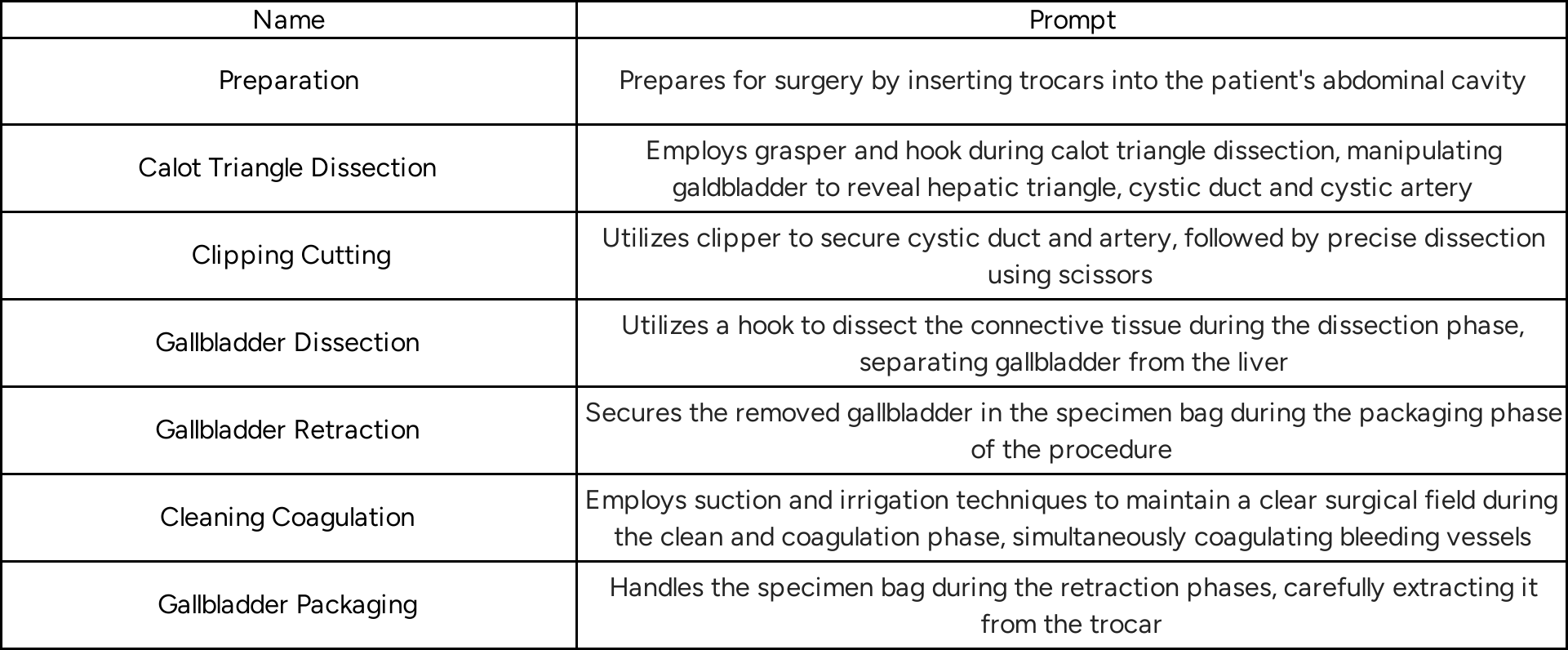}
    \caption{Zero-shot prompts for Cholec80 (phases).}
    \label{fig:cholec80_prompt}
\end{figure}

\begin{figure}[h]
  \centering
    \includegraphics[width=0.40\textwidth]{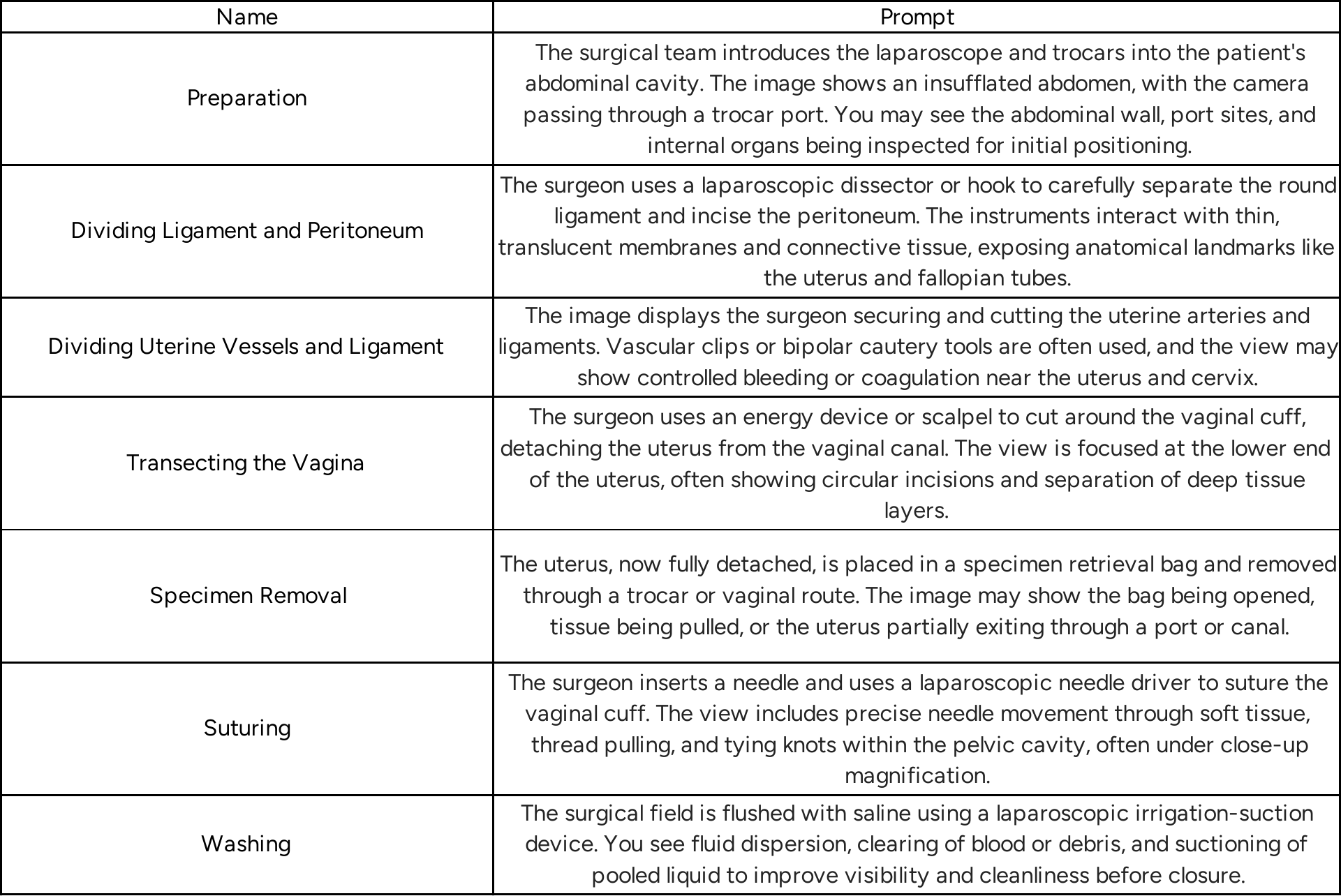}
    \caption{Zero-shot prompts for AutoLaparo (phases).}
    \label{fig:autolaparo_prompt}
\end{figure}

\begin{figure}[h]
  \centering
    \includegraphics[width=0.40\textwidth]{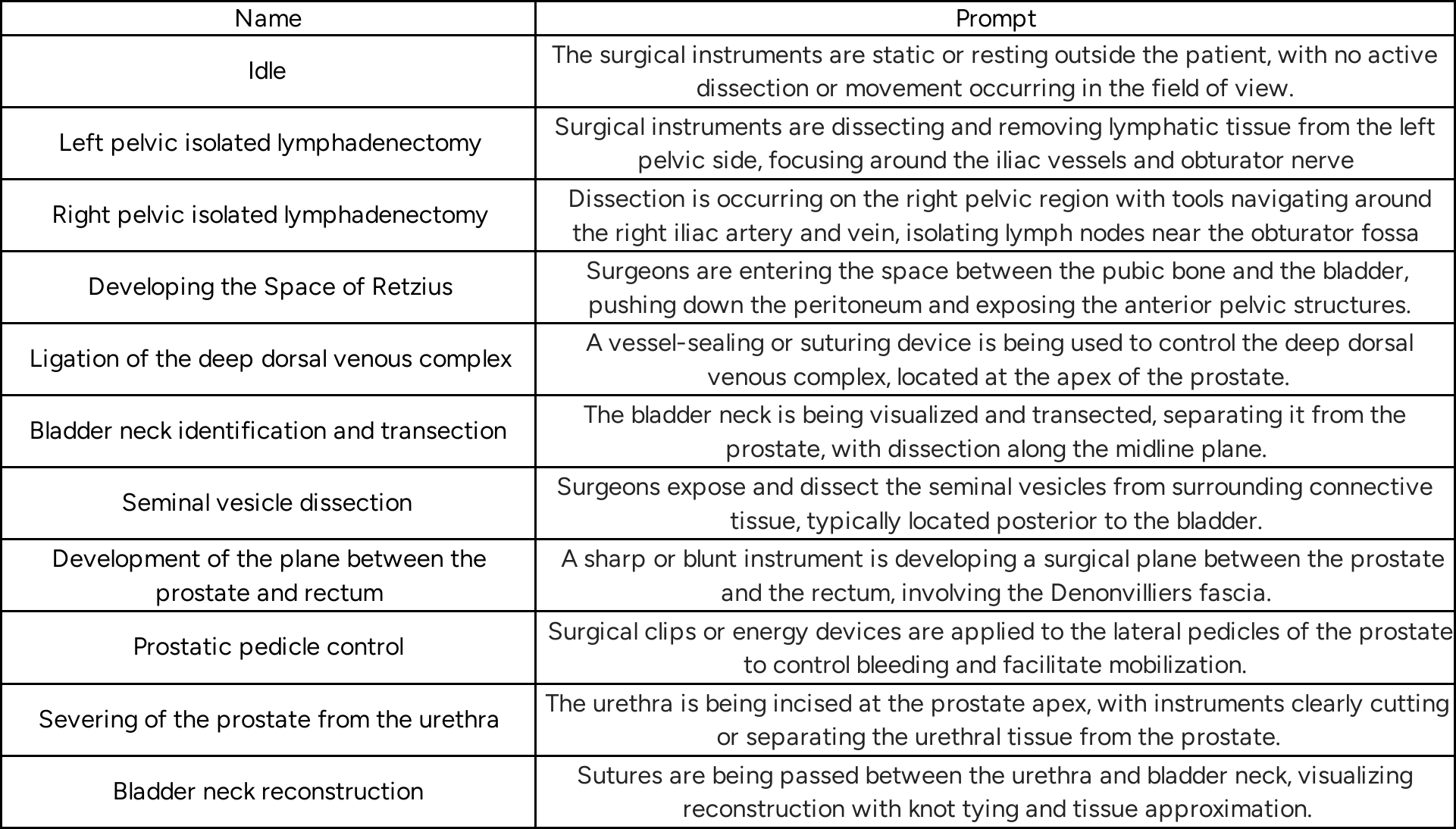}
    \caption{Zero-shot prompts for GraSP (phases).}
    \label{fig:grasp_phases_prompt}
\end{figure}

\begin{figure}[h]
  \centering
    \includegraphics[width=0.40\textwidth]{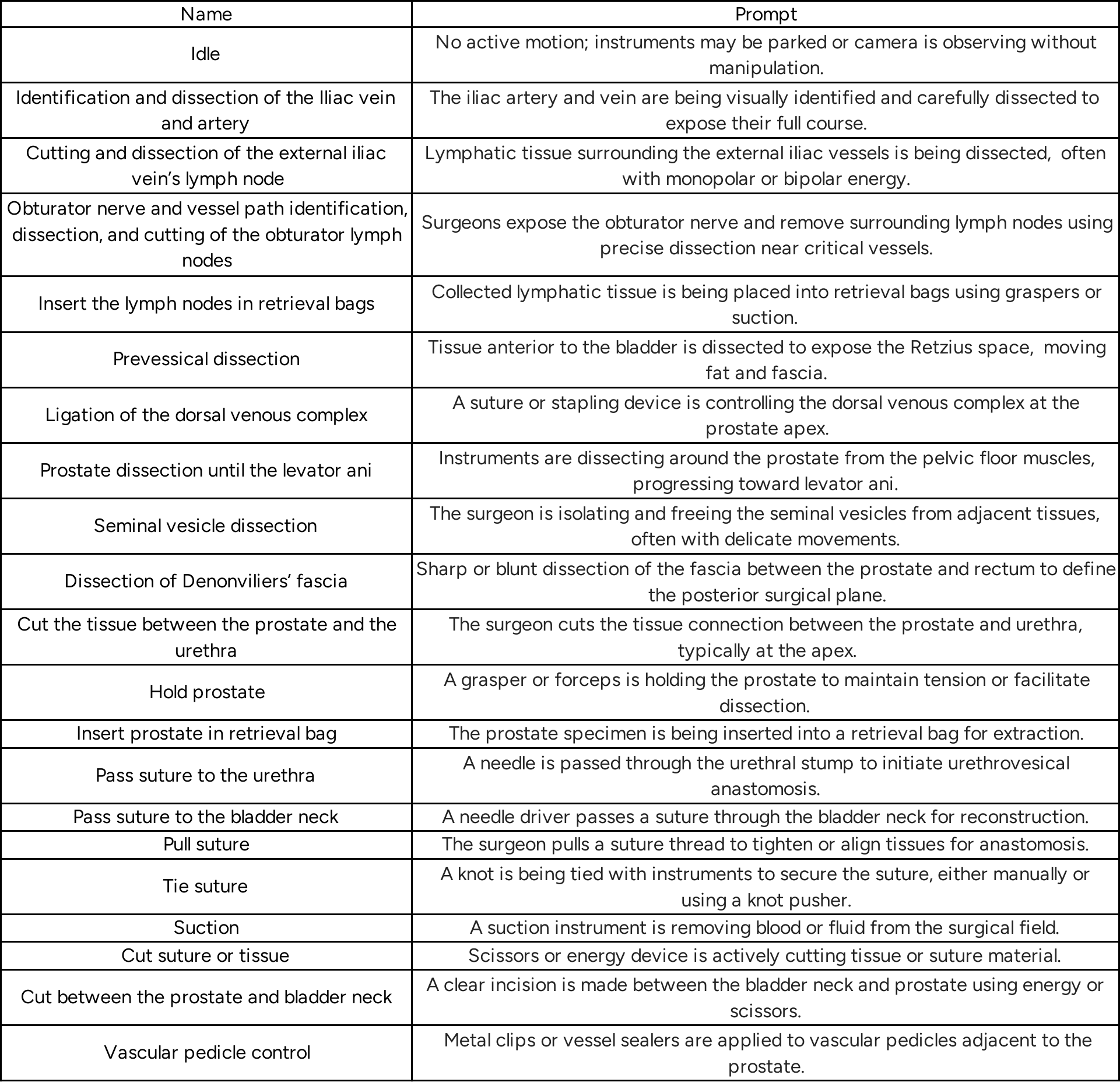}
    \caption{Zero-shot prompts for GRaSP (steps).}
    \label{fig:grasp_steps_prompt}
\end{figure}

\begin{figure}[h]
  \centering
    \includegraphics[width=0.40\textwidth]{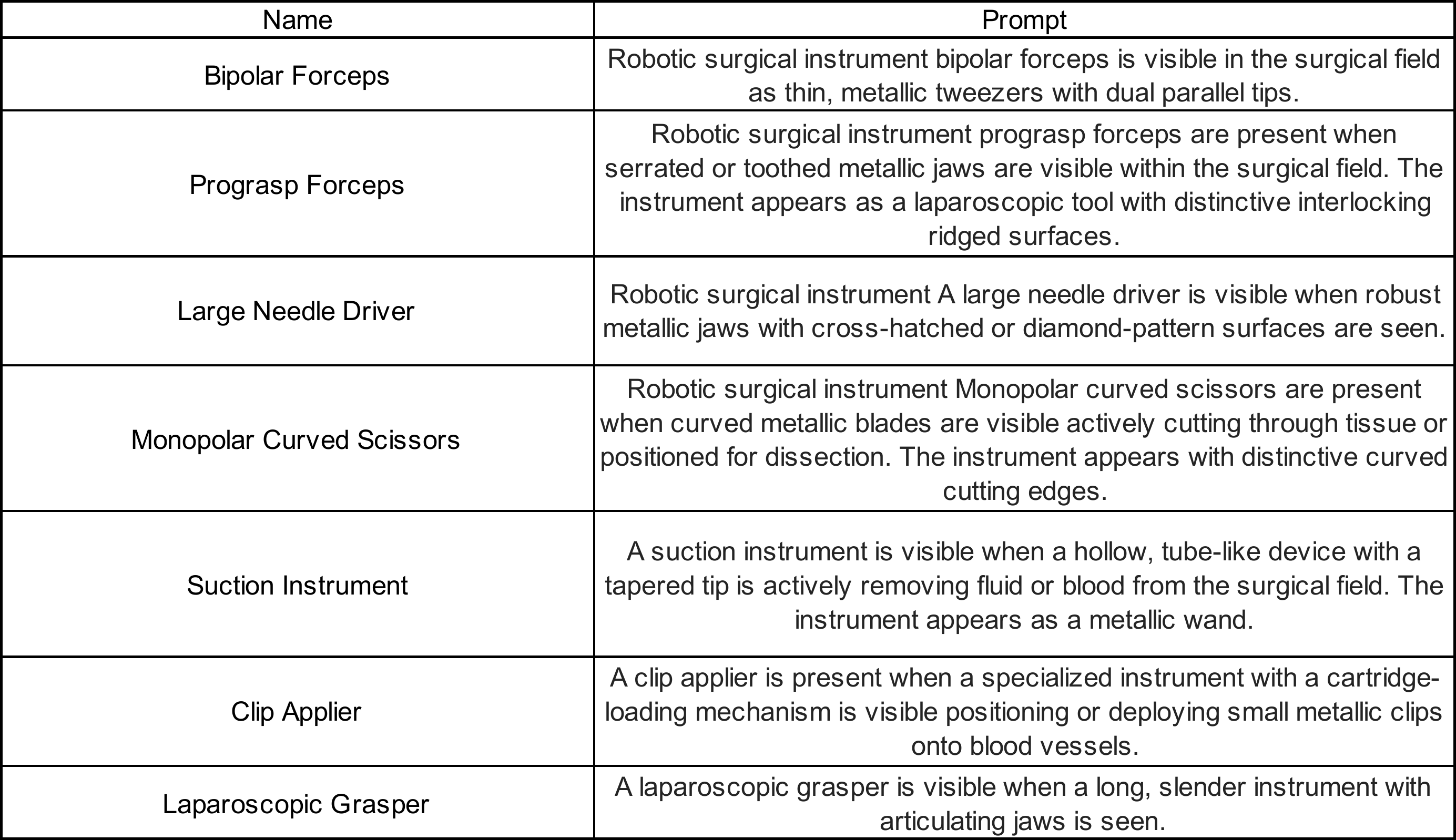}
    \caption{Zero-shot prompts for GRaSP (tools).}
    \label{fig:grasp_tools_prompt}
\end{figure}

\begin{figure}[h]
  \centering
    \includegraphics[width=0.40\textwidth]{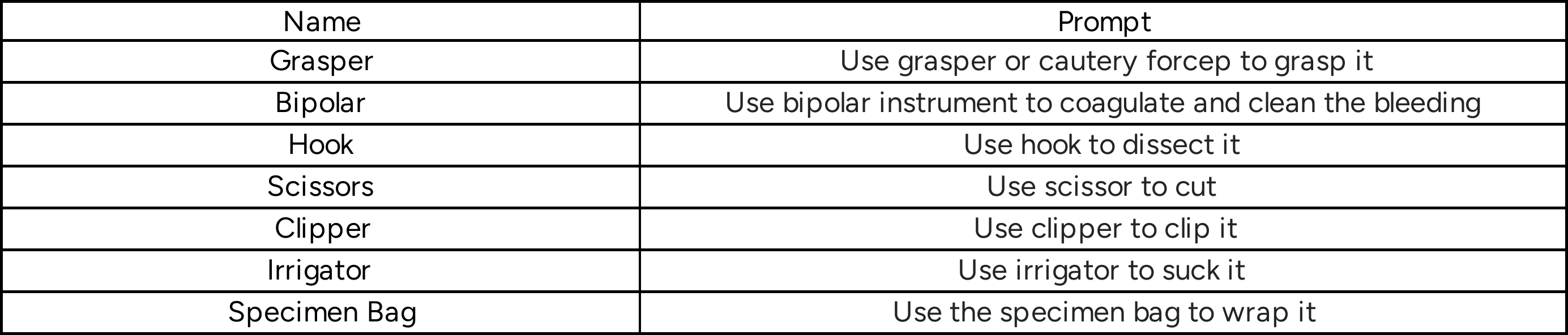}
    \caption{Zero-shot prompts for Cholec80 (tools).}
    \label{fig:cholec80_tools_prompt}
\end{figure}

\begin{figure}[h]
  \centering
    \includegraphics[width=0.45\textwidth]{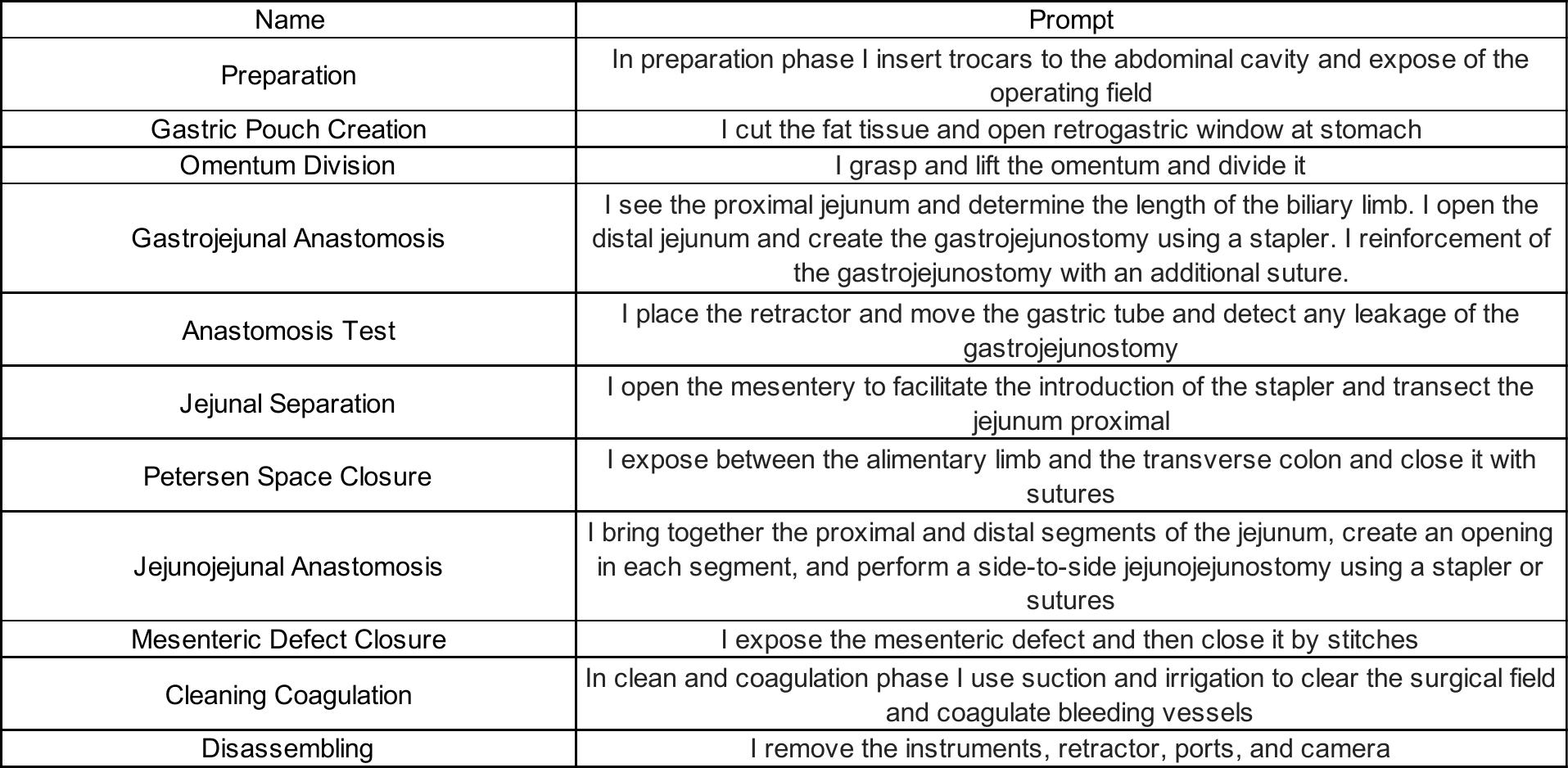}
    \caption{For zero-shot evaluation on BernBypass70 and StrasBypass70 (phases), we excluded the "other" class, following existing zero-shot studies \citep{peskavlp} to ensure fair comparison. }
    \label{fig:multibypass_prompt}
\end{figure}

\begin{figure}[h]
  \centering
    \includegraphics[width=0.40\textwidth]{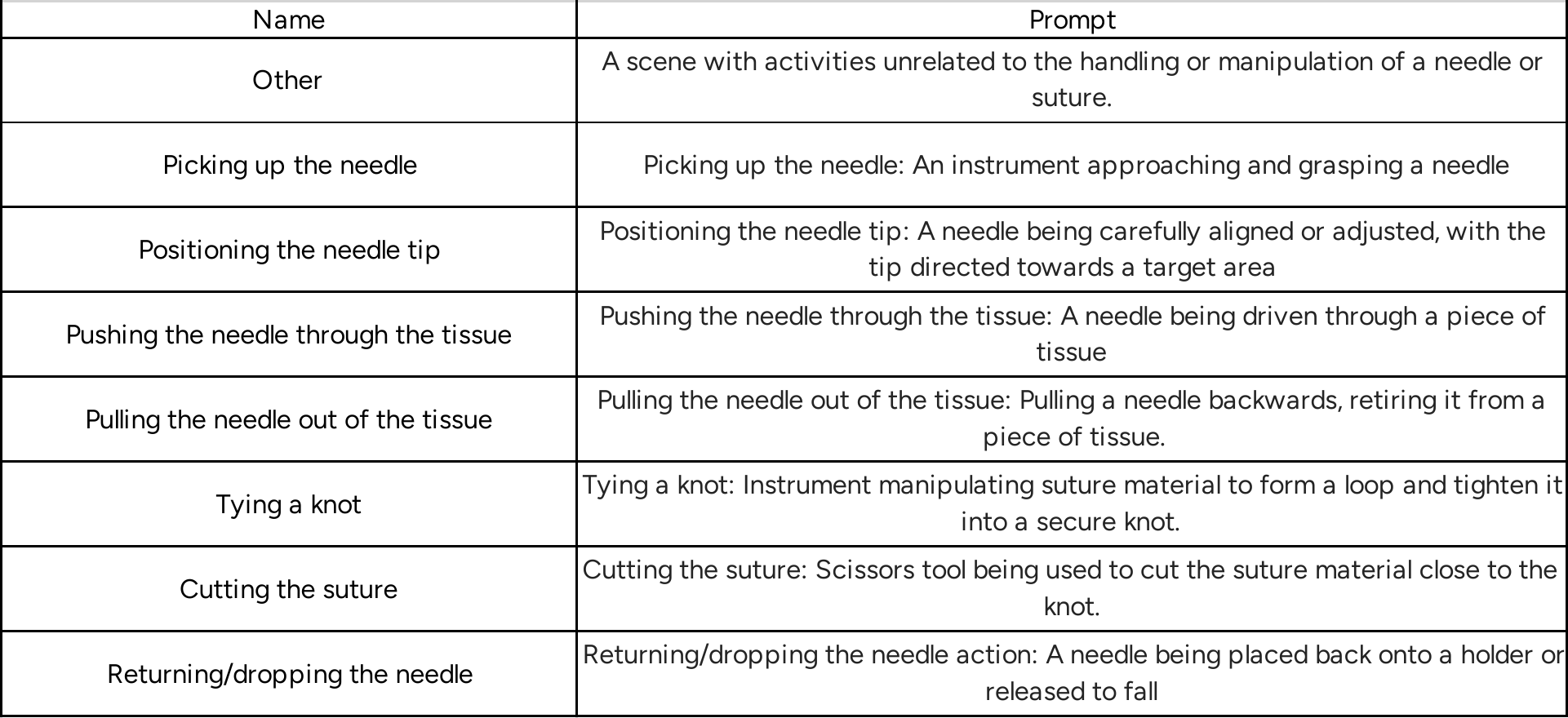}
    \caption{Zero-shot prompts for SAR-RARP50 (phases).}
    \label{fig:sarrarp50_prompt}
\end{figure}

\noindent Figures~\ref{fig:cholec80_prompt}–\ref{fig:sarrarp50_prompt} show all prompts used for zero-shot classification in downstream benchmarks.

\begin{figure}[b!]
  \centering
    \includegraphics[width=0.45\textwidth]{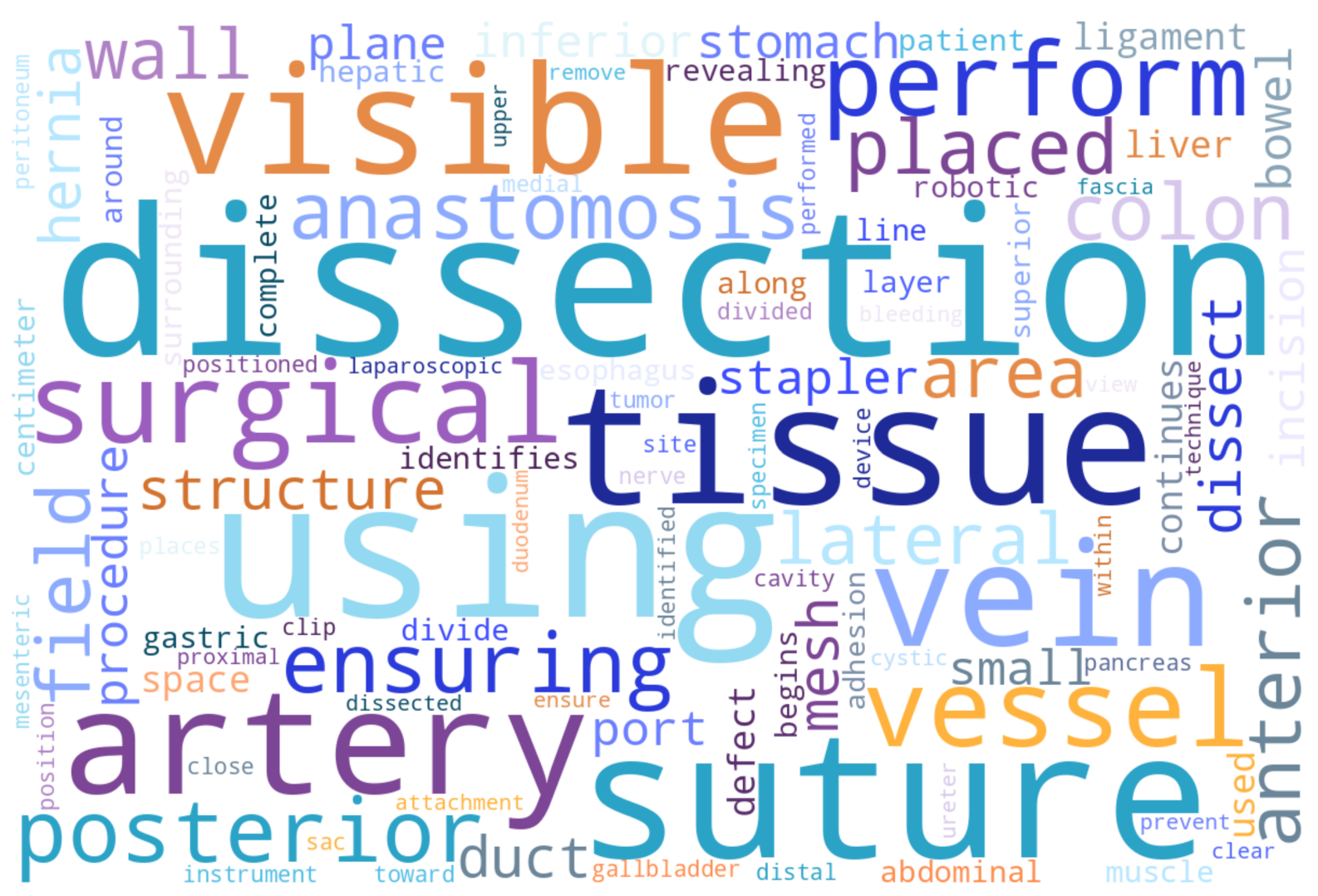}
    \caption{Word cloud of SurgLaVi captions, highlighting frequent surgical vocabulary.}
    \label{fig:wordcloud}
\end{figure}

\begin{figure}[b!]
  \centering
    \includegraphics[width=0.45\textwidth]{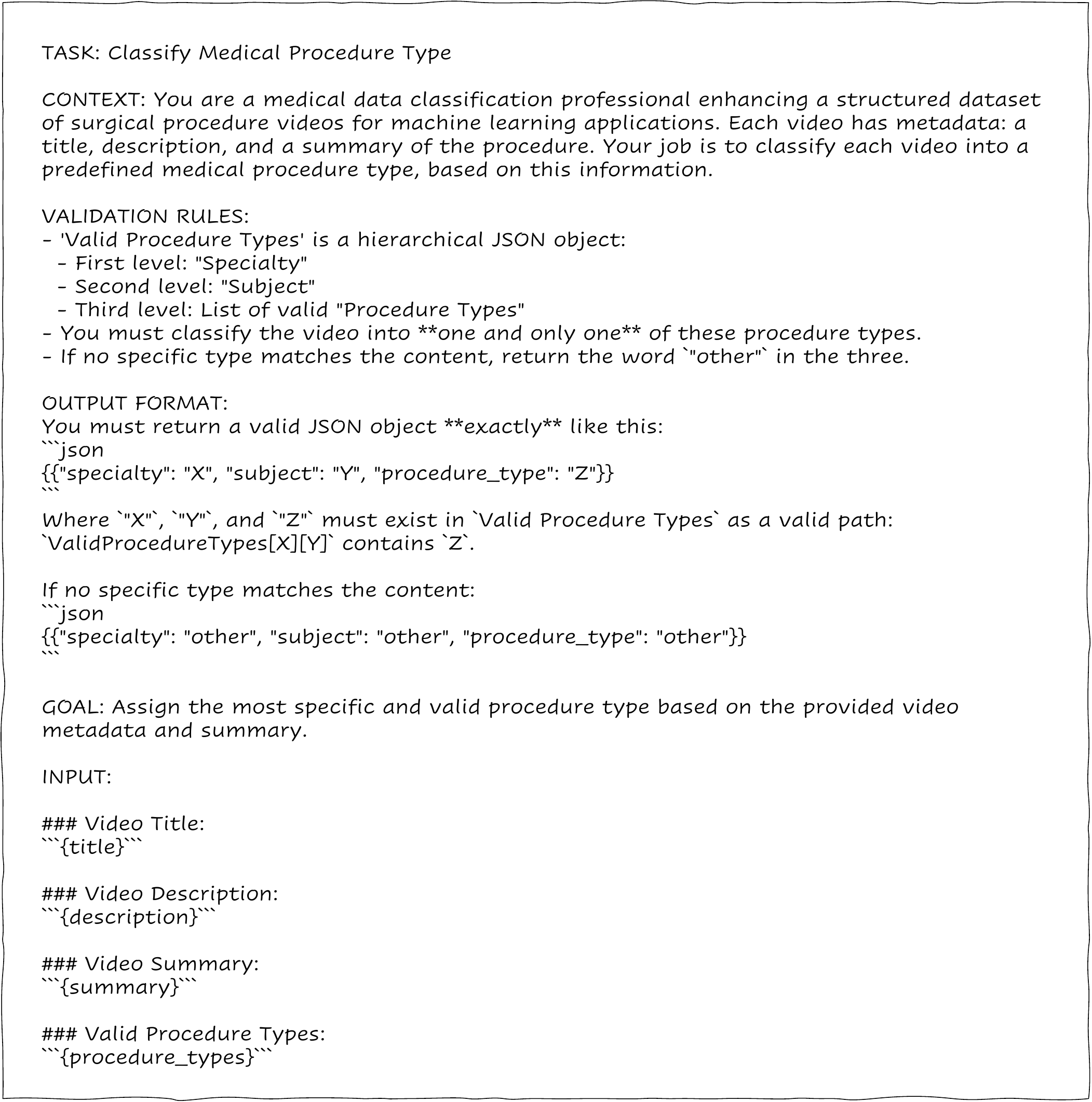}
    \caption{Prompt for assigning specialty, subject, and procedure type.}
    \label{fig:prompt_classification}
\end{figure}

% --------------------------

% --------------------------

\end{document}